\documentclass{article}

\usepackage[eandd, final]{neurips_2026}
\usepackage[table]{xcolor}
\usepackage{tcolorbox}
\tcbuselibrary{breakable}
\usepackage{enumitem}
\usepackage{float}
\usepackage{multirow}
\usepackage{siunitx}
\usepackage{pdflscape}
\usepackage{subcaption}
\usepackage{adjustbox}
\usepackage{booktabs, soul}

\usepackage{wrapfig}
\usepackage{caption}
\usepackage{titletoc}
\usepackage{etoc}
\usepackage{xltabular}
\usepackage{longtable}
\usepackage{lipsum} %

\definecolor{intradayblue}{RGB}{45,105,180}
\definecolor{earningsred}{RGB}{190,65,65}
\definecolor{intradaybg}{RGB}{235,243,252}
\definecolor{earningsbg}{RGB}{253,238,238}

\usepackage{tabularx}
\usepackage{booktabs}
\usepackage{amsmath}
\usepackage{lscape}
\usepackage{amssymb}
\usepackage{graphicx}
\usepackage[utf8]{inputenc} %
\usepackage[T1]{fontenc}    %
\usepackage{hyperref}       %
\usepackage{url}            %
\usepackage{booktabs}       %
\usepackage{amsfonts}       %
\usepackage{nicefrac}       %
\usepackage{microtype}      %

\usepackage{enumitem}

\title{LiveOption: Evaluating LLM Agents in Structured Option Trading with Nonlinear Payoffs}

\author{%
Haochen Luo$^{1}$,
~Yifan Li$^{1}$,
~Binh Minh An$^{1,}$,
~Xiaolong Luo$^{1}$,
~Zhengzhao Lai$^{3}$,\\
~\textbf{Yuan Zhang$^{2,\dagger}$},
~\textbf{Chen Liu$^{1,}$}\thanks{Corresponding authors: zhang.yuan@sufe.edu.cn, chen.liu@cityu.edu.hk}
\\
$^{1}$City University of Hong Kong 
$^{2}$Shanghai University of Finance and Economics \\
$^{3}$The Chinese University of Hong Kong (Shenzhen)
\\
\texttt{\{chester.hc.luo, yi3283-c, binhman2-c, xl.luo\}@my.cityu.edu.hk}
\\
\texttt{zhengzhaolai@link.cuhk.edu.cn}
}

\begin{document}

\maketitle

\begin{abstract}
Large language models (LLMs) and multi-agent systems (MAS) have shown promise in financial decision-making, yet existing evaluations focus on equity trading and primarily assess directional prediction, overlooking the structural complexity of derivative markets. 
Option trading introduces fundamentally different challenges, including nonlinear payoffs and multi-leg strategy construction, requiring structured decisions rather than simple directional bets.
We introduce \textbf{LiveOption}, an evaluation framework for LLM-based agents in option trading. LiveOption formulates the problem as structured sequential decision-making under realistic execution and capital constraints, and provides a reproducible environment with standardized interaction protocols.
The framework includes three task suites covering portfolio overlays, event-driven earnings trading, and 0DTE intraday trading. We further propose a hierarchical metric suite that evaluates action validity, decision quality, risk characteristics, and outcome-level performance.
Experiments show that current agents often fail to achieve competitive returns in most scenarios. LiveOption offers a principled testbed for evaluating structured decision-making beyond outcome-based metrics. Project Website: \href{https://liveoption.github.io/}{https://liveoption.github.io/}

{\color{red}\textbf{Warning:} This paper does not provide any investment advice. The methods in this article may involve substantial risk and could result in significant financial loss.}

\end{abstract}

\section{Introduction}

Recent advances in large language models (LLMs) and multi-agent systems (MAS) have rapidly expanded their applications in AI for Finance. Prior work has demonstrated promising results across a wide range of tasks, including financial text understanding~\citep{chen2021finqa,chen2024fintextqa,xie2023pixiu,reddy2024docfinqa}, alpha factor mining~\citep{luo2026alphabench,shi2025navigating,tang2025alphaagent,liu2025cognitive,wang2025alpha}, market prediction~\citep{abdelsamie2024comparative,fatemi2024finvision,zeng2025futurex,darwish2025stock,wang2024stocktime,shi2025kronos}, and autonomous trading~\citep{yu2024fincon,li2025time,fan2025aitraderbenchmarkingautonomousagents,xiao2024tradingagents,yu2025finmem,li2025investorbench,xiong2025quantagent}. Among these, trading agents have emerged as a particularly impactful direction, as they require sequential decision-making under uncertainty while integrating reasoning, tool use, and risk control.
However, alongside rapid progress, recent work~\citep{kong2026evaluating} critically points out an illusory sense of advancement in current LLM-based financial evaluation, exposing severe biases such as look-ahead bias induced by prior knowledge leakage from training data~\citep{glasserman2023assessing,xu2024benchmarking,lopez2025memorization}. This raises concerns about whether reported performance truly reflects real-world capability. In response, emerging benchmarks have begun to move beyond static backtesting. Dynamic ``live'' evaluation paradigms~\citep{li2025time} aim to mitigate data contamination by enforcing real-time market interaction, while agent-centric frameworks~\citep{fan2025aitraderbenchmarkingautonomousagents} further require models to actively acquire information through tool use in data-uncontaminated environments.
Despite these improvements, existing benchmarks remain confined to relatively narrow settings. Most focus on standard asset classes such as equities or cryptocurrencies, where the action space is typically limited to simple buy, hold, and sell decisions, and the underlying market dynamics are largely linear and direction-driven.

\begin{figure}[h]
        \vspace{-8pt}
    \centering
    \includegraphics[width=0.92\linewidth]{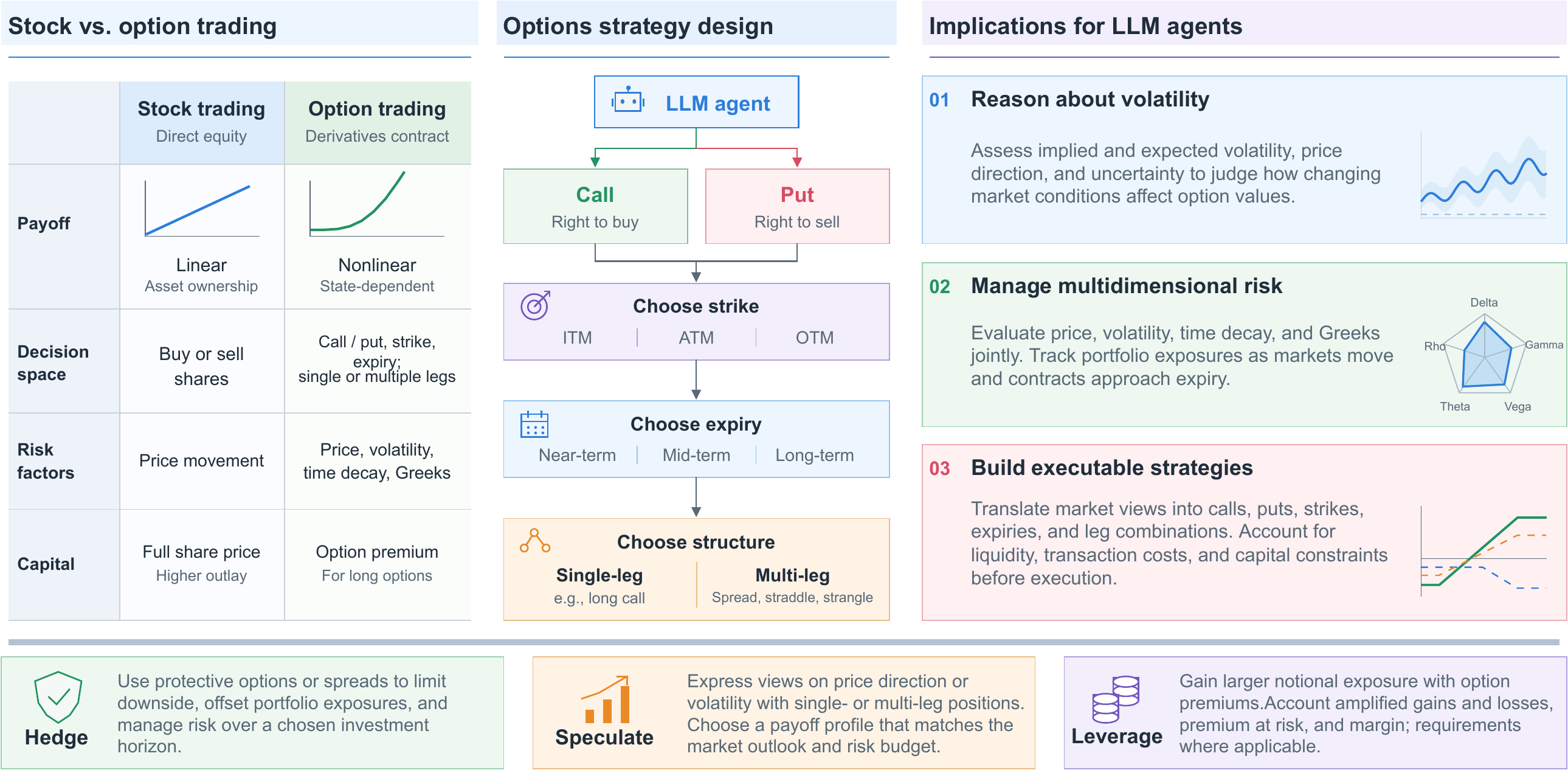}
    \caption{This figure outlines the key differences between stock trading (direct equity) and options trading (derivatives contract). It highlights their differences in properties. Additionally, it presents the different objects of options trading, including hedging, gambling/speculation, and leverage.}
    \label{fig:liveoption_intro}
    \vspace{-8pt}
\end{figure}

In contrast, real financial markets encompass a much richer ecosystem of \textbf{derivative products}, among which \textbf{options} are both highly active and structurally complex. As shown in Figure~\ref{fig:liveoption_intro}, option trading differs fundamentally from stock trading: it introduces nonlinear payoffs, time decay ($\theta$), volatility sensitivity ($\nu$), and strict margin constraints. The decision space is significantly larger, spanning multiple strikes, maturities, and strategy combinations, and risk must be managed across multiple dimensions (the Greeks), rather than relying solely on directional price prediction. We provide a detailed overview of option products and trading rules in Appendix~\ref{sec:preliminary}. These properties create a substantially more challenging and realistic learning environment, which remains largely underexplored in existing agent-based trading research. Previously, machine learning methods have been widely applied to option analysis, including pricing~\cite{amilon2003neural,fan2026machine,chen2012pricing}, forecasting~\cite{liang2009improving,crisostomo2018financial,chen2019forecasting,ayyagari2023developing,vrontos2021implied}, and trading~\cite{ayyagari2023developing,tan2024deep,chen2012pricing,cao2021deep}.
More recently, research has begun to explore how LLMs can support option-related tasks, such as pricing~\citep{banka2025options,dsouza2024leveraging}, hedging~\citep{yang2025dynamic,10.1145/3701716.3715232}, and the retrieval of option strategies from natural language queries~\citep{luo2026from}. However, despite this growing interest, a comprehensive and bias-free evaluation framework for assessing agent-level trading ability across diverse objectives in dynamic options environments remains lacking.

To narrow down this gap, we propose \textbf{LiveOption}, a framework designed specifically for LLM-driven option trading agents. Our central hypothesis is that option markets provide a more realistic and demanding stress test for financial agents than stock-only environments, as they require simultaneous reasoning about volatility dynamics, tail risk, capital efficiency, and the construction of structured strategies. Furthermore, option trading inherently embodies an online sequential decision process: agents must act under continuously evolving market conditions while satisfying strict risk and margin constraints at each step. To faithfully capture this complexity, LiveOption is built around two core design principles. First, it provides a multi-scenario evaluation suite spanning diverse trading objectives and constraint regimes, covering portfolio overlay, earnings-event-driven speculation, and intraday 0DTE trading, together reflecting the full spectrum of practical option use cases. Second, it incorporates a fully-featured backtesting engine that simulates realistic brokerage-style execution, including margin enforcement and multi-leg position management, ensuring agents are evaluated under conditions that closely mirror live market operation. The contributions of this work are:

\textit{\textbf{(a) Option-Centric Trading Agents.}} To the best of our knowledge, we present the first systematic study of LLM-based trading agents in option markets, explicitly addressing the structural challenges introduced by derivative products and establishing a foundation for future research in this domain.

\textit{\textbf{(b) A Dedicated Option Trading Benchmark.}} We establish a live evaluation benchmark covering three representative task scenarios: (i) portfolio overlay, (ii) earnings-event-driven trading, and (iii) intraday 0DTE trading. The benchmark formalizes observation spaces, action constraints, execution modeling, and evaluation metrics tailored to option markets, while providing a scalable infrastructure that supports user-defined scenario construction beyond the predefined tasks.

\textit{\textbf{(c) Behavioral Analysis and Empirical Insights.}} We provide a comprehensive behavioral analysis suite that examines agent decision-making patterns, risk management discipline, and execution quality beyond outcome-level metrics. Through extensive experiments across multiple state-of-the-art LLMs, we identify systematic failure modes unique to option trading and derive actionable insights to guide future agent design.

\section{Framework of LiveOption}

\subsection{Overall Design}

\begin{figure}[h]
    \centering
    \includegraphics[width=0.95\linewidth]{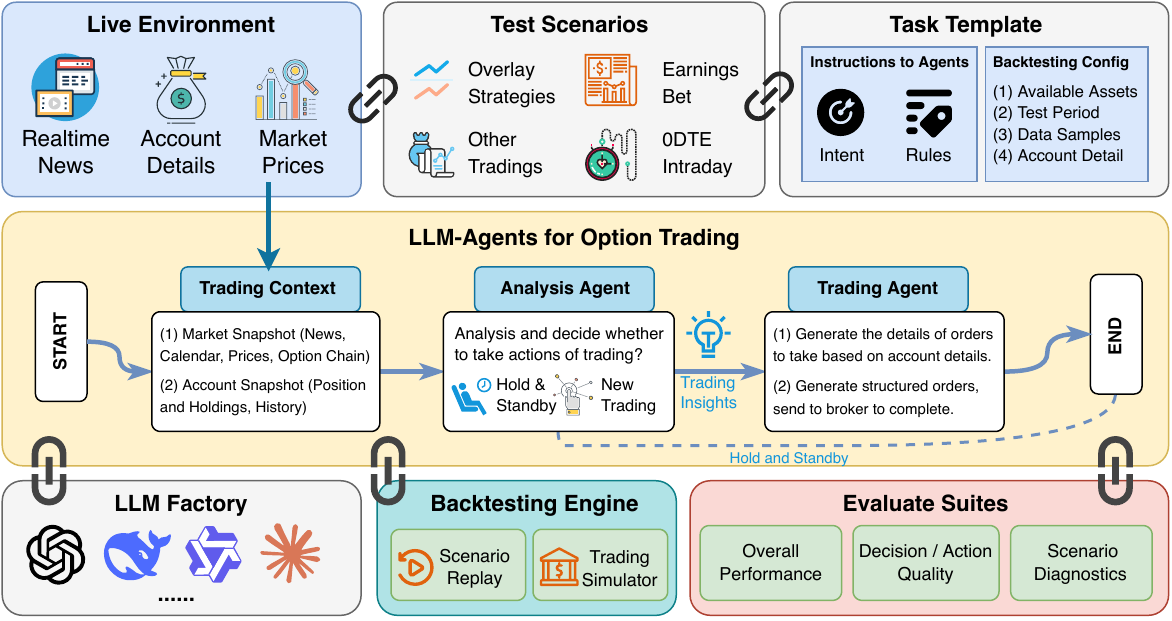}
    \caption{LiveOption framework for option trading, integrating live market context, scenario-based testing, agent-driven analysis and execution, backtesting, and multi-level evaluation.}
    \label{fig:liveoption_framrwork}
\end{figure}

As shown in Figure~\ref{fig:liveoption_framrwork}, LiveOption is designed to evaluate real-time option trading in a live environment. It consists of four main components: (1) A live environment specifying the available information, tasks, and user configurations; (2) A multi-agent system based on various LLMs for option trading; (3) An engine for executing backtesting; (4)  An evaluation suite to load specific metrics for assessing option trading performance.

\vspace{-14pt}
\subsection{Live Environments}

The live environment in LiveOption specifies the contexts of option trading, including the available information, the tasks and the user configurations. %
To cover a broad range of practical option use cases and investment intentions, LiveOption studies three tasks with different trading frequencies, investment styles, and trading rules. These tasks capture the core roles of options as in Figure~\ref{fig:liveoption_intro}. 
They offer a practical and challenging testbed for evaluating agent capabilities. We design one dedicated scenario for each role, with detailed specifications provided in Appendix~\ref{appendix:scenarios_design}.

\textbf{\textit{Overlay Strategies:}} This task focuses on utilizing specific option strategies, such as covered calls and protective puts, to enhance returns or hedge risk in a portfolio. The strategies are employed with longer expirations and are typically out-of-the-money (OTM), allowing for income generation or downside protection without actively trading the underlying asset.

\textbf{\textit{Earnings Bet:}} This task centers around short-term trading surrounding earnings announcements. The agent places options trades based on predictions related to earnings results, targeting significant price movements that typically occur before and after the earnings report. The agent can dynamically adjust its positions in response to changing news and price actions.

\textbf{\textit{0DTE Intraday:}} The 0DTE (Zero Days to Expiration) options scenario tests the agent's ability to make quick, high-frequency decisions on options that expire within the same trading day. The agent needs to leverage event-driven price movements, such as news releases or economic data, and manage rapid time decay, making split-second decisions before the options expire.

To improve the framework's flexibility and support user-customized scenarios, LiveOption enables users to customize trading environments with different objectives (e.g., return maximization, risk control, income generation), portfolio constraints (e.g., initial positions, capital allocation, leverage limits), and trading frequencies (e.g., intraday, daily, or event-driven). This extension allows researchers and practitioners to tailor evaluation settings to specific use cases, enabling more targeted analysis of agent behavior under diverse and realistic market conditions.

\subsection{LLM-based Trading Agents}

Option trading presents unique challenges for agent-based decision-making, including high-dimensional option chain data and a combinatorial action space under strict feasibility constraints. Unlike equity trading, where actions are typically directional, option trading requires structures that stay valid under margin, liquidity and contract-consistency rules.
To enable consistent and model-agnostic evaluation, we adopt a generic \emph{analyzer--executor} paradigm, which is widely used in practical agent systems~\citep{li2025time,fan2025aitraderbenchmarkingautonomousagents,xiong2025quantagent,li2025investorbench,lin2026squidagentparallelizewiselycoordinate}. In this abstraction, the \emph{analyzer} is responsible for interpreting market conditions and generating high-level trading decisions. The \emph{executor}, in turn, translates these decisions into executable orders under real-world constraints. We adapt this paradigm to the option trading domain through several key design choices:

\textbf{\textit{Compact volatility representation (AVSR):}}
    Instead of exposing the full implied volatility surface, we summarize it using a small set of anchor points defined over moneyness (or delta) and time-to-expiry. This $3\times3$ representation captures the essential structure of volatility (level, term structure, and skew) while remaining stable and comparable across assets The details are introduced in Appendix~\ref{appendix:avsr}.

\textbf{\textit{Quantized action space (QOCR):}}
    To reduce the combinatorial complexity of option chains, actions are specified in a discrete, structured form, including expiry buckets, moneyness levels, and option types. This abstraction enables efficient decision-making without requiring the agent to select from hundreds of individual contracts directly. We provide more details for it in Appendix~\ref{appendix:qocr}

\textbf{\textit{Backend order verification:}}
    All agent-generated option orders pass through a backend verification module that enforces realistic trading constraints, including margin requirements, position limits, and contract validity (strike, expiry, and multiplier). This layer ensures that only executable orders are admitted into the environment.

\subsection{Backtesting Engines}
The core infrastructure of LiveOption is a backtesting engine designed specifically for options. Unlike stock simulators that rely on simple buy/sell and mark-to-market logic, option evaluation demands contract-aware mechanics and rich market context. Therefore, the engine is organized into two layers: a \textbf{Scenario Replay} layer that reconstructs realistic market conditions, and a \textbf{Trading Simulator} layer that maintains brokerage-style account state and enforces realistic constraints.

\textbf{\textit{Scenario Replay:}}
The scenario replay layer streams time-indexed market states in chronological order, supporting both daily and minute-level resolution through configurable bars. It provides a deterministic step function that allows different agents to be evaluated under identical market conditions, ensuring reproducibility and fair comparison. At each step, the replay delivers a synchronized observation packet comprising the underlying price series, option-chain snapshots, real-time news feeds, and scheduled financial event calendars, giving agents the same rich information context they would encounter in live trading. Because raw option chains are large and highly redundant, the engine does not expose the full chain by default. Instead, a processing module filters and summarizes contracts according to liquidity and task constraints, producing a compact, agent-friendly observation that includes volatility surface and skew summaries, tradable contract candidates, and portfolio risk digests (details in Appendix~\ref{appendix:evaluation_metrics}). Together, these components form a live benchmark infrastructure that faithfully replicates the continuous flow of prices, events, and market signals that real option traders must navigate, while maintaining scalability and avoiding context overload.

\textbf{\textit{Trading Simulator:}}
Built on top of scenario replay, the trading simulator maintains a brokerage-style account state, including cash, open positions by contract, realized and unrealized PnL, transaction fees, and margin usage. At each step, it updates valuations according to core option mechanics, including contract multipliers, time decay, expiry settlement, and simplified assignment/exercise rules, so agents are evaluated under realistic trading constraints.
The simulator also supports multi-leg strategies and contract lifecycle events. It can open and manage spreads, straddles, strangles, condors, and calendars as grouped positions, while tracking fills, PnL, and Greeks for each leg. In addition, it handles expiry and forced-close policies, such as auto-close near expiry for intraday tasks, preventing agents from benefiting from unrealistic settlement assumptions.

\subsection{Evaluation Suites}

LiveOption provides a comprehensive evaluation workflow, including a web interface that enables users to conveniently conduct backtesting and visualize results, which was further introduced in Appendix~\ref{appendix:web}. The evaluation is carried out from two perspectives: (1) absolute performance metrics for task outcomes; (2) a set of analytical suites designed to examine agent behavior.

\textbf{\textit{Task Performance Evaluation.}} Since LiveOption contains scenarios with fundamentally different temporal structures, we evaluate task outcomes at two levels. At the \textit{portfolio level}, used for long-horizon overlay tasks, we report Annualized Return (AR), Sharpe Ratio (SR), and Maximum Drawdown (MDD) for absolute performance, together with benchmark-relative metrics (active return and Information Ratio) that isolate the value added by the option overlay over a passive buy-and-hold leg. At the \textit{episode level}, used for Earnings Bet and 0DTE Intraday where each trading day or event window is an independent unit, we report Win Rate, mean and median return (PnL\%), payoff ratio (P/L), profit factor (PF), and the no-trade rate, aggregated across episodes. Mean returns carry percentile bootstrap intervals from $10{,}000$ episode resamples. Cross-model and ablation comparisons are paired on shared episodes and tested with two-sided sign-flip tests under Holm correction within each comparison family. Overlay paths are continuous rather than episodic, so their Sharpe ratios are intervaled with a stationary block bootstrap instead (Appendix~\ref{appendix:detail_overlay}). This two-level design captures both the smoothed, long-run behavior of overlay strategies and the trade-level outcome distribution that dominates short-horizon option trading. Details are provided in Appendix~\ref{appendix:common_metrics}.

\textbf{\textit{Behavioral Analysis Evaluation.}} Outcome metrics summarize \textit{what} agents achieve but not \textit{how}. LiveOption therefore preserves the complete decision trace of every run, including thesis text, intent stream, fill log, and Greek snapshots, and uses it to diagnose agent behavior along three axes: \textit{strategy composition}, the empirical distribution over option structures actually traded; \textit{directional vs.\ structural reasoning}, classifying each trade by whether the LLM's thesis is dominated by a directional view or by explicit option-mechanic considerations (Greeks, IV-vs-realized, vega exposure); and \textit{position sizing and leverage control}, tracking capital deployed per trade and the effective leverage carried between sessions. We further decompose realized PnL into delta, gamma, theta, and vega contributions (Appendix~\ref{sec:pnl_decomposition}) to attribute returns to specific Greek exposures, distinguishing genuine option-aware edges from incidental directional gains.

\section{Benchmarks, Analysis and Discussion}
\label{sec:analysis}
\subsection{Experimental Settings}\label{sec:experimental}

\textbf{Data Integration.} 
LiveOption is deployed on QFinZero~\cite{luo-etal-2026-qfinzero}, which provides unified data access and simulated brokerage services. Historical market data are obtained from commercial third-party providers, while financial news is ingested through QFinZero's news pipeline. The inputs include underlying prices, option chains, news records, and financial event metadata. Users can also import their own market and news data by converting them to the documented schemas through the preprocessing pipeline. QFinZero serves these inputs through time-aligned interfaces for agent observations and replay. Data sources, schemas, and preparation procedures are described in Appendix~\ref{appendix:data}.

\textbf{Evaluation Period.} 
Although LiveOption supports flexible configuration to evaluate agents over arbitrary time ranges, we restrict the evaluation period to the full year of 2025 in this work. We select year 2025 for two main reasons: (1) LiveOption interacts with real-world financial data, and restricting the window to 2025 limits information leakage from pretraining for the models whose knowledge cutoff precedes it; (2) The market conditions in 2025 exhibit substantial diversity, including multiple regime shifts and high-volatility episodes, providing a comprehensive testbed for evaluating agent robustness under extreme or black-swan-like events.

\textbf{LLMs.} We evaluate eight LLMs spanning open-weight and proprietary families, with inference cost considered in model selection: DeepSeek-V4-Flash~\cite{deepseekai2026v4}, GPT-OSS-120B~\cite{openai2025gptoss}, Qwen3-235B-A22B~\cite{qwen3}, Llama-3.3-70B-Instruct~\cite{grattafiori2024llama3}, MiniMax-M3~\cite{lai2026minimax}, GLM-5.3-Flash~\cite{glm2026glm5,zai2026glm53}, GPT-5.6-Terra~\cite{openai_gpt56_terra}, and Gemini-3-Flash~\cite{google_gemini3_flash}. The first six models are accessed through a common inference provider, while the two proprietary models use separate gateways. All models are evaluated at temperature $0.5$ under the same task protocols. Model configurations, knowledge cutoffs, API identifiers, and detailed evaluation settings are provided in Appendix~\ref{appendix:eval_llm} (Table~\ref{tab:llm_detail}).

\begin{figure}[h]
    \centering
    \includegraphics[width=0.75\linewidth]{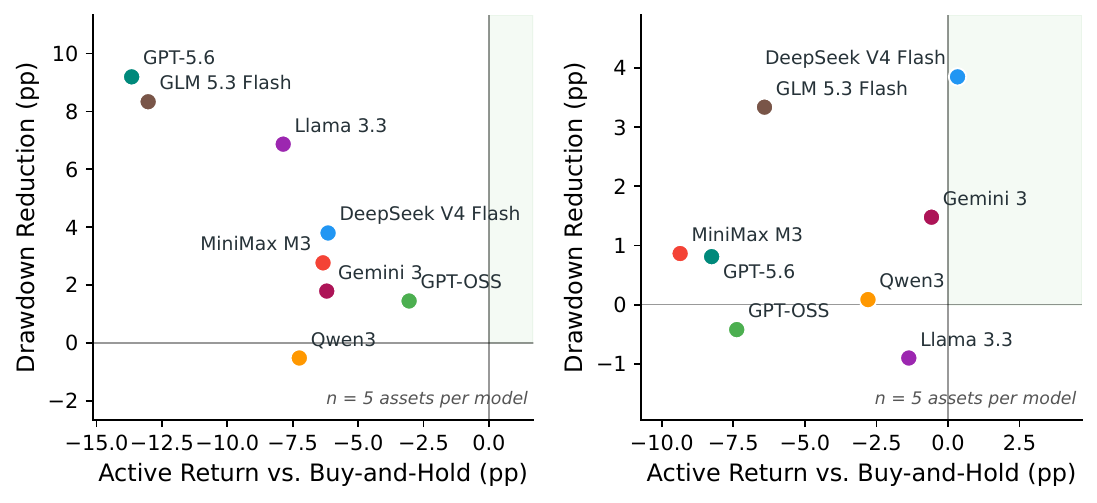}
    \caption{Overlay active return against drawdown reduction, both against the matched buy-and-hold portfolio, for Hedging (left) and Covered Call (right). Each point is a model's mean over five assets, and the shaded quadrant is where both improve.}
    \label{fig:overlay_tradeoff}
    \vspace{-12pt}
\end{figure}

\subsection{Overall Performance}

\begin{wraptable}{r}{0.5\textwidth}
\vspace{-\intextsep}
\centering
\caption{Overlay performance against the matched buy-and-hold portfolio, five assets per model and mandate. $\Delta$Ret is mean active return and $\Delta$MDD the mean drawdown reduction (\%).}
\label{tab:overlay_performance}
\setlength{\tabcolsep}{3pt}
\scriptsize
\begin{tabular}{@{}l rr rr@{}}
\toprule
\multirow{2}{*}{\textbf{Model}} & \multicolumn{2}{c}{\textbf{Hedging}} & \multicolumn{2}{c}{\textbf{Covered Call}} \\
\cmidrule(lr){2-3} \cmidrule(l){4-5}
 & \textbf{$\Delta$Ret} & \textbf{$\Delta$MDD} & \textbf{$\Delta$Ret} & \textbf{$\Delta$MDD} \\
\midrule
DeepSeek-V4-Flash & $-$6.15 & 3.80 & 0.34 & 3.85 \\
GPT-OSS-120B & $-$3.05 & 1.45 & $-$7.38 & $-$0.42 \\
Qwen3-235B-A22B & $-$7.25 & $-$0.52 & $-$2.79 & 0.09 \\
Llama-3.3-70B-Instruct & $-$7.86 & 6.87 & $-$1.37 & $-$0.90 \\
MiniMax-M3 & $-$6.34 & 2.76 & $-$9.36 & 0.86 \\
GLM-5.3-Flash & $-$13.02 & 8.33 & $-$6.41 & 3.34 \\
GPT-5.6-Terra & $-$13.65 & 9.19 & $-$8.26 & 0.81 \\
Gemini-3-Flash & $-$6.20 & 1.79 & $-$0.57 & 1.48 \\
\bottomrule
\end{tabular}
\vspace{-1em}
\end{wraptable}

Table~\ref{tab:overlay_performance} and Table~\ref{tab:consolidated_metrics} report performance across the three option-trading scenarios. In the Overlay scenario, active return against the matched buy-and-hold portfolio is negative for all eight models under hedging and for seven of eight under covered-call writing, and the mean information ratio is negative in all sixteen model-by-mandate cells, so no model delivers a risk-adjusted edge over passive equity exposure; per asset (Appendix~\ref{appendix:detail_overlay}) ten of eighty cells are positive, none of them by a wide margin. The drawdown effect is not uniform. Hedging reduces mean maximum drawdown for seven of eight models, by $1.45$ to $9.19$ percentage points, and increases it for Qwen3-235B-A22B by $0.52$ points; covered-call writing reduces it for six models and increases it for two. With five assets per model and mandate these are descriptive patterns, not population estimates. Per asset (Table~\ref{tab:overlay_per_asset}) at most one of five is positive under hedging and three of five under covered-call writing, so overlays here change the risk profile rather than adding alpha. Figure~\ref{fig:overlay_tradeoff} plots the two dimensions against each other: no model reaches the quadrant in which active return and drawdown both improve under hedging, and only DeepSeek-V4-Flash does so under covered-call writing.
In Earnings, all eight models have negative median returns, and seven also have negative means and profit factors below one. GPT-5.6-Terra is the only exception, with a positive mean of $0.40\%$ and a profit factor of $1.16$, but its negative median ($-0.68\%$) and bootstrap interval spanning zero suggest that gains are driven by a few favorable cases. Only MiniMax-M3 has a bootstrap interval excluding zero. Among 28 pairwise comparisons, only MiniMax-M3 versus GPT-5.6-Terra remains significant after Holm correction ($-2.22$ percentage points, adjusted $p=0.047$). Thus, the results distinguish the extremes but do not establish a reliable model ranking.
In 0DTE, mean returns range from $-0.73\%$ to $+0.14\%$, but all models have negative medians, all bootstrap intervals include zero, and no pairwise differences survive Holm correction. The three models with slightly positive means also lose their gains under more realistic execution costs. Replacing the nominal \$0.0001 slippage with the observed median quoted half-spread of \$0.0050 per contract eliminates all positive mean returns and reduces the pooled mean from $-0.216\%$ to $-0.452\%$ ($p=0.013$, sign-flip test). The resulting changes in model ranking further highlight the sensitivity to trading costs (Appendix~\ref{appendix:liquidity}).

Taken together the three scenarios give a consistent negative result on outcome and a positive one on diagnosis. Typical episode returns are negative in both episodic tasks, with every median below zero in each, overlay active returns are mostly negative while the drawdown effect goes both ways, and only the widest earnings gap in the panel is statistically separable, with no separable difference at all in 0DTE. What does vary measurably across models is behaviour: participation, tenor policy, structure mix and tail risk. This motivates reporting distributions and interventions rather than mean returns alone, and it is the view the remainder of this section takes.

\begin{table}[ht]
    \vspace{-10pt}
\centering
\caption{Episode-level performance on the two episodic tasks. Mean return carries a percentile 95\% bootstrap interval from 10{,}000 episode resamples; $n$ is the number of valid episodes, NT the no-trade rate, PF the profit factor.}
\label{tab:consolidated_metrics}
\setlength{\tabcolsep}{3pt}
\footnotesize
\begin{adjustbox}{width=\linewidth}
\begin{tabular}{l r l r r r r r l r r r r}
\toprule
\multirow{2}{*}{\textbf{Model}} & \multicolumn{6}{c}{\textbf{Earnings Bet}} & \multicolumn{6}{c}{\textbf{0DTE Intraday}} \\
\cmidrule(lr){2-7} \cmidrule(l){8-13}
 & \textbf{$n$} & \textbf{Mean \% [95\% CI]} & \textbf{Med \%} & \textbf{Win\%} & \textbf{NT\%} & \textbf{PF} & \textbf{$n$} & \textbf{Mean \% [95\% CI]} & \textbf{Med \%} & \textbf{Win\%} & \textbf{NT\%} & \textbf{PF} \\
\midrule
DeepSeek-V4-Flash & 94 & $-$0.59 [$-$2.67, 1.54] & $-$1.28 & 45.7 & 0.0 & 0.86 & 102 & $-$0.73 [$-$1.41, 0.02] & $-$2.12 & 29.4 & 0.0 & 0.59 \\
GPT-OSS-120B & 95 & $-$0.20 [$-$1.91, 1.60] & $-$0.53 & 44.2 & 2.1 & 0.94 & 102 & 0.14 [$-$1.20, 1.58] & $-$3.20 & 32.4 & 0.0 & 1.05 \\
Qwen3-235B-A22B & 94 & $-$1.23 [$-$3.57, 1.38] & $-$1.20 & 30.9 & 9.6 & 0.72 & 102 & $-$0.36 [$-$0.83, 0.13] & $-$0.70 & 36.3 & 0.0 & 0.67 \\
Llama-3.3-70B-Instruct & 90 & $-$0.37 [$-$2.24, 1.66] & $-$0.78 & 34.4 & 4.4 & 0.89 & 102 & $-$0.59 [$-$1.41, 0.31] & $-$2.18 & 25.5 & 0.0 & 0.70 \\
MiniMax-M3 & 95 & $-$1.81 [$-$3.37, $-$0.26] & $-$1.61 & 38.9 & 0.0 & 0.54 & 102 & 0.14 [$-$0.36, 0.72] & $-$0.26 & 39.2 & 0.0 & 1.18 \\
GLM-5.3-Flash & 95 & $-$0.48 [$-$2.32, 1.44] & $-$0.92 & 46.3 & 0.0 & 0.86 & 102 & 0.11 [$-$0.81, 1.17] & $-$1.68 & 33.3 & 0.0 & 1.07 \\
GPT-5.6-Terra & 96 & 0.40 [$-$1.10, 1.95] & $-$0.68 & 41.7 & 4.2 & 1.16 & 102 & $-$0.18 [$-$0.87, 0.55] & $-$1.15 & 36.3 & 0.0 & 0.87 \\
Gemini-3-Flash & 96 & $-$0.68 [$-$2.84, 1.62] & $-$0.17 & 37.5 & 10.4 & 0.83 & 102 & $-$0.11 [$-$0.71, 0.49] & $-$0.08 & 47.1 & 0.0 & 0.91 \\
\bottomrule
\end{tabular}
\end{adjustbox}
    \vspace{-10pt}
\end{table}

\subsection{Challenges for Current LLM-based Agents in Option Trading}

In this section, we further analyze the evaluation results and examine agent behaviors in various tasks. We identify several key phenomena and critical issues, then we investigate the challenges in employing current LLM-based agents in option trading. For the results of each scenario, we conduct a detailed analysis, which is provided in Appendix~\ref{appendix:perf_detail}.

\begin{table}[h]
\vspace{-10pt}
    \centering
    \caption{Overlay instrument choice by mandate ($\mu \pm \sigma$ over option fills). \textbf{Fills} counts executed option contracts over the five assets, \textbf{Moneyness} is strike over underlying at fill ($K/S$), and \textbf{DTE} is days to expiry at entry against prescribed bands of 7--45 days (Income) and 7--60 days (Hedge). Every hedge leg in the record is a put and every income leg a call, so a call-share column is omitted.}
    \label{tab:combined_option_stats}
    \setlength{\tabcolsep}{4pt}
    \footnotesize
    \begin{tabular*}{\linewidth}{@{\extracolsep{\fill}} l rcc rcc @{}}
        \toprule
        & \multicolumn{3}{c}{\textbf{Hedge}} & \multicolumn{3}{c}{\textbf{Income}} \\
        \cmidrule(lr){2-4} \cmidrule(l){5-7}
        \textbf{Model} & \textbf{Fills} & \textbf{Moneyness} & \textbf{DTE} & \textbf{Fills} & \textbf{Moneyness} & \textbf{DTE} \\
        \midrule
        DeepSeek-V4-Flash & 511 & 0.970 $\pm$ 0.079 & 23.2 $\pm$ 13.4 & 222 & 1.057 $\pm$ 0.075 & 25.9 $\pm$ 13.9 \\
        GPT-OSS-120B & 807 & 0.988 $\pm$ 0.057 & 14.1 $\pm$ 11.6 & 852 & 1.037 $\pm$ 0.055 & 12.5 $\pm$ 9.3 \\
        Qwen3-235B-A22B & 703 & 0.951 $\pm$ 0.068 & 16.7 $\pm$ 15.2 & 397 & 1.033 $\pm$ 0.051 & 14.2 $\pm$ 15.6 \\
        Llama-3.3-70B-Instruct & 588 & 0.979 $\pm$ 0.053 & 6.4 $\pm$ 5.4 & 164 & 1.012 $\pm$ 0.037 & 3.3 $\pm$ 4.3 \\
        MiniMax-M3 & 789 & 0.955 $\pm$ 0.078 & 27.1 $\pm$ 12.9 & 405 & 1.063 $\pm$ 0.076 & 23.2 $\pm$ 14.0 \\
        GLM-5.3-Flash & 415 & 0.965 $\pm$ 0.084 & 26.4 $\pm$ 12.8 & 458 & 1.045 $\pm$ 0.064 & 21.5 $\pm$ 14.4 \\
        GPT-5.6-Terra & 456 & 0.993 $\pm$ 0.047 & 28.1 $\pm$ 12.4 & 546 & 1.107 $\pm$ 0.065 & 19.4 $\pm$ 11.1 \\
        Gemini-3-Flash & 233 & 0.973 $\pm$ 0.068 & 30.5 $\pm$ 14.0 & 375 & 1.065 $\pm$ 0.078 & 27.3 $\pm$ 14.3 \\
        \bottomrule
    \end{tabular*}
    \vspace{-10pt}
\end{table}

\subsubsection{Overlay Behavior: Instrument Choice, Tenor and Exposure}

LLM agents underperform the buy-and-hold baseline in the overlay scenario (Table~\ref{tab:overlay_performance}), despite reasonable instrument selection. They do, however, beat a mechanical rule: against deterministic $0.30$-delta covered-call and protective-put overlays run through the same configurations, the panel is ahead by $17.5$ points of active return under the covered call and roughly level under the hedge mandate, where the fixed rule instead buys about twice the drawdown reduction (Appendix~\ref{appendix:detail_overlay}). The moneyness distribution (Table~\ref{tab:combined_option_stats}) shows no lottery-like tilt: agents concentrate around $K/S \approx 0.95$--$0.99$ in Hedge and $1.01$--$1.06$ in Profit. Tenor, by contrast, is not a shared bias but a model property. Mean entry maturity spans $6.4$ to $27.1$ days in Hedge and $3.3$ to $25.9$ days in Profit against prescribed bands of $7$--$45$ days (Income) and $7$--$60$ days (Hedge), so the panel contains both weekly-rolling and monthly-tenor policies; only Llama-3.3-70B-Instruct writes income legs shorter than a week on average.
Figure~\ref{fig:amd-portfolio-trades-2025} to Figure~\ref{fig:tsla-portfolio-trades-2025} show how tenor interacts with the market regime: Hedge agents accumulate positive excess returns through the February--March downtrend as their puts pay off, then give it back once markets stabilise and rolling short-dated puts pays premium without offsetting protection. In Income the symmetric pattern leaves active return below zero for five of six models.

The runs record no short-call assignment event, so the rolling penalty has to be read off the stock leg instead. Every one of the $60$ overlay cases ends the year holding exactly its target share count, and in $56$ of them the stock leg is below target on at most $8$ of the $250$ trading days, which is the opening ramp from a flat account. Llama-3.3-70B-Instruct under the hedge mandate is the exception, below target on $58$, $144$, $164$ and $246$ of the $250$ days on AMD, GOOG, NVDA and TSLA, with a deepest shortfall on each equal to the entire target position, so its hedged-equity return on those assets is earned with the equity leg absent or incomplete for much of the year.
What the record supports is therefore narrower than a retail-behaviour verdict: instrument selection is sensible, tenor policy differs by model rather than skewing short, and the one case of sustained exposure drift belongs to a single model and mandate.

\subsubsection{Action Concentration Within Tasks}

\begin{figure}[h]
    \vspace{-10pt}
    \centering
    \includegraphics[width=0.82\linewidth]{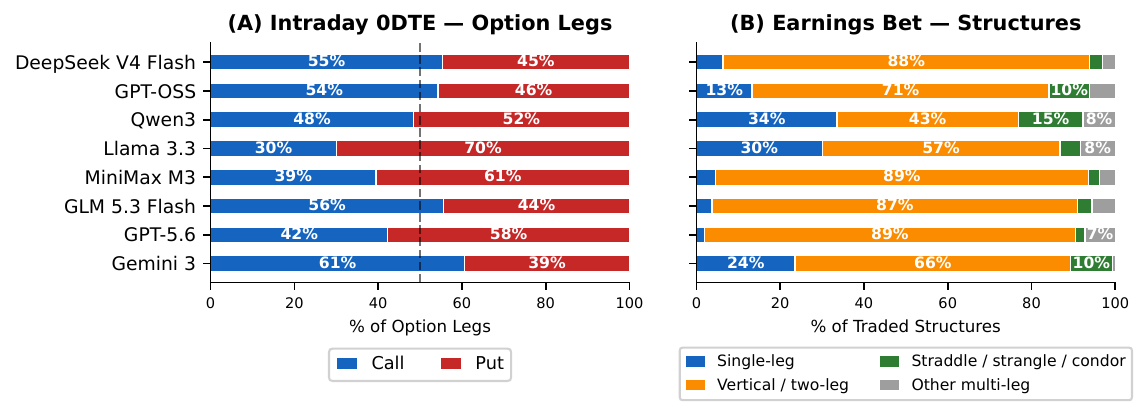}
    \caption{Action composition in the recorded traces: \textbf{(A)} call versus put legs in intraday 0DTE and \textbf{(B)} traded structures in Earnings. }
    \label{fig:action_composition}
    \vspace{-10pt}
\end{figure}

Each agent shows a persistent trading style along two dimensions: how much it trades and which structures it opens. We describe the concentration as recorded; the design does not intervene on training data, so it does not identify a cause. For trading activity (Table~\ref{tab:activity-by-scenario}) we focus on Hedge, Income and 0DTE, as fill counts in Earnings track each model's directional thesis more than its propensity. Activity varies by roughly a factor of two to five between models on the same task, the widest spread being the Income mandate at $164$ to $852$ fills. Structure choice shows a parallel pattern (Figure~\ref{fig:action_composition}), but one concentrated \emph{within} a task rather than carried across tasks: single-leg positions account for $49.2\%$ to $96.1\%$ of classified 0DTE structures, while vertical and two-leg positions account for $43.3\%$ to $89.1\%$ in Earnings. The same agent therefore concentrates on different structures in the two tasks, which is evidence against reading the concentration as a single fixed template.

\begin{table}[h]
    \vspace{-10pt}
  \centering
  \caption{Trading activity by model and scenario, as executed option fills in the selected main runs. The two right-hand columns normalise by episode, per earnings case and per 0DTE session; 0DTE totals are the per-session mean over the 102 selected sessions.}
  \label{tab:activity-by-scenario}
  \setlength{\tabcolsep}{6pt}
  \footnotesize
  \begin{tabular}{@{}l rrrr rr@{}}
    \toprule
    \multirow{2}{*}{\textbf{Model}} & \multicolumn{4}{c}{\textbf{Total Fills}} & \multicolumn{2}{c}{\textbf{Per Episode}} \\
    \cmidrule(lr){2-5} \cmidrule(l){6-7}
    & \textbf{Hedge} & \textbf{Income} & \textbf{Earnings} & \textbf{0DTE} & \textbf{Earnings} & \textbf{0DTE} \\
    \midrule
    DeepSeek-V4-Flash & 511 & 222 & 616 & 3,274 & 6.6 & 32.1 \\
    GPT-OSS-120B & 807 & 852 & 682 & 3,427 & 7.2 & 33.6 \\
    Qwen3-235B-A22B & 703 & 397 & 803 & 2,519 & 8.5 & 24.7 \\
    Llama-3.3-70B-Instruct & 588 & 164 & 678 & 2,397 & 7.5 & 23.5 \\
    MiniMax-M3 & 789 & 405 & 572 & 1,499 & 6.0 & 14.7 \\
    GLM-5.3-Flash & 415 & 458 & 717 & 2,071 & 7.5 & 20.3 \\
    GPT-5.6-Terra & 456 & 546 & 533 & 1,693 & 5.6 & 16.6 \\
    Gemini-3-Flash & 233 & 375 & 468 & 2,009 & 4.9 & 19.7 \\
    \bottomrule
  \end{tabular}
      \vspace{-10pt}
\end{table}

\subsubsection{Episode-Return Distributions and Tail Risk}

Both episodic distributions are right-skewed. Every model loses on the majority of 0DTE sessions, $54$ to $76$ of $102$, and on roughly half of earnings cases, $50$ to $58$ of $90$ to $96$, and every median is negative in both tasks. The left half of Table~\ref{tab:episode_shape_greeks} shows where the means come from. Tail episodes are rare and almost all gains: across both tasks the panel has $38$ upside tails against $5$ downside, and Qwen3-235B-A22B has no 0DTE tail at all. GPT-OSS-120B is the outlier with $10$ upside 0DTE tails and a $90$th percentile of $+9.8\%$, which lifts its mean above zero while its median sits at $-3.20\%$. The positive means in Table~\ref{tab:consolidated_metrics} therefore rest on a handful of sessions rather than on a shifted distribution, so mean return is the wrong summary statistic here and a risk-control mechanism has to be evaluated by intervention instead, as Section~\ref{sec:ablations} does.

  \subsubsection{Where Risk Is Carried: Greek Attribution}%
  \label{appendix:greek}%
  Greek attribution decomposes a realized return into the risk dimensions an agent ended up bearing. It describes carried exposure, not whether that exposure was mispriced. The right half of Table~\ref{tab:episode_shape_greeks} carries the decomposition beside the distribution shape it explains. In Earnings, theta is large and negative for every model, $-625$ to $-3{,}108$ dollars per case over a holding window of $1.5$ to $2.4$ days, the mechanical cost of carrying long premium across the announcement. Vega is small beside it, $+11$ to $+658$ dollars, and delta is positive for seven of eight models, $+239$ to $+1{,}166$, so nominally volatility-oriented positions accumulate directional exposure while held; Gemini-3-Flash is the one exception at $-65$ dollars. The residual dominates and varies most, $-31{,}340$ to $+1{,}685$ dollars per case, which is why first-order attribution should not be read as a profit explanation here.
In 0DTE the same structure appears at a much smaller scale: theta $-98$ to $-271$ dollars per session, vega $+12$ to $+32$, gamma $+76$ to $+211$, and a residual that is negative for every model at $-7$ to $-32$. Delta straddles zero, $-36$ to $+73$, so direction selection is not what separates the models. The attribution is consistent across both tasks and locates where the cost arises without establishing that agents priced it incorrectly. Tables~\ref{tab:intraday_greek_decomp_spy} and~\ref{tab:earnings_fills_greeks} add the case counts, the net PnL and the Vanna and Volga.

\begin{table*}[h]
\centering
\caption{
Episode-level return shape and BSM PnL attribution for 0DTE Intraday (left) and Earnings Bet (right). $\uparrow/\downarrow$ count positive/negative tail episodes, $q_{90}$ is the 90th-percentile return, and attribution reports mean PnL per episode by Greek plus residual. Detail metrics are reported in Tables~\ref{tab:intraday_greek_decomp_spy} and~\ref{tab:earnings_fills_greeks}.
}
\label{tab:episode_shape_greeks}

\setlength{\tabcolsep}{2.8pt}
\renewcommand{\arraystretch}{1.05}
\scriptsize

\begin{tabular}{@{}l rrr rrrrr rrr rrrrr@{}}
\toprule
\multirow{3}{*}{\textbf{Model}}
& \multicolumn{8}{c}{\textbf{0DTE Intraday}}
& \multicolumn{8}{c}{\textbf{Earnings Bet}} \\
\cmidrule(lr){2-9}
\cmidrule(l){10-17}

& \multicolumn{3}{c}{\textbf{Shape}}
& \multicolumn{5}{c}{\textbf{BSM attribution}}
& \multicolumn{3}{c}{\textbf{Shape}}
& \multicolumn{5}{c}{\textbf{BSM attribution}} \\
\cmidrule(lr){2-4}
\cmidrule(lr){5-9}
\cmidrule(lr){10-12}
\cmidrule(l){13-17}

& $\uparrow$ & $\downarrow$ & $q_{90}$
& $\delta$ & $\gamma$ & $\theta$ & $\nu$ & Resid.
& $\uparrow$ & $\downarrow$ & $q_{90}$
& $\delta$ & $\gamma$ & $\theta$ & $\nu$ & Resid. \\
\midrule

DeepSeek-V4-Flash
& 2 & 0 & +3.6
& $-$19 & +123 & $-$163 & +17 & $-$32
& 2 & 0 & +13.5
& +553 & +9659 & $-$1642 & +11 & $-$8737 \\

GPT-OSS-120B
& 10 & 2 & +9.8
& +73 & +211 & $-$271 & +32 & $-$32
& 2 & 0 & +10.6
& +659 & +20157 & $-$1503 & +260 & $-$19642 \\

Qwen3-235B-A22B
& 0 & 0 & +2.3
& $-$9 & +86 & $-$133 & +21 & $-$18
& 3 & 0 & +9.3
& +239 & +32689 & $-$3108 & +658 & $-$31340 \\

Llama-3.3-70B-Instruct
& 3 & 0 & +5.1
& $-$36 & +120 & $-$186 & +14 & $-$16
& 2 & 0 & +11.7
& +750 & +271 & $-$1681 & +202 & +279 \\

MiniMax-M3
& 1 & 0 & +3.1
& +25 & +76 & $-$98 & +16 & $-$7
& 0 & 1 & +8.2
& +257 & $-$864 & $-$1408 & +176 & +1043 \\

GLM-5.3-Flash
& 3 & 0 & +4.3
& +38 & +108 & $-$146 & +25 & $-$16
& 2 & 0 & +9.0
& +1166 & +13988 & $-$1178 & +319 & $-$14463 \\

GPT-5.6-Terra
& 4 & 0 & +3.9
& +1 & +109 & $-$131 & +12 & $-$13
& 0 & 0 & +12.3
& +609 & $-$1465 & $-$625 & +97 & +1685 \\

Gemini-3-Flash
& 1 & 0 & +3.9
& +11 & +93 & $-$121 & +18 & $-$15
& 3 & 2 & +11.4
& $-$65 & +6852 & $-$1973 & +369 & $-$5427 \\

\bottomrule
\end{tabular}

\vspace{-8pt}
\end{table*}

\subsection{Targeted Ablations}
\label{sec:ablations}

The ablation panel comprises $684$ selected runs for DeepSeek-V4-Flash, GPT-OSS-120B and Qwen3-235B-A22B over six profiles, each removing one input or constraint from the standardized agent context and compared against the matched baseline run for the same case, with $24$ paired cases per profile and episodic task. Appendix~\ref{appendix:ablation_detail} carries the per-model cells, intervals and overlay tasks.
Three readings hold. First, no return effect is significant at the $5\%$ level after Holm correction, and the largest are the most outlier-driven: removing portfolio state raises the paired \emph{mean} 0DTE return for all three models while the paired \emph{median} falls for all three. Second, the clear effect is on participation: removing Greeks and multi-timeframe inputs cuts fills per session from $29.5$ to $4.4$ and drives the no-trade rate to $95.8\%$ for DeepSeek-V4-Flash against a baseline of zero. Third, removing portfolio state raises mean maximum drawdown from $4.4\%$ to $12.4\%$, and for every model individually. Context ablation therefore separates agents that outcome comparison cannot. The design principles these results support are set out in Appendix~\ref{appendix:implications}.

\begin{table}[h]
\vspace{-10pt}
\centering
\caption{Targeted context ablations averaged over three models (24 paired cases each). $\Delta$Ret is the paired return change; Fills and MDD show baseline$\to$ablation; NT is the no-trade rate; and $p$ is the minimum Holm-adjusted sign-flip $p$-value. Under \emph{No Greeks}, lower drawdown reflects reduced trading rather than better risk control. Per-model results are in Table~\ref{tab:ablations_by_task}.}
\label{tab:targeted_ablations}
\setlength{\tabcolsep}{5pt}
\footnotesize
\begin{tabular}{@{}l l r r r r r@{}}
\toprule
\textbf{Ablation} & \textbf{Task} & \textbf{$\Delta$Ret (pp)} & \textbf{Fills} & \textbf{NT\%} & \textbf{MDD (\%)} & \textbf{min $p$} \\
\midrule
Fixed Chain & \multirow{3}{*}{0DTE} & $-$0.90 & 29.5 $\to$ 26.0 & 0.0 & 4.4 $\to$ 4.2 & 0.085 \\*
No Portfolio State &  & +2.36 & 29.5 $\to$ 16.1 & 0.0 & 4.4 $\to$ 12.4 & 0.698 \\*
No Task Rules &  & +0.31 & 29.5 $\to$ 30.7 & 0.0 & 4.4 $\to$ 4.5 & 0.211 \\
No News & \multirow{2}{*}{Earnings} & $-$0.31 & 7.1 $\to$ 7.3 & 4.2 & 7.0 $\to$ 8.0 & 1.000 \\*
No IV &  & $-$2.82 & 7.1 $\to$ 7.3 & 5.6 & 7.0 $\to$ 8.3 & 0.059 \\
No Greeks / Multi-TF & 0DTE & +0.25 & 29.5 $\to$ 4.4 & 47.2 & 4.4 $\to$ 1.0 & 0.835 \\
\bottomrule
\end{tabular}
\vspace{-10pt}
\end{table}

\subsection{Implications for Robust and Reliable Agent Design in Option Trading}
\label{appendix:implications}
Based on our findings, we distill three key design principles for future work on building more robust and reliable LLM-based agents in option markets.

\textbf{Execution-layer risk verification is non-negotiable.} Regardless of how well an agent reasons at the analysis stage, all generated orders should pass through a strict backend verification module before execution, enforcing hard constraints on maximum loss per trade, portfolio-level Value-at-Risk, margin feasibility, and gamma exposure. Prompt engineering alone cannot prevent the generation of tail-risk strategies, and in option markets, a single unconstrained position can exceed the entire capital budget. Risk guardrails should therefore be treated as infrastructure.

\textbf{Participation and tail risk belong in the report, next to return.} The measurable effects here are on whether an agent trades at all and on how deep its drawdown runs: no return effect survives correction for multiple comparisons, while removing Greeks pushes one model to no-trade on $95.8\%$ of sessions. Agent reports should carry no-trade and rejection rates and drawdown distributions.

\textbf{Information dependence should be tested by intervention, not inferred from outcomes.} Whether an agent uses the Greeks, the volatility surface or its own portfolio state cannot be read off returns that are statistically indistinguishable across models. Removing each input and re-running the matched case does separate them, and exposes what outcome-only comparison misses.

\section{Related Works}

\textbf{LLMs in Financial Trading.}
Several works have begun to explore the application of LLMs in the domain of trading.
TradingGPT~\citep{li2023tradinggpt} first introduced a systematic LLM-powered trading agent framework,
while FinMem~\citep{yu2025finmem} enhanced agent capabilities through layered memory and character design.
CryptoTrade~\citep{li2024cryptotrade} extended the trading domain to cryptocurrencies by proposing a reflection-based agent system.
Flag-Trader~\citep{xiong2025flag} and FinCon~\citep{yu2024fincon} further incorporated reinforcement learning frameworks to improve agent performance.
However, these works focus primarily on trading assets such as stocks or cryptocurrencies, which exhibit linear payoffs under relatively simple action spaces.
In the options domain, OQL~\citep{luo2026from} is the first to explore LLMs' capability to generate option strategies with specific intent.
Despite this growing interest, the application of LLM-based agents to options trading remains largely underexplored, particularly given the non-linear payoff structures and complex strategy spaces that options present.

\textbf{Benchmarking LLMs in financial Domain} 
With the increasing adoption of LLMs in finance, early benchmarks focused mainly on fundamental tasks like financial document understanding and QA such as FinQA~\citep{chen2021finqa}, FinanceBench~\citep{islam2023financebench}, and CFBenchmark~\citep{lei2023cfbenchmark}. As LLM-powered trading systems emerged, newer benchmarks such as InvestorBench~\citep{li2025investorbench} and StockBench~\citep{chen2025stockbench} began evaluating decision-making in more complex scenarios. However, existing trading benchmarks suffer from prior knowledge leakage and unrealistic assumptions such as ignoring transaction costs~\citep{kong2026evaluating}. Recent work proposes live, dynamic benchmarks that simulate real trading environments~\citep{li2025time,qian2026agents}, but these focus mainly on equities and overlook more complex instruments like options, leaving a critical gap in the evaluation of LLMs for option trading.

\section{Conclusion}

We present LiveOption, a benchmark for evaluating LLM agents in option trading across diverse scenarios, including intraday 0DTE trading, earnings-driven strategies, and portfolio overlay. This provides a structured testbed for agent behavior under nonlinear, constraint-rich decisions.
Our results reveal systematic limitations in current LLM agents: typical episode returns are negative in every task, only the widest earnings gap survives correction for multiple comparisons, and removing portfolio state roughly trebles mean maximum drawdown (Table~\ref{tab:targeted_ablations}). These findings underscore the need for domain-specific benchmarks beyond directional prediction, and for reporting distributions and interventions rather than means alone. We hope LiveOption serves as a foundation for future work on more reliable and transparent trading agents.

\medskip
\newpage
\bibliography{references}
\bibliographystyle{unsrt}

\newpage
\appendix

\section*{Appendix Contents of LiveOption}   %
\startcontents[appendices]
\printcontents[appendices]{l}{1}{} 
\newpage

\section{Broader Impacts}
\label{appendix:board}
LiveOption is an option-focused benchmark for studying the trading behaviour of large language models, extending the evaluation of financial agents beyond conventional stock trading into a domain with leverage, nonlinear payoffs, and structured strategy design. Beyond the evaluation infrastructure itself, our work documents several behavioural patterns that outcome-only evaluation does not reveal: tenor policy varies widely across agents, with median entry maturities from $2.5$ to $28$ days against prescribed bands of $7$--$45$ and $7$--$60$ days, and only $20.2\%$ of overlay fills shorter than a week; action structure is concentrated within each task but differs between tasks; risk rejection rates span $7.0\%$ to $66.1\%$ in 0DTE; and removing portfolio state roughly doubles to quintuples mean maximum drawdown. From these observations we distil design principles that the evidence supports---treating execution-layer risk verification as infrastructure rather than prompt-level guidance, reporting participation and tail risk alongside return, and testing information dependence by intervention rather than inferring it from outcomes. We hope these contributions support future research on more reliable and transparent AI systems for finance by making model strengths and limitations easier to study and providing a concrete diagnostic vocabulary for option-aware agent design.

\section{Limitations}
Our evaluation has several limitations. (1) Each model--case configuration is represented by a single selected rollout, so our uncertainty estimates primarily reflect variation across market episodes rather than repeated-generation variability; moreover, the analysis conditions on valid completed runs and therefore does not measure end-to-end API reliability. (2) The evaluation universe remains limited: Overlay contains only five assets per model and mandate, while the broader benchmark covers 2025 data, SPY 0DTE sessions, five large-cap equities for Overlay, and 24 single names for Earnings. Results should therefore not be extrapolated to other sectors, asset classes, or market regimes without further evaluation. (3) Execution costs are modeled in detail only for 0DTE, where quote coverage permits spread- and depth-based analysis; Overlay and Earnings lack comparable quote coverage, and our fixed-decision re-pricing does not capture behavioral adaptation to latency, queue priority, or market impact. (4) Computational constraints limit our model and agent coverage: smaller open-weight models and alternative architectures, such as multi-agent, tool-augmented, or specialist agents, remain unexplored.

\section{Preliminary}
\label{sec:preliminary}

\subsection{Introduction of Option}
\label{sec:prelim_option_intro}

An \textbf{option} is a derivative contract that gives its holder the \textbf{right, but not the obligation}, to trade an underlying asset (e.g., a stock or ETF) at a pre-specified price within a pre-specified time window. We denote the underlying price by $S_t$, the pre-specified \textbf{strike price} by $K$, and the expiration date by $T$. Options mainly come in two forms. A \textbf{call} grants the right to buy the underlying at $K$, while a \textbf{put} grants the right to sell the underlying at $K$. From the position side, an investor can go \textbf{long} an option by paying a premium to buy it, or \textbf{short} (write) an option by selling it for a premium in exchange for the obligation to fulfill the contract if exercised.

The option price can be understood as the sum of \textbf{intrinsic value} and \textbf{time value}. Intuitively, intrinsic value is the immediate exercise value: for a call it is $\max(S_t - K, 0)$, and for a put it is $\max(K - S_t, 0)$. Time value is the extra premium beyond intrinsic value, reflecting uncertainty of future movements before expiration and the remaining time to $T$. This decomposition is helpful because many trading behaviors (e.g., holding until expiration versus exiting earlier) can be explained through the fast decay of time value, especially for short-maturity contracts.

\subsection{Option Pricing Model}
\label{sec:prelim_pricing}
 
Option pricing establishes the relationship between an option's fair value and the dynamics of its underlying asset.
The standard analytical baseline is the \textbf{Black--Scholes--Merton (BSM)} framework~\citep{black1973pricing}, which models the underlying price as a geometric Brownian motion:
\begin{equation}
dS_t = \mu\, S_t\, dt + \sigma\, S_t\, dW_t,
\end{equation}
where $\mu$ is the drift rate, $\sigma$ is the constant volatility, and $W_t$ is a standard Brownian motion.
Under no-arbitrage and frictionless market assumptions, the fair value of a European call option with strike $K$ and maturity $T$ admits the closed-form solution:
\begin{equation}
C = S_0\, \Phi(d_1) - K\, e^{-rT}\Phi(d_2),
\end{equation}
with the corresponding put price given by put--call parity as:
\begin{equation}
P = K\, e^{-rT}\Phi(-d_2) - S_0\, \Phi(-d_1),
\end{equation}
where $r$ is the risk-free rate, $\Phi(\cdot)$ denotes the standard normal CDF, and
\begin{equation}
d_1 = \frac{\ln(S_0/K) + (r + \tfrac{1}{2}\sigma^2)\,T}{\sigma\sqrt{T}}, \qquad
d_2 = d_1 - \sigma\sqrt{T}.
\label{eq:bsm}
\end{equation}
 
Although real markets violate several BSM assumptions---most notably through the volatility smile, jump dynamics, and transaction costs---the framework provides a tractable reference model and, more importantly, motivates the key price sensitivities known as \emph{Greeks} (Delta, Gamma, Theta, Vega, etc.) that underpin modern option risk management.
 
In practice, option prices are commonly quoted in terms of \textbf{implied volatility} (IV): given an observed market price, one inverts the BSM formula to recover the $\sigma$ that reproduces that price. IV provides a normalized scale for comparing options across different strikes and expiries, and is a key determinant of strategy performance---separating, for example, short-volatility premium harvesting from long-volatility event speculation.

\subsection{Greeks of Options}
\label{sec:prelim_greeks}
 
The partial derivatives of the option value $V$ with respect to its pricing parameters are collectively known as the \emph{Greeks}.
They quantify how sensitive an option position is to changes in each underlying risk factor, and form the basis for hedging, risk management, and---as we develop in Section~\ref{sec:pnl_decomposition}---PnL attribution.
 
Under the BSM framework with $d_1$ and $d_2$ defined in Eq.~\eqref{eq:bsm}, the Greeks for a European call option are given in closed form as follows. Table~\ref{tab:greeks_summary} summarizes the Greeks, their risk interpretations, and the trading scenarios in our benchmark where each is most relevant.
 
\textbf{First-order Greeks.}
First-order Greeks are the direct partial derivatives of the option value with respect to each individual risk factor, describing how the option price responds to a small, isolated change in one variable while holding all others constant.

\begin{itemize}[leftmargin=12pt]
    \item \textit{Delta} ($\delta = \partial V / \partial S = N(d_1)$) measures the sensitivity of option value to changes in the underlying price and is the primary determinant of directional exposure.
\item \textit{Theta} ($\theta = -\partial V / \partial T = -\frac{S\,N'(d_1)\,\sigma}{2\sqrt{T}} - r\,K\,e^{-rT}N(d_2)$) quantifies time decay: long option positions lose value as expiration approaches, making theta the cost of holding optionality.
\item \textit{Vega} ($\nu = \partial V / \partial \sigma = S\sqrt{T}\,N'(d_1)$) captures sensitivity to implied volatility; it is central to volatility trading strategies and is typically largest for at-the-money options.
\item \textit{Rho} ($\rho = \partial V / \partial r = K\,T\,e^{-rT}N(d_2)$) measures sensitivity to the risk-free rate. Since $\rho$ scales with time to maturity $T$ and interest rate changes $\Delta r$ are typically small over short holding periods, its contribution is negligible for short-dated options and is omitted from our analysis unless otherwise noted.
\end{itemize}

\textbf{Second-order Greeks.}
Second-order Greeks are the second partial derivatives of the option value, capturing how the first-order sensitivities themselves change as market conditions evolve. \textit{Gamma} ($\gamma = \partial^2 V / \partial S^2 = N'(d_1) / (S\,\sigma\sqrt{T})$) is the rate of change of delta with respect to the underlying price.
High gamma means delta shifts rapidly as the underlying moves, which is both an opportunity (gamma scalping) and a risk (hedging instability).
Gamma is most pronounced for at-the-money options near expiration.

\begin{table}[h]
\centering
\caption{Summary of option Greeks and their relevance to LiveOption evaluation scenarios.}
\label{tab:greeks_summary}
\small
\begin{tabular}{@{}llll@{}}
\toprule
\textbf{Greek} & \textbf{Definition} & \textbf{Risk Dimension} & \textbf{Key Scenario(s)} \\
\midrule
Delta ($\delta$)  & $\partial V / \partial S$                        & Directional exposure     & Overlay               \\
Gamma ($\gamma$)  & $\partial^2 V / \partial S^2$                    & Price convexity          & 0DTE                  \\
Theta ($\theta$)  & $-\partial V / \partial T$                       & Time decay               & 0DTE, Overlay \\
Vega ($\nu$)      & $\partial V / \partial \sigma$                   & Volatility level         & Earnings Bet \\
Rho ($\rho$)      & $\partial V / \partial r$                        & Interest rate            & (negligible in short-dated) \\
\bottomrule
\end{tabular}
\end{table}

\subsection{Option Trading Rules}
\label{sec:prelim_rules}

The exchange standardizes option contracts. In equity options, each contract represents a fixed number of shares of the underlying (commonly 100), giving each contract leveraged exposure compared to trading the underlying directly. Options are identified by a tuple $(\text{underlying}, K, T, \omega)$, where $\omega \in \{\text{call}, \text{put}\}$. Traders place orders on these contracts, and positions are marked to market as option prices change.

A key mechanism is \textbf{exercise} and \textbf{assignment}. Exercising a call gives the holder the right to buy the underlying at $K$; exercising a put gives the right to sell at $K$. For the writer (short side), assignment means being obligated to sell the underlying (if short a call) or buy it (if short a put) at the strike. American-style options can be exercised any time before expiration, while European-style options can only be exercised at expiration. In practice, most retail and institutional traders close their option positions before expiration rather than exercising. However, assignment risk still matters for short positions, particularly at expiration when the underlying settles near the strike (pin risk), and before ex-dividend dates for short ITM calls on dividend-paying stocks.

Trading also faces practical constraints, including margin requirements for short positions, position limits, and liquidity conditions. Short options typically require higher margin because of their asymmetric payoff --- large potential losses against a capped premium gain --- while long options have losses bounded by the premium paid. These rules shape which strategies are feasible under different risk budgets, and they explain why many agents prefer spreads (bounded risk) to naked short options.

\begin{table}[h]
  \caption{Common option strategies and their constructions. Payoff intuition is described at expiration $T$; profit equals payoff minus net premium, and short legs flip the corresponding payoff.}
  \label{tab:option_strategies}
  \centering
  \small
  \begin{tabularx}{\textwidth}{@{}l l X@{}} %
    \toprule
    \textbf{Name} & \textbf{Construction} & \textbf{Payoff / Profit Intuition (at $T$)} \\
    \midrule
    Long Call & Buy 1 call $(K,T)$, pay $c$ & Profit $= (S_T-K)^+ - c$; bullish with limited loss and uncapped upside. \\
    \addlinespace[0.5em]
    Long Put & Buy 1 put $(K,T)$, pay $p$ & Profit $= (K-S_T)^+ - p$; bearish protection with limited loss and downside convexity. \\
    \addlinespace[0.5em]
    Covered Call & Long stock + sell 1 call $(K,T)$ & Collects premium; upside capped above $K$, downside partially cushioned by premium. \\
    \addlinespace[0.5em]
    Protective Put & Long stock + buy 1 put $(K,T)$ & Downside bounded (insurance-like); sets a floor near $K$ on portfolio value. \\
    \addlinespace[0.5em]
    Cash-Sec. Put & Sell 1 put $(K,T)$ (cash-backed) & Receives premium; profits if $S_T > K$; risk of buying stock at $K$ on drops. \\
    \addlinespace[0.5em]
    Bull Call Spread & Buy call $K_1$, sell call $K_2$ ($K_1 < K_2$) & Moderately bullish; bounded profit/loss; reduces cost but caps upside. \\
    \addlinespace[0.5em]
    Bear Put Spread & Buy put $K_2$, sell put $K_1$ ($K_1 < K_2$) & Moderately bearish; bounded profit/loss; cheaper than long put but capped. \\
    \addlinespace[0.5em]
    Long Straddle & Buy call $(K,T)$ + buy put $(K,T)$ & Direction-agnostic; profits from volatility; max loss is premium paid. \\
    \addlinespace[0.5em]
    Long Strangle & Buy OTM call $K_2$ + OTM put $K_1$ & Cheaper than straddle; requires larger move; direction-agnostic. \\
    \bottomrule
  \end{tabularx}
\end{table}

\subsection{Profit and Strategy}
\label{sec:prelim_strategies}

\textbf{Profit (payoff) structure.} Option trading can be understood from its payoff at expiration. Let $S_T$ be the underlying price at maturity $T$ and $K$ be the strike. The payoff of a long call is $(S_T-K)^+$, while a long put is $(K-S_T)^+$, where $(x)^+=\max(x,0)$. The \emph{profit} further subtracts the entry premium: for a long call the profit is $(S_T-K)^+ - c$, and for a long put it is $(K-S_T)^+ - p$, where $c$ and $p$ are the paid premiums. In contrast, short (written) options invert the profit profile: the trader receives premium up front but takes the obligation at exercise, leading to limited upside and potentially large losses. This asymmetric profit profile is the main reason why options are widely used for hedging, income generation, and event-driven trading.

\textbf{From payoff to Option Strategy.} An \textbf{option strategy} is a structured combination of options (and sometimes the underlying) designed to shape the final profit curve. Practically, strategies can be viewed as engineering a desired trade-off among direction exposure and risk sensitivities (e.g., delta, gamma, theta, and vega), but for reviewers new to options, it is often easiest to interpret a strategy by (i) what instruments it combines, and (ii) the qualitative payoff shape it creates. For example, a protective put behaves like insurance for an existing long stock position by paying a premium to cap downside losses, while a covered call sells upside beyond a strike to collect premium and enhance yield. Similarly, spreads combine multiple option legs to bound risk and cost, and straddles/strangles express a view on volatility (large vs.\ small moves) rather than a strict direction. Common option strategies and their payoff are introduced in Table~\ref{tab:option_strategies}.

\subsection{Option PnL Decomposition via Greeks}
\label{sec:pnl_decomposition}

A central challenge in evaluating option trading agents is that raw profit and loss (PnL) conflates multiple independent risk dimensions: an agent may profit purely from a favorable directional move while its volatility positioning and time management are structurally flawed.
To disentangle these effects, we decompose realized PnL into contributions from each Greek using a Taylor expansion of the option value $V$ with respect to its underlying risk factors.

\paragraph{Derivation.}
Let $S$ denote the underlying price, $\sigma$ the implied volatility of the specific contract, $r$ the risk-free interest rate, and $t$ calendar time.
Between two consecutive evaluation steps, the change in option value can be approximated as:
\begin{equation}
\label{eq:pnl_taylor}
\Delta \text{PnL} \;\approx\;
\underbrace{\delta \cdot \Delta S}_{\text{Delta PnL}}
\;+\; \underbrace{\tfrac{1}{2}\,\gamma \cdot (\Delta S)^2}_{\text{Gamma PnL}}
\;+\; \underbrace{\theta \cdot \Delta t}_{\text{Theta PnL}}
\;+\; \underbrace{\nu \cdot \Delta\sigma}_{\text{Vega PnL}}
\;+\; \underbrace{\rho \cdot \Delta r}_{\text{Rho PnL}}
\;+\; \underbrace{\varepsilon}_{\text{Unexplained}},
\end{equation}
where $\Delta S = S_{t+1} - S_t$, $\Delta\sigma = \sigma_{t+1} - \sigma_t$, $\Delta r = r_{t+1} - r_t$, and $\Delta t$ is the elapsed calendar time (in years).
The five named Greek terms arise from the first- and second-order partial derivatives of $V$ with respect to the underlying risk factors $(S, \sigma, r, t)$, and $\varepsilon$ captures higher-order residuals including cross-terms such as Vanna ($\partial^2 V / \partial S\,\partial\sigma$) and Volga ($\partial^2 V / \partial\sigma^2$).

\paragraph{Unexplained Residual.}
The residual $\varepsilon$ in Eq.~\eqref{eq:pnl_taylor} absorbs higher-order and cross-derivative terms, including Volga ($\partial^2 V / \partial\sigma^2$), Vanna ($\partial^2 V / \partial S\,\partial\sigma$), Charm ($\partial\delta / \partial t$), and Speed ($\partial\gamma / \partial S$), as well as discrete rebalancing error from finite time steps and any model mismatch between Black--Scholes Greeks and actual market dynamics.
A persistently large $|\varepsilon|$ relative to the explained components signals either that the evaluation time step is too coarse, or that the position is in a regime where the first-order expansion is insufficient (e.g., near-expiry options with extreme gamma, or large volatility jumps around earnings).

\paragraph{Application to Evaluation.}
The decomposition in Eq.~\eqref{eq:pnl_taylor} is computed at every time step of the backtesting engine for each open position, then aggregated across legs for multi-leg strategies.
This enables scenario-specific diagnostic metrics:
in \textit{Earnings Bet}, we expect vega PnL to dominate around announcements (implied volatility crush or expansion);
in \textit{0DTE}, theta and gamma should be the primary drivers;
in \textit{Overlay}, delta PnL attribution reveals hedging efficiency.
By inspecting the decomposition, we can assess not just \emph{whether} an agent profited, but \emph{why}, distinguishing genuine option-aware reasoning from incidental directional gains.

\section{Data Preparation and Framework Design}
\label{appendix:data}

\subsection{Dataset Preparation Pipeline}
\label{appendix:data_pre}

LiveOption uses the data and news pipelines provided by QFinZero~\cite{luo-etal-2026-qfinzero}. The reference pipeline provides native ingestion support for market data from commercial third-party providers, including Massive and Databento; Benzinga is the native source for financial news. These sources are accessed through QFinZero's standardized interfaces rather than through provider-specific calls in the trading agents.
Users can also import their own market data and news corpora by converting them to the documented schemas. Imported news records retain their publication timestamps for time-aligned retrieval during replay. This allows researchers to run the same framework with their own data assets; substituting data sources does not imply reproducing identical numerical results. The preparation pipeline proceeds as follows:

\begin{enumerate}[leftmargin=12pt]
    \item Users obtain the required option data, underlying price histories, news records, and event metadata from sources for which they have the necessary access and usage rights.
    \item QFinZero's ingestion and preprocessing pipeline converts provider data or user-imported market and news records into the documented schemas and aligns them for decision-time queries.
    \item The deployed QFinZero services expose standardized REST interfaces queried by LiveOption's backtesting engine and agent modules, decoupling data access from experiment execution.
\end{enumerate}

The open-source release provides the framework, preprocessing code, schemas, and experiment configurations. Access to and redistribution of source data remain subject to the applicable provider licenses and usage rights; restricted source data are not included in the public release.

\begin{figure}[h]
    \centering
    \includegraphics[width=0.99\linewidth]{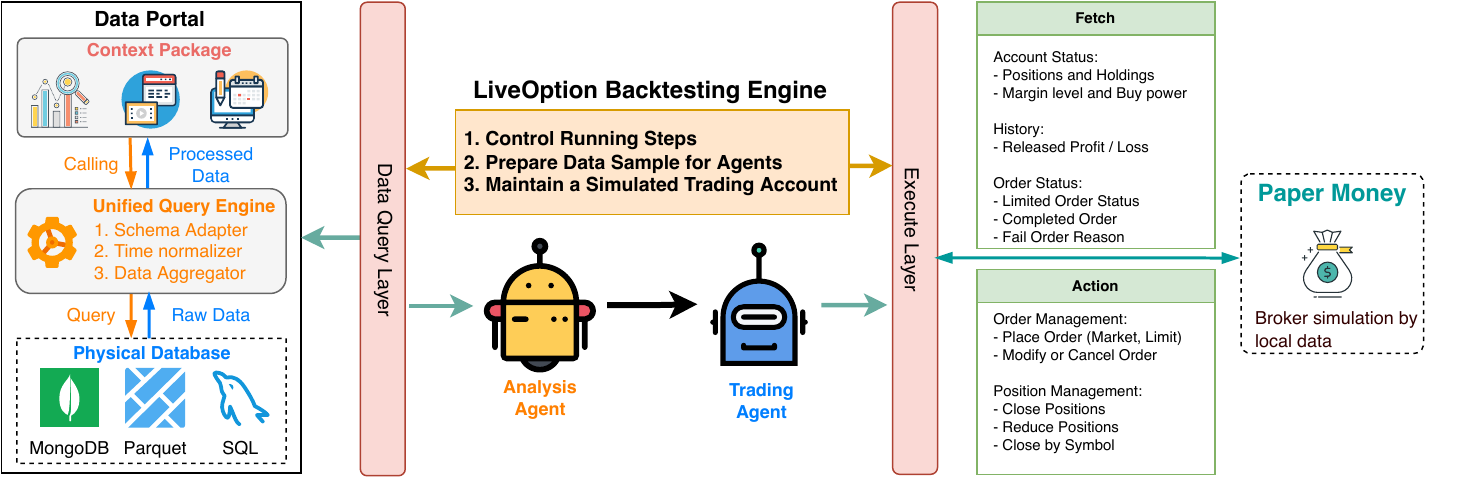}
    \caption{Overview of the LiveOption backtesting engine. The system integrates a unified multi-source data portal, a time-aware query layer, and a simulated brokerage execution environment. At each step, the engine prepares decision-time market context for the analysis and trading agents while maintaining a paper-money account for realistic order execution, position management.}
    \label{fig:bt_engine}
\end{figure}

\subsection{Design of Framework}

The LiveOption framework consists of three main components: \textbf{(1) Data Portal, (2) Paper Money Simulator, and (3) Backtesting Engine}. Together, these modules provide a realistic and reproducible environment for evaluating autonomous option trading agents.

\begin{enumerate}[leftmargin=12pt]
    \item \textbf{Data Portal.} LiveOption accesses the resources described in Appendix~\ref{appendix:data_pre} through QFinZero's data and news services. The portal organizes provider-sourced and user-imported records into unified context packages for agent analysis and trading decisions. The supported data include both \textit{unstructured information}, such as financial news and event reports, and \textit{structured market data}, such as underlying asset prices, volatility indicators, and option chain snapshots. Through schema normalization, time alignment, and efficient querying, the portal enables agents to access multi-modal financial information in a consistent format.
    
\item \textbf{Paper Money Simulator.} A simulated brokerage. It exposes account queries and order operations (market and limit orders, modification, cancellation, closing) and maintains cash, buying power, margin usage, realized and unrealized PnL and open positions. It also models the option-specific lifecycle: expiration, assignment, exercise and multi-leg position management.
\item \textbf{Backtesting Engine.} The engine drives the experiment. It steps each run against the trading calendar and termination conditions; at every step it assembles the agent's context from news, price history and the option chain, restricted to information timestamped before that step so that future data cannot leak into a decision; and it synchronizes with the simulator to execute actions, update the portfolio and record the logs the evaluation reads.
\end{enumerate}

\subsection{Web Interface}
\label{appendix:web}

We provide a web-based interface that enables users to configure and run experiments conveniently. Users can specify key experimental settings, including the backtesting date range, selected base model, account configurations, and scenario-specific parameters.
During execution, the interface allows users to monitor the full backtesting process in real time, including the contextual information provided to the agent, actions taken by the agent, and changes in account value at each decision step.
After backtesting is complete, users can visualize several important trading metrics, including portfolio returns, drawdown, win rate, position exposure, and profit/loss statistics.

The interface reads the same artifacts the benchmark writes, rather than a
separate store, so what it shows is exactly what the reported analyses consume.
Table~\ref{tab:web_mapping} maps each view to the fields behind it. For a
finished run it replays the recorded artifacts; during an active run it renders
the same records as they are produced.

\begin{table}[h]
\centering
\small
\caption{Web interface views and the recorded fields each one reads.}
\label{tab:web_mapping}
\begin{tabular}{@{}p{0.30\linewidth} p{0.64\linewidth}@{}}
\toprule
\textbf{View} & \textbf{Source fields} \\
\midrule
Market path and chain & Underlying bars and option-chain snapshots from the query layer, at the decision timestamp \\
News and event triggers & Time-aligned news and calendar entries from the event layer \\
Agent context and rationale & The observation packet and thesis text recorded per decision step \\
Orders and fills & Submitted intents, verification outcome, and the resulting fills with price, quantity and fees \\
Portfolio and risk & Cash, equity, open positions, margin usage and per-position Greeks from the portfolio record \\
Performance panels & Episode return, drawdown, win rate and exposure, recomputed from the portfolio series \\
\bottomrule
\end{tabular}
\end{table}

In future versions, we plan to extend the platform further to support fully customized testing scenarios, including user-defined context inputs, custom prompts, and flexible asset universes.


\section{Scenarios Design}
\label{appendix:scenarios_design}

\subsection{Takeaway}

Our three core design scenarios span from long-term portfolio management to ultra-short-term execution, and we summarize the main settings of experiments for each scenario in Table~\ref{tab:scenario_settings}. The Portfolio Overlay scenario demonstrates how systematic option selling can generate extra yield or provide tail-risk mitigation through protective hedging to reshape core equity distributions. Moving to event-driven dynamics, the Earnings Bet scenario outlines a three-phase timeline---Betting, Reaction, and Aftershock---designed to capture volatility expansion and post-event price drifts. Finally, the 0DTE Intraday scenario focuses on high-frequency execution within a single session, managing rapid theta decay and gamma risk from early-session positioning through to the final expiration-horizon close-out. Together, these scenarios provide a comprehensive benchmark for evaluating option strategies across varying time horizons and volatility regimes.

\begin{table}[h]
    \caption{Implementation and experimental settings across the three scenarios.}
    \label{tab:scenario_settings}
    \centering
    \small
    \renewcommand{\arraystretch}{1.15}
    \begin{tabular}{@{}p{0.18\linewidth} p{0.27\linewidth} p{0.23\linewidth} p{0.23\linewidth}@{}}
        \toprule
        \textbf{Setting} & \textbf{Overlay} & \textbf{Earnings Bet} & \textbf{0DTE Intraday} \\
        \midrule
        Universe
            & QQQ, NVDA, AMD, GOOG, TSLA
            & 24 single names across mega-cap tech, semis, software, consumer, fintech (Tab.~\ref{tab:earnings_bet_world})
            & SPY \\
        Evaluation horizon
            & Full-year 2025, weekly cadence
            & 2025 fiscal-year scheduled releases (95 event cases, 90--95 valid per model)
            & Wednesdays \& Fridays of 2025 (102 sessions) \\
        Bar frequency
            & 1-day
            & 1-min (intraday) + daily close gates
            & 1-min \\
        Decision frequency
            & Weekly rebalance (Monday close)
            & Daily close at $T{-}2$, $T{-}1$; every 30~min on $T$ to $T{+}2$
            & Every 15~min, 09:30--16:00~ET \\
        Initial capital
            & 10{,}000 shares per name + 20\% cash buffer
            & \$50{,}000 cash (event-betting account)
            & \$100{,}000 cash \\
        Allowed structures
            & \textit{Income:} covered call, cash-secured put. \textit{Hedge:} long put, bear put spread. No naked premium selling.
            & Long call/put, vertical call/put spreads, straddles, strangles. No naked short legs.
            & Long call/put, vertical call/put spreads, straddles, strangles. \\
        DTE constraint
            & \textit{Income:} 7--45~d. \textit{Hedge:} 7--60~d.
            & 7--14~d (weekly/bi-weekly cycle around the event)
            & 0--7~d, $\delta\!\in\![0.20,0.60]$ \\
        Position-sizing limits
            & Effective $|\Delta|$ capped at the initial stock $\Delta$; \textit{Hedge:} weekly premium budget
            & $\le 50$ contracts/trade; $\le \$30$k cost/trade; 40\% daily-loss circuit breaker
            & $\le 5$ concurrent positions; $\le 50\%$ equity exposure; 2\% daily-loss flatten, 3\% alert \\
        Exit rules
            & Agent-driven roll/close; auto-close ITM legs at expiry
            & Force-flat at $T{+}2$ EOD; no overnight carry beyond $T{+}2$
            & Force-flat at 15:55~ET; no overnight carry \\
        Context fed to agent
            & Spot, mark, full Greeks ($\Delta,\Gamma,\Theta,\mathcal{V}$), IV, DTE, position table, compressed chain (IV surface, moneyness buckets, top-tradable contracts)
            & All of Overlay's context, plus historical earnings reactions (gap\%, post-move\%), EPS/revenue estimates, last-5-day news sentiment, pre-/post-event chain snapshots
            & All of Overlay's context, plus aggregated 15-min OHLCV bars, minutes-to-close countdown, stale-quote flags \\
        \midrule
        \multicolumn{4}{l}{\textit{Trading frictions (shared simulation engine)}} \\
        Option commission
            & \multicolumn{3}{l}{\$0.65 per contract} \\
        Stock commission
            & \multicolumn{3}{l}{\$0.0005 per share} \\
        Slippage
            & \multicolumn{3}{l}{1.0~bp applied to fill price} \\
        Bid--ask spread
            & \multicolumn{3}{l}{Mid-quote fills (0~bp explicit spread cost)} \\
        \bottomrule
    \end{tabular}
\end{table}

\subsection{Overlay Strategy}
\label{appendix:sc_overlay}

\subsubsection{Background}

Option overlay strategies are among the most widely adopted applications of derivatives in institutional portfolio management. The defining characteristic of an overlay is that the core equity holding remains fixed; rather than liquidating or rebalancing the underlying position, the manager systematically layers option structures on top of it to reshape the portfolio's payoff distribution. This approach is attractive precisely because it decouples the equity allocation decision from the risk and income management decision, allowing both objectives to be pursued independently.

In practice, overlay strategies serve two complementary economic objectives. The first is income enhancement, in which the manager sells option premiums to generate supplemental yield on an existing holding, most commonly through covered calls or cash-secured puts. The second is downside hedging, in which the manager purchases protective structures, such as long puts or put spreads, to truncate left-tail exposure without liquidating the position. Although these two objectives involve different risk profiles, they share the same structural premise: a fixed equity base, a defined cash budget, and a periodic decision process governing contract selection, strike choice, and maturity. We therefore evaluate them under a unified overlay framework, treating income enhancement and downside hedging as two subtasks of the same problem class.

\subsubsection{Settings}

For each underlying, every run begins with a fixed long-equity base of $10{,}000$ shares plus a cash buffer equal to $20\%$ of the initial stock notional value. The equity base is held constant throughout the evaluation period; agents may not liquidate or trade the underlying directly. The cash buffer serves a dual purpose: in the income subtask, it collateralizes short put positions and absorbs assignment, while in the hedging subtask, it funds the periodic purchase of protective structures. Identical capitalization is used across the two subtasks to ensure that differences in available capital do not confound cross-task comparisons.

Agents make rebalancing decisions weekly, executed at Monday's close. At every decision point, the agent receives: (i) the current spot price, mark prices, and full Greeks ($\Delta, \Gamma, \Theta, \mathcal{V}$) along with implied volatility for all open positions; (ii) a position table containing unrealized PnL, entry price, DTE, and spread-group labels; (iii) a compressed snapshot of the option chain summarizing the implied-volatility surface, moneyness buckets, and a short list of liquid contracts within the admissible DTE band; and (iv) recent stock-level price action over the trailing window. All option legs are restricted to short- to medium-term expiries: $7$--$45$ days for the income subtask and $7$--$60$ days for the hedging subtask.

In the income subtask, the agent decides when to enter short option positions, which strike and maturity to select, and which premium-generating structure to apply. The strategy space is restricted to two structures: covered calls, where short calls are secured by the existing equity holding, and cash-secured puts, where short puts are collateralized by the cash buffer. Naked (uncollateralized) short positions are disallowed. The portfolio's net delta is constrained to not exceed, in magnitude, the delta of the unencumbered stock position, preventing the agent from converting an income overlay into a leveraged directional bet. The objective is to enhance portfolio income while maintaining controlled risk exposure relative to the buy-and-hold benchmark.

In the hedging subtask, by contrast, the agent operates as a risk manager rather than a premium seller. The strategy space is restricted to long protective structures: long puts and bear put spreads, both of which truncate the lower tail of the portfolio's payoff distribution by providing explicit downside protection at a known maximum cost. Cash from the buffer may only be used to purchase protective structures, and net short-vega exposure is disallowed. The objective is to mitigate downside risk while controlling hedging cost under a constrained premium budget.

\subsubsection{Experiment}

Each (model, subtask) pair is evaluated over the full 2025 trading year ($250$ trading days) on each of the five underlyings, yielding $5 \times 2 = 10$ runs per model. The five-name panel and the two subtasks are treated as independent runs rather than a single multi-asset portfolio, so that per-name attribution and dispersion across regimes can be measured directly. Every run is initialized with the same fixed equity base, cash buffer, and admissible strategy space described above; the only quantities that vary across runs are the underlying ticker and the subtask label.

\subsection{Earnings Bet}
\label{appendix:sc_earnings}

\subsubsection{Background}

Earnings announcements are among the most consequential event-driven catalysts in equity markets. Because scheduled announcements introduce discrete, time-concentrated uncertainty, they routinely trigger large price jumps and sharp volatility spikes upon release. This combination of predictable timing and unpredictable magnitude makes earnings events a natural laboratory for event-driven option trading, and they have long been a focus of both practitioner strategies and academic study.

Earnings trading presents several structural challenges that distinguish it from conventional directional trading. In the period leading up to an announcement, implied volatility typically rises as market participants price in uncertainty about the forthcoming results. Once the announcement is released, this uncertainty resolves and implied volatility often collapses sharply, a phenomenon widely known as the \textit{IV crush}. This volatility contraction directly affects option pricing and can substantially erode or reverse gains from an otherwise correct directional prediction. Furthermore, traders commonly use short-dated contracts around earnings events to maximize leverage on the anticipated move, but the accelerated time decay and sensitivity to volatility changes at short maturities reduce the effective win rate of simple directional positions. Successful earnings trading, therefore, demands more than a directional view: the agent must simultaneously reason about expected price movement, anticipate volatility dynamics, and manage position risk across a rapidly evolving information environment.

\subsubsection{Settings}

An idealized earnings trade unfolds across three distinct phases. The first is the \textit{betting phase}, occurring one to several trading days before the announcement. During this window, the agent forms expectations about the earnings outcome based on available contextual signals, including historical earnings reactions, recent price trends, and implied volatility levels embedded in the option chain, and initiates a position accordingly. The second is the \textit{reaction phase}, occurring at and immediately after the announcement. The market processes the new information, often producing a large directional move, and the agent must decide in near real time whether to take profits, cut losses, or adjust the position in response to the realized price and volatility shock. The third is the \textit{aftershock phase}, covering the short post-event window, during which prices may continue to drift or partially reverse as the initial reaction settles and implied volatility normalizes. The agent must determine whether to exit or continue holding through this residual period.

The strategy space spans the core option structures used in practice for earnings positioning. For directional bets, the agent may use long calls or long puts to express a view on the direction of the post-announcement move. To manage the cost of premiums and reduce sensitivity to IV crush, the agent may deploy vertical spreads, either call spreads or put spreads, which partially offset the long premium with a short leg at a further strike. For volatility-oriented positions that are agnostic to direction, the agent may construct straddles or strangles, which profit from a sufficiently large move in either direction regardless of the sign. The choice among these structures requires the agent to balance directional conviction, premium cost, and exposure to volatility compression, capturing the core decision-making complexity of earnings trading.

\begin{table}[h]
\vspace{-12pt}
\caption{Tradable universe in the Earnings Bet scenario.}
\label{tab:earnings_bet_world}
\centering
\small
\begin{tabularx}{\linewidth}{l X X}
\toprule
Category & Assets & Description \\
\midrule

Mega-Cap Tech
& AAPL, MSFT, TSLA, AMZN, META, NVDA
& High liquidity; large absolute post-earnings moves \\

\addlinespace
Semiconductor
& AMD, INTC, MU, AVGO, SMCI
& Cyclical earnings sensitivity and hardware-sector volatility \\

\addlinespace
Consumer \& Streaming
& NFLX, DIS, NKE, CELH
& Subscriber, brand demand, and consumer spending reactions \\

\addlinespace
Cloud / SaaS
& CRM, SNOW, PLTR, SHOP
& Revenue-growth sensitive; wide analyst forecast dispersion \\

\addlinespace
Fintech \& Crypto-Adjacent
& HOOD, COIN, SOFI, PYPL
& Highly volatile around earnings; retail-flow driven \\

\addlinespace
High Beta / Growth
& RKLB
& Elevated volatility and strong reaction to earnings surprises \\

\bottomrule
\end{tabularx}
\end{table}

\subsubsection{Experiment}

We select the companies listed in Table~\ref{tab:earnings_bet_world} and collect their scheduled earnings announcements over 2025 to construct 95 independent test cases, of which 90 to 95 are valid per model. Each test case corresponds to a single earnings event and is fully self-contained, providing the agent with the company's historical earnings reactions and associated price movements, recent price trends leading into the announcement, and the available option chain at the time of position initiation.

Each trading session begins two trading days before the earnings announcement, denoted $T-2$. The agent receives all contextual information at this point and may choose to open a position at the market close of $T-2$ or defer entry to $T-1$. Following the announcement, the environment transitions into an active management phase in which the agent may exit, hold, or adjust its position in response to realized market conditions. The session concludes once the agent voluntarily closes the position, or at a maximum horizon of $T+2$, at which point all remaining positions are automatically closed and the final portfolio outcome is recorded.
Performance is evaluated on realized profit and loss across the full set of test cases, with additional diagnostic metrics examining the agent's ability to anticipate IV crush, construct appropriately structured strategies relative to its directional conviction, and execute timely exits across the three trading phases.

\subsection{0DTE Intraday Trading}
\label{appendix:sc_0dte}

\subsubsection{Background}

Zero days-to-expiration (0DTE) options are contracts that expire within the same trading session in which they are opened. These ultra-short maturity instruments exhibit a distinctive set of characteristics that set them apart from conventional option contracts. As time-to-expiry approaches zero, gamma exposure becomes extremely elevated, meaning that even modest intraday price fluctuations can produce disproportionately large changes in option value~\citep{augen2009day, augen2009trading, beckmeyer2023retail}. Simultaneously, theta decay accelerates sharply, eroding the time value of held positions at a rate far exceeding that of longer-dated contracts. Together, these properties create a highly nonlinear payoff environment in which option value dynamics are dominated by short-term price sensitivity rather than longer-horizon volatility expectations.

Because of these characteristics, 0DTE contracts have become widely used instruments for intraday speculation, event-driven positioning, and same-session hedging around high-impact announcements. Recent evidence suggests that their adoption has grown substantially among speculative retail participants~\citep{brogaard2023does}, reflecting broader accessibility to intraday option markets. Trading activity is necessarily concentrated within a single session, which demands rapid reaction to market developments, strict intraday risk control, and efficient capital deployment under extreme time pressure.

\subsubsection{Settings}

An idealized 0DTE trading session unfolds across three operationally distinct phases. The first is the early-session positioning phase, covering the market open through approximately the first hour of trading. During this window, the agent assesses the prevailing intraday trend, observes the term structure of implied volatility in the 0DTE option chain, and initiates positions at strikes and structures consistent with its short-horizon view. The second is the intra-session management phase, spanning the remainder of the morning through mid-afternoon. During this period, the agent must respond to evolving intraday price action, identify and react to volatility spikes associated with scheduled macroeconomic releases or other catalysts, and continuously manage gamma exposure as the proximity to expiration intensifies price sensitivity. The third is the expiration-horizon phase, covering the final hour of trading. As options approach settlement, the agent must close or roll remaining positions within the available exit window, balancing the risk of holding through expiration against the cost of early liquidation.

The strategy space available to the agent includes the core instruments used in intraday option trading. For directional positioning, the agent may take long calls or long puts to express a view on the near-term price direction of the underlying. For volatility-driven or non-directional positioning, the agent may construct straddles or strangles to profit from a large intraday move regardless of direction. To manage the premium cost and cap the risk of short-horizon positions, the agent may deploy vertical spreads, either call spreads or put spreads, which reduce net premium outlay at the cost of capped upside. All positions are subject to the constraint that they must be resolved within the same trading session, either through voluntary exit or automatic settlement at expiration.

\subsubsection{Experiment}

We select \textbf{SPY} as the underlying asset for the 0DTE scenario. SPY is among the most actively traded instruments in both U.S. equity and option markets and offers daily option expirations, a structure that makes it uniquely suited to same-session trading~\citep{zarattini2024beat}. Unlike most underlyings that provide only weekly maturities, the daily expiration calendar of SPY allows positions to be constructed specifically around same-day catalysts without exposure to overnight risk. We select all Wednesdays and Fridays in 2025 that fall on active trading days as test dates, yielding a total of 102 independent sessions. This selection targets the two most heavily used expiration days among 0DTE practitioners and captures a representative cross-section of intraday market conditions across the evaluation period.

Each test session spans the full regular trading session from 9:30 AM to 4:00 PM ET. The agent receives intraday price and option chain data at each decision point and must manage its positions through the three phases described above. We construct sub-scenarios within sessions that are centered on identifiable volatility events, including scheduled macroeconomic releases and intraday liquidity shocks, to stress the agent's responsiveness under rapidly changing conditions specifically.

\subsection{Statistics of Scenarios}

\subsubsection{Selected Assets for Overlay Scenario}
\label{appendix:overlay_stat}

\begin{figure} [h]
\vspace{-12pt}
    \centering
    \includegraphics[width=0.95\linewidth]{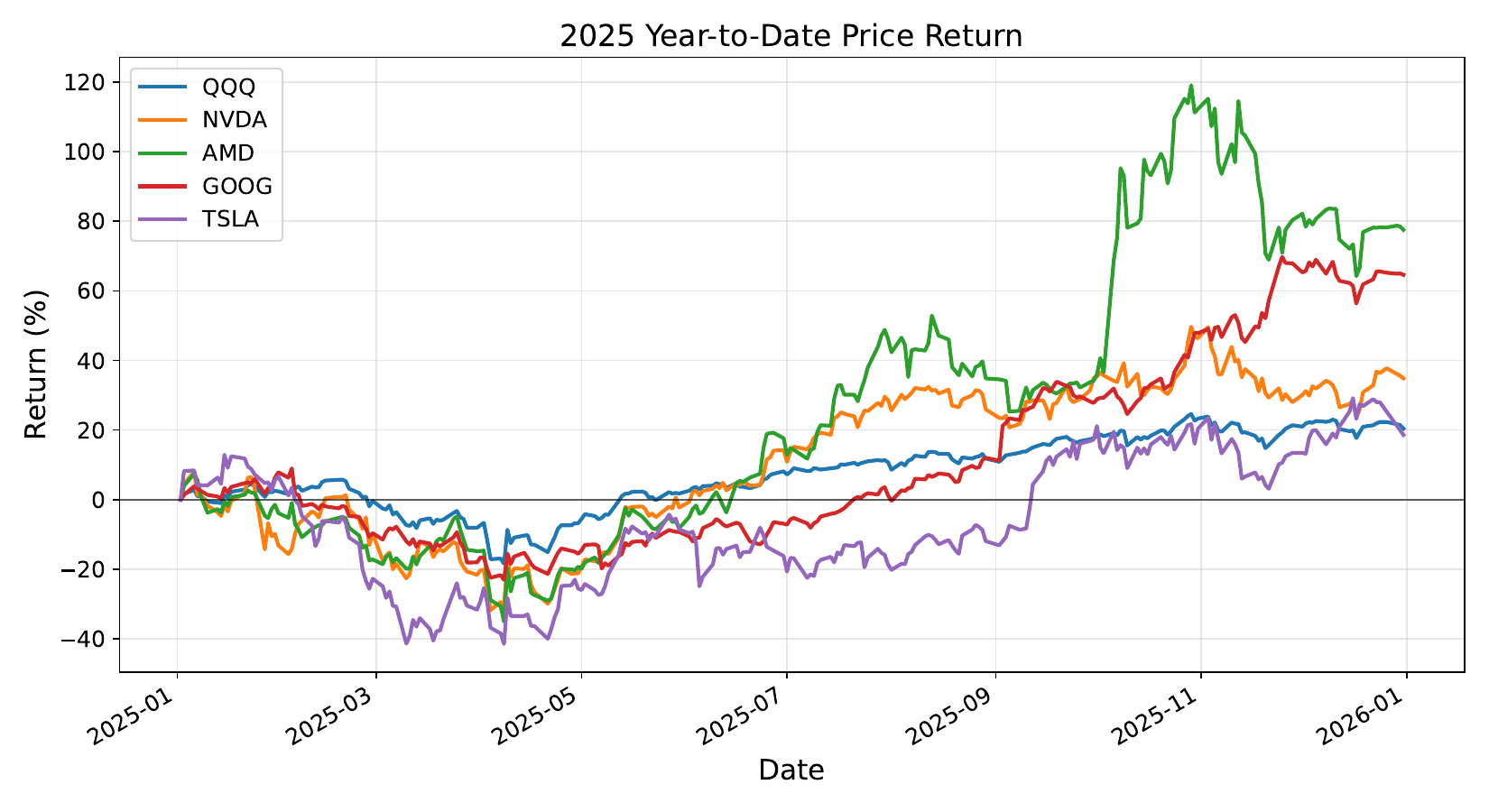}
    \caption{2025 year-to-date price return for QQQ and four single-name underlyings (NVDA, AMD, GOOG, TSLA), rebased to the first 2025 trading day (2025-01-02 close).}
    \label{fig:ytd_plot}
\end{figure}

The five overlay underlyings, QQQ, NVDA, AMD, GOOG, and TSLA, are chosen to span the regimes a single-name option overlay must be evaluated against. Figure~\ref{fig:ytd_plot} and Table~\ref{tab:ticker_performance} show that the 2025 total returns range from $+18.57\%$ (TSLA) to $+77.54\%$ (AMD), a $\approx 59$ pp spread around the index benchmark QQQ ($+20.40\%$); the rebased curves do not collapse into a single bundle, so per-name PnL is driven by distinct return processes rather than a single common factor.

The panel also spans a wide risk and tail range. Annualized volatility runs from $23.62\%$ (QQQ) to $63.31\%$ (TSLA), realized Sharpe at $r_f=0$ from $0.30$ to $2.05$, and maximum drawdown from $-22.88\%$ to $-48.19\%$, so protective and premium-collecting overlays are exercised across drawdown depths an index-only test would not reach. Higher moments diverge too, from QQQ's fat right tail (excess kurtosis $17.81$) to GOOG's near-Gaussian shape and AMD's asymmetric best and worst days ($+23.82\%$ against $-8.90\%$), implying different gamma and vega exposures by name. All five are US large-cap technology or AI-adjacent equities, so the results describe overlay behaviour within that regime.

\begin{table}[h]
\vspace{-12pt}
  \caption{Performance metrics and risk characteristics of selected technology tickers. Returns and volatility are annualized; Sharpe ratio is calculated assuming a risk-free rate of zero ($r_f=0$). Metrics are derived from daily closing prices for the 2025 fiscal year.}
  \label{tab:ticker_performance}
  \centering
  \small
  \begin{tabular}{@{} l S[table-format=2.2] S[table-format=2.2] S[table-format=2.2] S[table-format=1.2] S[table-format=-1.2] S[table-format=2.2] S[table-format=2.2] S[table-format=-2.2] S[table-format=-2.2] @{}}
    \toprule
    \textbf{Ticker} & {\textbf{Tot. Ret.}} & {\textbf{Ann. Ret.}} & {\textbf{Ann. Vol.}} & {\textbf{Sharpe}} & {\textbf{Skew.}} & {\textbf{Kurt.}} & {\textbf{Best}} & {\textbf{Worst}} & {\textbf{MDD}} \\
    & {(\%)} & {(\%)} & {(\%)} & & & & {(\%)} & {(\%)} & {(\%)} \\
    \midrule
    QQQ  & 20.40 & 20.67 & 23.62 & 0.88 & 1.31  & 17.81 & 12.00 & -6.21  & -22.88 \\
    NVDA & 34.84 & 35.33 & 49.64 & 0.71 & -0.08 & 8.00  & 18.72 & -16.97 & -36.89 \\
    AMD  & 77.54 & 78.77 & 60.73 & 1.30 & 1.79  & 10.68 & 23.82 & -8.90  & -39.63 \\
    GOOG & 64.61 & 65.60 & 31.99 & 2.05 & 0.33  & 3.99  & 9.88  & -7.51  & -29.43 \\
    TSLA & 18.57 & 18.82 & 63.31 & 0.30 & 0.41  & 4.79  & 22.69 & -15.43 & -48.19 \\
    \bottomrule
  \end{tabular}
\end{table}

\begin{table}[h]
  \caption{Earnings-announcement price reactions by ticker in 2025. Values represent percentage changes ($\Delta\%$) aligned by decimal point. \textbf{Open} denotes the reaction from $T-1$ close to $T+1$ open; \textbf{Close} denotes $T-1$ close to $T+1$ close. Information is curated from evaluated technical reports.}
  \label{tab:merged_earnings_complete}
  \centering
  \scriptsize
  \setlength{\tabcolsep}{1pt}
  \begin{tabular*}{\textwidth}{@{\extracolsep{\fill}} l *{10}{S[table-format=-2.2]} @{}}
    \toprule
    & \multicolumn{2}{c}{\textbf{Q1}} & \multicolumn{2}{c}{\textbf{Q2}} & \multicolumn{2}{c}{\textbf{Q3}} & \multicolumn{2}{c}{\textbf{Q4}} & \multicolumn{2}{c}{\textbf{Overall}} \\
    \cmidrule(lr){2-3} \cmidrule(lr){4-5} \cmidrule(lr){6-7} \cmidrule(lr){8-9} \cmidrule(lr){10-11}
    \textbf{Ticker} & {\textbf{Open}} & {\textbf{Close}} & {\textbf{Open}} & {\textbf{Close}} & {\textbf{Open}} & {\textbf{Close}} & {\textbf{Open}} & {\textbf{Close}} & {\textbf{Open}} & {\textbf{Close}} \\
    \midrule
    AAPL & -3.20 & -4.03 & -4.79 & -6.76 & -1.48 & -2.03 & -0.36 & -0.87 & -2.46 & -3.42 \\
    AMD  & -7.17 & -7.82 & +3.35 & +3.12 & -4.29 & -1.10 & +1.37 & -4.94 & -1.69 & -2.68 \\
    AMZN & -3.47 & -2.38 & -1.94 & -2.02 & -7.14 & -9.59 & +14.58& +13.97& +0.51 & -0.01 \\
    AVGO & +5.66 & +2.79 & -5.73 & -6.02 & +11.97& +12.92& -10.95& -16.38& +0.24 & -1.67 \\
    CELH & +27.09& +22.88& +3.72 & +5.61 & +19.68& +21.55& -23.72& -30.71& +6.69 & +4.83 \\
    COIN & -6.48 & -11.23& +1.18 & +0.35 & -15.41& -15.77& +3.76 & +0.58 & -4.24 & -6.52 \\
    CRM  & -3.70 & -3.08 & -2.88 & -3.86 & -4.53 & -2.22 & +5.07 & +9.15 & -1.51 & -0.00 \\
    DIS  & -1.25 & -1.07 & +12.80& +14.05& -2.17 & -4.60 & -7.83 & -9.30 & +0.39 & -0.23 \\
    HOOD & +13.23& +16.76& -4.75 & -1.06 & -7.82 & -5.84 & -13.13& -8.51 & -3.12 & +0.34 \\
    INTC & -5.20 & -3.15 & -6.38 & -4.56 & -8.00 & -8.62 & +0.60 & +3.62 & -4.74 & -3.18 \\
    META & +2.14 & +1.88 & +6.28 & +8.75 & +9.42 & +7.88 & -10.27& -13.75& +1.90 & +1.19 \\
    MSFT & -5.28 & -6.17 & +9.23 & +10.12& +4.24 & +2.12 & -2.34 & -4.38 & +1.46 & +0.42 \\
    MU   & -6.70 & -5.88 & -1.66 & -1.96 & -5.42 & -5.76 & +11.63& +17.91& -0.54 & +1.08 \\
    NFLX & +10.13& +13.24& +3.32 & +6.92 & -5.35 & -3.21 & -9.22 & -10.29& -0.28 & +1.66 \\
    NKE  & -4.59 & -6.22 & +14.25& +13.59& +7.56 & +6.94 & -10.12& -12.81& +1.77 & +0.37 \\
    NVDA & -10.10& -4.84 & +2.90 & +0.24 & -1.92 & -4.09 & -2.83 & -4.10 & -2.99 & -3.20 \\
    PLTR & +21.54& +21.04& -12.77& -10.74& +6.54 & +11.75& -8.77 & -9.31 & +1.64 & +3.19 \\
    PYPL & -12.62& -11.52& +0.35 & +1.40 & -7.71 & -10.88& +3.49 & -0.81 & -4.12 & -5.45 \\
    RKLB & +7.11 & -5.90 & -5.82 & -7.97 & +1.00 & +1.83 & +0.06 & -3.72 & +0.59 & -3.94 \\
    SHOP & -0.71 & +0.49 & -0.03 & -2.89 & +21.62& +18.95& -7.74 & -5.79 & +3.29 & +2.69 \\
    SMCI & +7.74 & -5.69 & -0.03 & -2.52 & -17.66& -18.49& -10.34& -14.91& -5.07 & -10.40\\
    SNOW & +3.51 & +6.56 & +10.82& +11.68& +18.23& +19.10& -11.85& -13.66& +5.18 & +5.92 \\
    SOFI & -10.04& -9.77 & -3.14 & -5.23 & -0.90 & +4.04 & +4.98 & +3.00 & -2.28 & -1.99 \\
    TSLA & +3.19 & +3.98 & +5.27 & +9.05 & -7.16 & -4.96 & +1.79 & -1.20 & +0.77 & +1.72 \\
    \bottomrule
  \end{tabular*}
\end{table}

\subsubsection{Analysis of Earning Events}

The earnings sample, summarized in Table~\ref{tab:merged_earnings_complete},
is intentionally compact yet demonstrably diverse in both magnitude and
pattern. 

\textbf{Volatility is amply represented}: of the 24 constituents,
18 (75\%) print at least one single-event move of $\geq 10\%$, 9 (38\%)
exceed $15\%$ in at least one quarter, and 3 (13\%) cross the $20\%$
threshold. The realized cross-sectional dispersion spans roughly 58
percentage points, with the most extreme positive print at $+27.09\%$ and
the most extreme negative at $-30.71\%$. 

\textbf{Pattern heterogeneity is equally pronounced.} Classifying each ticker by the sign sequence of its
four quarterly close-reactions, the universe partitions into four
qualitatively distinct archetypes: (i) \emph{persistent drift} (same sign
in all four quarters), 2 names; (ii) \emph{single-reversal regime change}
(exactly one sign flip across the year, typically a late-year capitulation
or breakout), 11 names; (iii) \emph{whipsaw} (two or more sign flips,
indicative of alternating beat/miss regimes), 11 names; and (iv)
\emph{tail-event dominated} (single-quarter $\geq 15\%$ shock that drives
the annual average), with 9 names exhibiting this property in addition to
a primary archetype. The full-year overall reaction is roughly balanced
across direction, with 11 of 24 names net-positive and 13 net-negative on
the close-to-close column. The simultaneous presence of these regimes
within a 24-name universe ensures the benchmark stresses agents across
both volatility-expansion and volatility-collapse states, persistent-trend
and mean-reverting trajectories, and asymmetric tail events in either
direction, rather than concentrating risk in a single behavioral mode.

\subsubsection{Intraday Statistics}
We characterize the data landscape of SPY on Wednesdays and Fridays to evaluate the model's 0DTE intraday trading performance. We define two key indicators below:

Daily return is the log ratio of the session's last close to its first open, from minute bars. Daily volatility is the Garman--Klass estimator~\citep{garman1980estimation} applied to 5-minute OHLC bars and aggregated across the session, which uses the intraday range and so is more efficient than a close-to-close estimate at this sample size.

\begin{table}[h]
    \caption{Summary statistics of daily returns and GK volatility by
             day-of-week group over 250 trading days.
             All values are in percentage~(\%).}
    \label{tab:summary_stats}
    \centering
    \small
    \begin{tabular}{lcrrrrrrrr}
        \toprule
        & & \multicolumn{4}{c}{Return (\%)}
          & \multicolumn{4}{c}{GK Volatility (\%)} \\
        \cmidrule(lr){3-6}\cmidrule(lr){7-10}
        Group & $N$ & Mean & Std & P25 & P75
                    & Mean & Std & P25 & P75 \\
        \midrule
        Overall   & 250
            & $0.0911$ & $1.2558$ & $-0.3760$ & $0.4846$
            & $0.8214$ & $0.8630$ & $0.4289$ & $0.8557$ \\
        Wednesday &  52
            & $0.2181$ & $1.4654$ & $-0.2520$ & $0.4612$
            & $0.8762$ & $1.1162$ & $0.4289$ & $0.8599$ \\
        Friday    &  50
            & $-0.0409$ & $0.9903$ & $-0.3955$ & $0.5506$
            & $0.7643$ & $0.4816$ & $0.4389$ & $0.8477$ \\
        \bottomrule
    \end{tabular}
\end{table}

\begin{wraptable}{r}{0.6\textwidth}
    \centering
    \small
    \vspace{-12pt} %
    \caption{Intraday and tail-risk statistics by day-of-week group. Tail counts report observations exceeding the stated multiple of the full-sample standard deviation, with the group percentage in parentheses.}
    \label{tab:intraday_stats}
    \begin{tabular}{lrrr}
        \toprule
        Metric & Wed. & Fri. & Overall \\
        \midrule
        Mean intraday range (\%)  & 1.2697 & 1.1902             & 1.1791  \\
        Mean absolute return (\%) & 0.6677 & 0.6916             & 0.6115  \\
        Skewness                  & 5.3916 & $-$1.2568          & 2.6255  \\
        Kurtosis                  & 30.239 & \phantom{$-$}2.636 & 32.623  \\
        \midrule
        Up ${>}2\sigma$   & 1 (1.9\%) & 1 (2.0\%) & 3 (1.2\%) \\
        Up ${>}3\sigma$   & 1 (1.9\%) & 0 (0.0\%) & 2 (0.8\%) \\
        Down ${>}2\sigma$ & 0 (0.0\%) & 2 (4.0\%) & 4 (1.6\%) \\
        Down ${>}3\sigma$ & 0 (0.0\%) & 1 (2.0\%) & 2 (0.8\%) \\
        \midrule
        Max single, day gain (\%) & $+$9.8099 & $+$2.1432 & $+$9.8099 \\
        Max single, day loss (\%) & $-$1.1688 & $-$3.5116 & $-$4.9755 \\
        \bottomrule
    \end{tabular}
\end{wraptable}

Tables~\ref{tab:summary_stats} and~\ref{tab:intraday_stats} separate the two session groups. Wednesday carries the higher mean return, $0.2181\%$ against a full-sample $0.0911\%$, and the higher GK volatility, $0.8762\%$, but the mean rests on a thin right tail: skewness $5.39$, kurtosis $30.24$, and a largest single-day gain of $+9.81\%$. Friday is flat in the mean, $-0.0409\%$, and mildly left-skewed at $-1.26$, with losses beyond $2\sigma$ on $4.0\%$ of sessions against none on Wednesday and a worst day of $-3.51\%$ against Wednesday's $-1.17\%$. Volatility interquartile ranges are close across the two groups, so the difference sits in the tails rather than in the level of activity.
The intraday track runs on the Wednesday and Friday subset rather than the full week, for cost reasons. The two days give $102$ of the $250$ trading days, $N=52$ and $N=50$. They were fixed \emph{a priori} by their position in the weekly expiry cycle, not by realized returns, so the subset is not selected on the outcome; and since they are the most positively and most negatively skewed day-of-week groups in the sample, the pair does not narrow the range of regimes the track sees.

\section{Details in Evaluation}
\label{appendix:evaluation_metrics}

\subsection{Common Metrics}
\label{appendix:common_metrics}

LiveOption contains scenarios with two fundamentally different temporal
structures, and therefore requires two complementary evaluation
protocols. \textbf{Overlay Strategies} is
treated as \emph{continuous long-horizon portfolio management} tasks:
the agent operates over an extended trading window and is judged at the
portfolio level. \textbf{0DTE Intraday Trading} and \textbf{Earnings
Bet} are treated as \emph{episode-based option trading} tasks: each
trading day or earnings event is an independent episode, performance is
measured per episode, and statistics are aggregated across episodes.
Accordingly, we adopt two groups of metrics:
(i) \emph{portfolio metrics} for long-horizon scenarios, in both
absolute (own portfolio) and relative (vs.\ benchmark) form, and
(ii) \emph{episode metrics} for short-horizon option trading, focused on
trade-level outcome statistics (Win\%, PnL\%, P/L, PF).

\paragraph{Portfolio Metrics.}
For the Overlay Strategies task, let $V_t$ be the portfolio
value at time $t$ and $r_t = V_t/V_{t-1}-1$ the periodic return
(daily). We report three standard absolute portfolio measures.

\underline{\textit{Annualized Return (AR)}}:
\[
\text{AR} \;=\; \left(\frac{V_T}{V_0}\right)^{A/T} - 1,
\]
where $A=252$ is the daily annualization factor and $T$ is the number
of periods in the evaluation window.

\underline{\textit{Sharpe Ratio (SR)}}:
\[
\text{SR} \;=\; \frac{\mathbb{E}[r_t - r_f]}{\mathrm{Std}[r_t - r_f]} \cdot \sqrt{A},
\]
with risk-free rate $r_f$ set to $0$ when not specified.

\underline{\textit{Maximum Drawdown (MDD)}}:
\[
\text{MDD} \;=\; \max_{t}\!\left(1 - \frac{V_t}{\max_{u \le t} V_u}\right).
\]

These three numbers describe the portfolio's profitability,
risk-adjusted return, and worst-case capital loss \emph{on its own
terms}, without reference to any benchmark.

\paragraph{Relative Metrics.}
Absolute metrics alone are not sufficient for the Overlay Strategy: the
overlay sits on top of a buy-and-hold underlying, so a higher SR or
lower MDD may simply reflect the underlying's behavior rather than any
edge added by the option overlay. Worse, the option leg distorts the
return distribution non-linearly through gamma, vega, and jump
exposure, so the Sharpe ratio of the combined portfolio is no longer a
clean signal of overlay quality. We therefore complement the absolute
metrics with two \emph{relative} metrics that explicitly benchmark the
agent portfolio against the buy-and-hold leg on the same underlying.

Let $r_{p,t}$ be the daily return of the agent portfolio and
$r_{b,t}$ the daily return of the buy-and-hold benchmark. Define the
\emph{active return} on day $t$ as
\[
    \Delta r_t \;=\; r_{p,t} - r_{b,t}.
\]

\underline{\textit{Cumulative Delta Return ($\Delta R$)}}:
\[
    \Delta R \;=\; \prod_{t=1}^{T}\!\bigl(1 + r_{p,t}\bigr) \;-\; \prod_{t=1}^{T}\!\bigl(1 + r_{b,t}\bigr),
\]
the incremental cumulative PnL attributable to the option overlay
after netting out the directional exposure already supplied by the
underlying. $\Delta R$ measures the \emph{magnitude} of the overlay's
contribution.

\underline{\textit{Information Ratio (IR)}}:
\[
    \mathrm{IR} \;=\; \frac{\overline{\Delta r}}{\sigma(\Delta r)} \cdot \sqrt{A}
    \;=\; \frac{\tfrac{1}{T}\sum_{t} \Delta r_t}{\sqrt{\tfrac{1}{T-1}\sum_{t}\bigl(\Delta r_t - \overline{\Delta r}\bigr)^{2}}} \cdot \sqrt{A},
\]
which normalizes the active return by its tracking error. Unlike SR,
IR penalizes only the variance of \emph{deviations from the
benchmark}, so an overlay that closely tracks the underlying while
delivering a small but consistent edge scores well even when the
absolute portfolio volatility is inflated by the option leg. IR thus
measures the \emph{consistency} of the overlay's contribution.

\paragraph{Episode Metrics.}
For episode-based scenarios, each episode (one trading day for 0DTE,
one earnings window for Earnings Bet) yields a closed PnL; metrics are
computed over the population of episodes and reported as cross-episode
averages. Let $N$ be the total number of episodes, $\pi_i$ the realized
PnL of episode $i$, and $\rho_i = \pi_i / C_i$ its return relative to
the capital $C_i$ committed in that episode.

\underline{\textit{Win Rate (Win\%)}}:
\[
    \text{Win\%} \;=\; \frac{|\{i : \pi_i > 0\}|}{N},
\]
the fraction of episodes that close profitably. Win\% measures
\emph{how often} the agent makes money, but says nothing about the
size of wins versus losses, so it is always read jointly with the next
three metrics.

\underline{\textit{Average PnL Return (PnL\%)}}:
\[
    \text{PnL\%} \;=\; \frac{1}{N}\sum_{i=1}^{N} \rho_i,
\]
the mean per-episode return on committed capital. PnL\% is the
headline expectancy of the strategy: a positive value means the
average episode is profitable after combining wins, losses, and
breakevens.

\underline{\textit{Payoff Ratio (P/L)}}:
\[
    \text{P/L} \;=\; \frac{\overline{\pi}_{+}}{\bigl|\,\overline{\pi}_{-}\,\bigr|}
    \;=\; \frac{\mathbb{E}[\pi_i \mid \pi_i > 0]}{\bigl|\,\mathbb{E}[\pi_i \mid \pi_i < 0]\,\bigr|},
\]
the ratio of the average winning episode to the average losing
episode. P/L describes the \emph{shape} of the payoff distribution:
$\text{P/L} > 1$ means the agent's wins are larger than its losses on
average, which can compensate for a sub-50\% Win\%. Conversely, a high
Win\% with $\text{P/L} \ll 1$ flags a strategy that wins frequently
but is vulnerable to a few large losers wiping out many small gains.

\underline{\textit{Profit Factor (PF)}}:
\[
    \text{PF} \;=\; \frac{\sum_{i:\,\pi_i > 0} \pi_i}{\bigl|\sum_{i:\,\pi_i < 0} \pi_i\bigr|},
\]
the gross dollars won divided by the gross dollars lost across all
episodes. PF combines Win\% and P/L into a single capital-efficiency
ratio: $\text{PF} > 1$ means the strategy is net profitable in
aggregate, $\text{PF} = 1$ is breakeven, and values $\gg 1$ indicate
that winning episodes dominate losing episodes both in frequency and
in size. PF is the most concise summary of episode-based performance
because it is invariant to episode count and directly answers ``for
every dollar lost, how many dollars were won?''

\paragraph{Uncertainty and multiplicity.}
Mean returns and mean drawdowns are reported with percentile bootstrap
intervals from $10{,}000$ episode resamples at fixed seeds. Medians and
$10\%$ and $20\%$ trimmed means accompany every mean so that right-tail
sensitivity is visible. Cross-model comparisons are paired on shared
cases; ablation effects are paired within model and case. Paired effects
are tested with two-sided Monte Carlo sign-flip tests using $50{,}000$
draws, and Holm correction is applied within each comparison family, the
$15$ pairs of a task or the three models of an ablation profile. A
no-trade episode has zero fills and stays in the distribution with its
observed return.

\subsection{Behavior Analysis}
\label{appendix:behavior_analysis}

The metrics defined in Appendix~\ref{appendix:common_metrics}
characterize trading \emph{outcomes}: the magnitude of profit, the
frequency of success, and the associated risk profile. They do not,
however, reveal \emph{how} an agent arrives at these outcomes, that
is, whether profits originate from a sound option-mechanic edge or
from a fortunate directional bet, and whether losses arise from
incorrect market views or from execution-level cascades. Since LLM
agents reason in natural language and emit structured trade intents,
LiveOption preserves the complete decision trace of every run,
including the thesis text, the intent stream, the fill log, and the
Greek snapshots. This trace enables fine-grained behavioral auditing
that aggregate PnL cannot provide. We therefore complement the
outcome metrics with a \emph{behavior analysis} that diagnoses the
underlying causes of an agent's performance along five axes. The
analyses presented in Sections~\ref{sec:analysis} and onward all
operate within this framework, which we make explicit below.

\paragraph{Strategy Composition.}
We tabulate, for each model and scenario, the empirical distribution
over option structures actually traded, including single legs (naked
calls, naked puts), vertical spreads (bull and bear, debit and
credit), volatility structures (straddles, strangles), and overlay
structures (covered calls, cash-secured puts, collars). The
composition is interpreted jointly with the episode metrics: a high
Win\% accompanied by a structure mix that is overwhelmingly
direction-agnostic (such as straddles) carries different implications
than the same Win\% obtained from directional spreads. Strategy
composition further exposes \emph{strategy collapse}, in which a
model defaults to a single structure across heterogeneous scenarios
regardless of the prevailing market regime.

\paragraph{Directional versus Structural Reasoning.}
For each trade, we extract two signals from the LLM's thesis: a
directional polarity (bullish, bearish, or neutral) and any explicit
reference to option mechanics
(Greeks, implied versus realized volatility, vega exposure, IV
crush). Each trade is then classified as \emph{directional}, when the
decision is dominated by a directional view, or \emph{structural},
when an option-mechanic edge dominates the decision. The
resulting split is reported per model and is used to test whether
profitable runs are driven by repeatable structural edges or by
fortunate directional streaks.

\paragraph{Position Sizing and Leverage Control.}
We track the capital deployed per trade as a fraction of available
buying power and the gross notional carried into the next session.
The failure mode this axis detects is \emph{leverage stacking}, in
which rolling, pyramiding, or wing-overwriting drives the effective
leverage above the initial allocation. The latter mechanism is the proximate cause of the
sub-$-100\%$ earnings returns documented in
Section~\ref{sec:analysis}.

\subsection{Details of Evaluated LLMs}
\label{appendix:eval_llm}

We evaluate eight LLMs in Table~\ref{tab:llm_detail}, spanning eight model families and a wide parameter range. Six are served through a single hosted inference provider, so serving conditions are held constant within that group; the two proprietary models are reached through a separate gateway, on the wire protocol each vendor exposes. All experiments use 2025 market data. The reported cutoffs are listed alongside each model and they do not line up behind that window: two models are trained to a date that precedes it, one reports a cutoff of January 2026 that postdates it entirely, one falls inside it, and four vendors report no cutoff at all. The evaluation window is therefore not a contamination control for this panel, and we do not present it as one. What limits look-ahead within a run is the observation gating described in Appendix~\ref{appendix:data}: at every decision step the engine exposes only information timestamped before that step, so an agent cannot read prices, news or events it has not yet reached. That constrains the use of future information during a run; it cannot rule out that a model recalls 2025 outcomes from pretraining, and results for the models whose cutoffs cover the window should be read with that possibility open.

\begin{table}[h]
  \caption{Evaluated large language models. Recorded identifiers correspond
  to the exact model strings logged during evaluation. Knowledge cutoffs
  are based on publicly available model documentation. All models are
  evaluated at temperature $0.5$.}
  \label{tab:llm_detail}
  \centering
  \small
  \setlength{\tabcolsep}{5pt}
  \begin{adjustbox}{width=\linewidth}
  \begin{tabular}{@{}llllc@{}}
    \toprule
    \textbf{Family} & \textbf{Model} & \textbf{Recorded identifier} &
    \textbf{Knowledge Cutoff} & \textbf{Source} \\
    \midrule
    DeepSeek & DeepSeek-V4-Flash
      & \texttt{deepseek-ai/DeepSeek-V4-Flash-0731}
      & N/A & \cite{deepseekai2026v4} \\
    OpenAI & GPT-OSS-120B
      & \texttt{openai/gpt-oss-120b}
      & Jun.\ 2024 & \cite{openai2025gptoss} \\
    Qwen & Qwen3-235B-A22B
      & \texttt{Qwen/Qwen3-235B-A22B-Instruct-2507}
      & Jun.\ 2025$^{\dagger}$ & \cite{qwen3} \\
    Meta & Llama-3.3-70B-Instruct
      & \texttt{meta-llama/Llama-3.3-70B-Instruct}
      & Dec.\ 2023 & \cite{grattafiori2024llama3} \\
    MiniMax & MiniMax-M3
      & \texttt{MiniMaxAI/MiniMax-M3}
      & Jan.\ 2026 & \cite{lai2026minimax} \\
    Z.ai & GLM-5.3-Flash
      & \texttt{zai-org/GLM-5.3-Flash}
      & N/A & \cite{glm2026glm5,zai2026glm53} \\
    OpenAI & GPT-5.6-Terra
      & \texttt{gpt-5.6-terra}
      & Feb.\ 2026 & \cite{openai_gpt56_terra} \\
    Google & Gemini-3-Flash
      & \texttt{gemini-3-flash}
      & Jan.\ 2025 & \cite{google_gemini3_flash} \\
    \bottomrule
  \end{tabular}
  \end{adjustbox}

  \vspace{2pt}
  {\footnotesize
  $^{\dagger}$ The reported cutoff falls within the 2025 evaluation
  window. N/A indicates that no official knowledge cutoff was identified.
  }
\end{table}

\section{Details in Agent Design}

This section details the two representations the agent sees in place of a raw option chain: an anchored summary of the implied-volatility surface, and a quantized chain that keeps the tradable candidates within the context budget.

\subsection{Anchor-Based Volatility Surface Representation (AVSR)}
\label{appendix:avsr}
The implied volatility (IV) surface is a fundamental component in option pricing and trading, encoding rich information about market expectations across strikes and maturities. However, directly exposing the full surface to LLM agents is impractical due to its high dimensionality and irregular structure. To address this, we propose an \emph{Anchor-Based Volatility Surface Representation (AVSR)}, which approximates the surface using a small set of structured anchor points.

Formally, let $S_t$ denote the underlying price at time $t$. We define a set of \emph{moneyness anchors}:
\[
\mathcal{M} = \{m_1, m_2, m_3\},
\]
where each $m_i$ corresponds to a standardized moneyness level, e.g.,
\[
m_1 = 0.95,\quad m_2 = 1.0,\quad m_3 = 1.05,
\]
representing out-of-the-money put, at-the-money, and out-of-the-money call, respectively.

We further define a set of \emph{time-to-expiry anchors}: $\mathcal{T} = \{\tau_1, \tau_2, \tau_3\}$, for example,
$\tau_1 = 7\text{D}, \tau_2 = 30\text{D}, \tau_3 = 90\text{D}.$ The anchor grid is defined as the Cartesian product:
$
\mathcal{A} = \mathcal{M} \times \mathcal{T}.
$

For each anchor $(m, \tau) \in \mathcal{A}$, we compute the implied volatility:
\[
\sigma(m, \tau) = \text{IV}(K = m \cdot S_t,\; T = \tau),
\]
where $\text{IV}(\cdot)$ denotes the implied volatility obtained from the option chain. In practice, if no exact contract exists at $(K, T)$, we select the nearest available contract or apply interpolation over strikes and maturities.

The final AVSR representation is given by:
$
\mathbf{V}_t = \left[ \sigma(m_i, \tau_j) \right]_{i=1,2,3;\; j=1,2,3} \in \mathbb{R}^{3 \times 3}.
$

To enhance interpretability, we optionally compute summary statistics derived from the anchor grid:

\[
\text{Risk Reversal}_{\tau_j} = \sigma(m_3, \tau_j) - \sigma(m_1, \tau_j),
\]

\[
\text{Butterfly}_{\tau_j} = \frac{\sigma(m_1, \tau_j) + \sigma(m_3, \tau_j)}{2} - \sigma(m_2, \tau_j),
\]

\[
\text{Term Spread} = \sigma(m_2, \tau_3) - \sigma(m_2, \tau_1).
\]

\begin{figure}[h]
    \centering
    \includegraphics[width=1.01\linewidth]{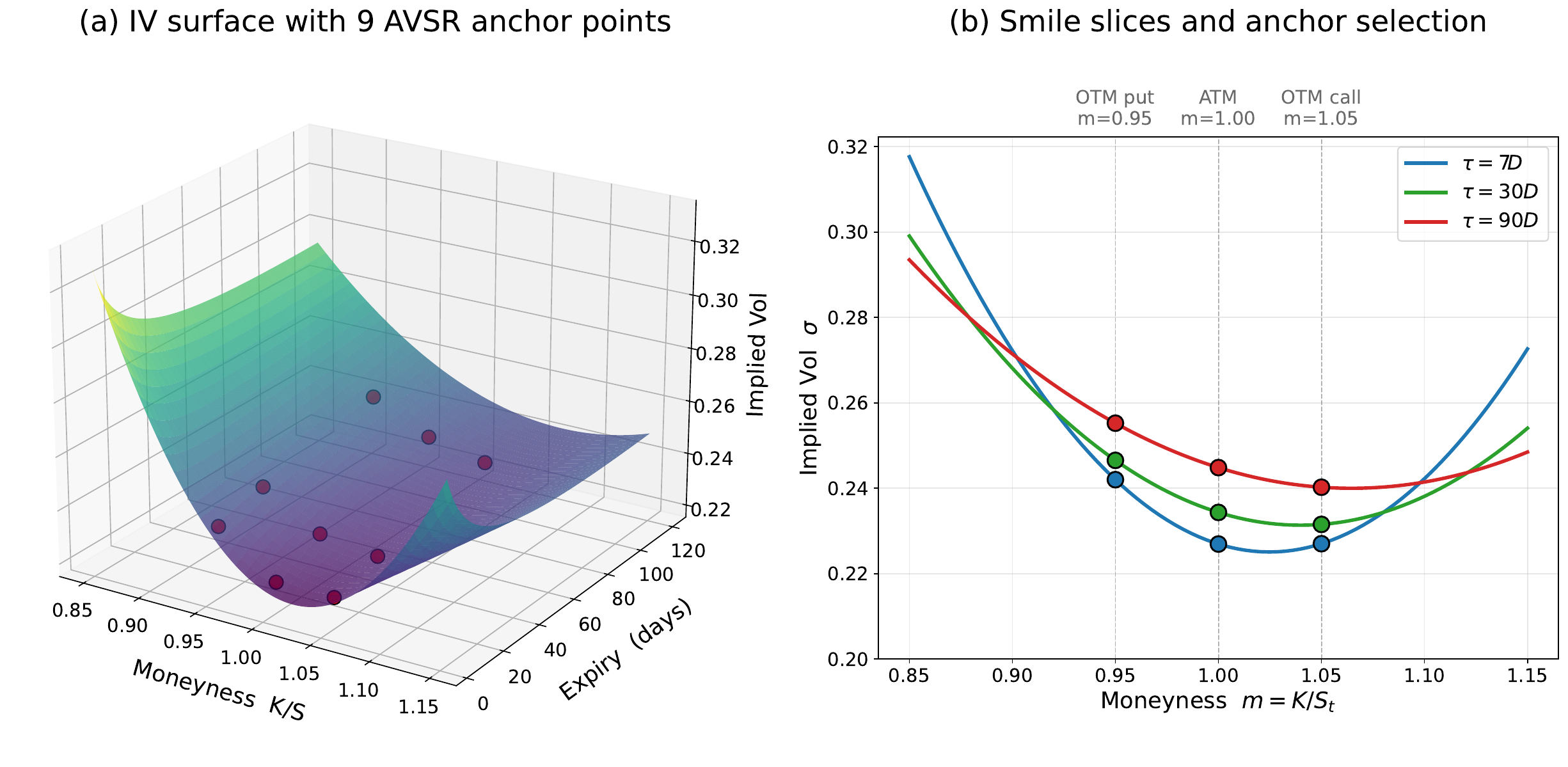}
  \caption{%
    Illustration of the Anchor-Based Volatility Surface Representation
    (AVSR). \textbf{(a)} A synthetic implied-volatility surface
    $\sigma(m,\tau)$ over moneyness $m=K/S_t$ and time-to-expiry $\tau$
    (days), with the nine red markers indicating the AVSR anchors at the
    grid $\mathcal{A}=\mathcal{M}\times\mathcal{T}$ with
    $\mathcal{M}=\{0.95,1.00,1.05\}$ and $\mathcal{T}=\{7\text{D},
    30\text{D}, 90\text{D}\}$.
    \textbf{(b)} The same surface visualized as smile slices at the three
    anchor maturities; the three filled markers on each curve are the
    selected anchors at $m \in \mathcal{M}$, jointly forming the compact
    representation
    $\mathbf{V}_t = [\sigma(m_i,\tau_j)]\in\mathbb{R}^{3\times 3}$ that
    is exposed to the LLM agent. By design, AVSR collapses the full,
    irregular surface into a fixed-size summary that captures level,
    skew, and term structure while remaining interpretable.%
  }
    \label{fig:example_avsr}
\end{figure}

We provide an example in Figure~\ref{fig:example_avsr}, these quantities capture the key structural properties of the IV surface, including skew, curvature, and term structure, while maintaining a compact and fixed-size representation suitable for LLM-based agents.

\subsection{Quantized Option Chain Representation (QOCR)}
\label{appendix:qocr}

While the IV surface provides a compact view of market conditions, executing option trades requires access to the option chain, which specifies all available contracts across strikes and expirations. In practice, the option chain is large, irregular, and highly redundant. To reduce complexity and improve tractability for LLM agents, we introduce a \emph{Quantized Option Chain Representation (QOCR)}, which discretizes the option chain into a structured action space.

We define a set of \emph{moneyness buckets}:
\[
\mathcal{B}_M = \{b_1, b_2, b_3\},
\]
for example:
\[
b_1 = [0.95, 0.99),\quad b_2 = [0.99, 1.01],\quad b_3 = (1.01, 1.05],
\]
representing OTM put, ATM, and OTM call regions in terms of normalized strike $K/S_t$.

Similarly, we define a set of \emph{expiry buckets}:
\[
\mathcal{B}_T = \{e_1, e_2, e_3\},
\]
such as:
\[
e_1 = [0, 14]\text{D},\quad e_2 = (14, 45]\text{D},\quad e_3 = (45, 120]\text{D}.
\]

Each bucket pair $(b, e) \in \mathcal{B}_M \times \mathcal{B}_T$ corresponds to a subset of contracts:
\[
\mathcal{C}(b, e) = \{(K, T) \mid K/S_t \in b,\; T \in e\}.
\]

We define a mapping function that selects a representative contract from each bucket:
\[
(K^*, T^*) = \arg\min_{(K, T) \in \mathcal{C}(b, e)} \left| \frac{K}{S_t} - \bar{m}(b) \right| + \lambda |T - \bar{\tau}(e)|,
\]
where $\bar{m}(b)$ and $\bar{\tau}(e)$ denote the center of the moneyness and expiry buckets, and $\lambda$ is a weighting parameter balancing strike and maturity proximity.

Under QOCR, an agent action is defined over discrete buckets rather than raw contracts:
\[
a_t = (u, s, b, e, q),
\]
where:
\begin{itemize}
    \item $u$ is the underlying asset,
    \item $s \in \{\text{call}, \text{put}\}$ is the option type,
    \item $b \in \mathcal{B}_M$ is the moneyness bucket,
    \item $e \in \mathcal{B}_T$ is the expiry bucket,
    \item $q$ is a quantized position size.
\end{itemize}

For multi-leg strategies, the action extends to:
\[
a_t = \{a_t^{(1)}, a_t^{(2)}, \dots, a_t^{(L)}\},
\]
where each leg $a_t^{(l)}$ follows the same bucketed structure.

This quantized formulation significantly reduces the size of the action space:
\[
|\mathcal{A}| = |\mathcal{B}_M| \times |\mathcal{B}_T| \times |\mathcal{S}| \times |\mathcal{Q}|,
\]
making it fixed and independent of the underlying option chain size.

By operating on discretized buckets, QOCR removes unnecessary granularity, enforces a consistent interface across assets, and enables efficient reasoning and execution for LLM-based trading agents.

\begin{figure}[h]
    \centering
    \includegraphics[width=0.95\linewidth]{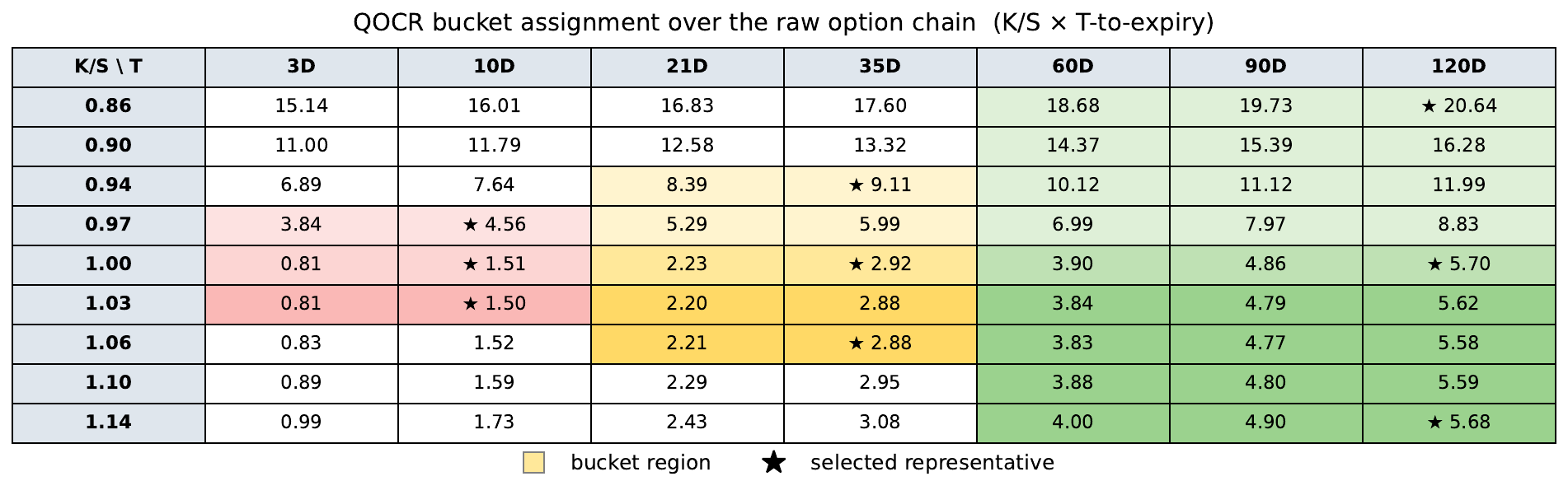}
  \caption{%
    \textbf{QOCR bucket assignment over a raw option chain.}
    The raw chain is partitioned into a $3 \times 3$ grid of
    (moneyness $\times$ time-to-expiry) buckets: rows by $K/S$
    (OTM put / ATM / OTM call, shaded red / yellow / green) and
    columns by $\tau$ (short / mid / long, shaded light / medium / dark).
    Moneyness wings widen with $\tau$, short-tenor options use a tight
    $\pm 3\%$ wing while long-tenor options span $\pm 14\%$,
    reflecting that far-dated contracts trade meaningfully across a
    wider strike range. The ATM band is fixed at
    $K/S \in [0.99,\, 1.01]$ for all maturities.
    Within each of the buckets, one representative contract
    ($\star$) is selected at the outer edge: the most extreme $K/S$
    for the wings and the $K/S$ closest to $1.0$ for ATM, with the
    longest $\tau$ as tiebreaker. These representatives form the
    QOCR feature vector handed to the policy. Cells outside any bucket
    (white) are discarded; mid-prices shown are illustrative.%
  }
    \label{fig:example_qocr}
\end{figure}

\newpage
\section{Additional Results}
\label{appendix:additional_results}

\subsection{Detail Performance }

\label{appendix:perf_detail}
\subsubsection{Overlay Strategies}
\label{appendix:detail_overlay}

We analyze the overlay performance per ticker and per mandate across the
eight evaluated models. Table~\ref{tab:overlay_per_asset}
reports return, risk-adjusted metrics, and option-flow statistics for the
Hedge and Profit~(Income) mandates over the full-year 2025 backtest, and
Figure~\ref{fig:overlay_active_return_grid} shows the daily active
return against the matched benchmark for every model, asset and mandate,
including the Feb--Apr~2025 downtrend where the protective value of the
overlay is most visible. Two conventions govern how the tables should be
read. First, $\Delta\text{Ret}\%$ is measured against a matched
buy-and-hold portfolio holding the same target share count, so it is the
column that captures the full economic effect of the overlay, including
contracts that expire without an explicit closing trade. Second,
$\Delta\text{MDD}$ is signed so that a positive value means the strategy
drew down less than that benchmark.

\setlength{\LTleft}{0pt}
\setlength{\LTright}{0pt}
\renewcommand{\arraystretch}{1.05}
\setlength{\tabcolsep}{4pt}
\footnotesize
\begin{longtable}{@{\extracolsep{\fill}} l r r r r r r r r @{}}
\caption{Overlay backtest performance per asset, full-year 2025. $\Delta$Ret is active return against the matched buy-and-hold portfolio, $\Delta$MDD the reduction in maximum drawdown (positive means shallower than the benchmark), IR the information ratio, and Fees the total commission and exchange fees paid. Each sub-heading gives that asset's buy-and-hold return and maximum drawdown, which are the benchmark the $\Delta$ columns are measured against.}\label{tab:overlay_per_asset} \\
\toprule
\textbf{Model} & \textbf{Ret\%} & \textbf{$\Delta$Ret} & \textbf{Sharpe} & \textbf{MDD\%} & \textbf{$\Delta$MDD} & \textbf{IR} & \textbf{Fills} & \textbf{Fees (\$)} \\
\midrule
\endfirsthead
\multicolumn{9}{l}{\footnotesize\itshape (continued from previous page)} \\
\toprule
\textbf{Model} & \textbf{Ret\%} & \textbf{$\Delta$Ret} & \textbf{Sharpe} & \textbf{MDD\%} & \textbf{$\Delta$MDD} & \textbf{IR} & \textbf{Fills} & \textbf{Fees (\$)} \\
\midrule
\endhead
\midrule
\multicolumn{9}{r}{\footnotesize\itshape continued on next page} \\
\endfoot
\bottomrule
\endlastfoot
\multicolumn{9}{@{}l}{\textbf{(a) QQQ}\quad buy-and-hold 17.00\%, maximum drawdown 19.24\%} \\
\midrule
\multicolumn{9}{@{}l}{\itshape Hedge} \\
\quad DeepSeek-V4-Flash & 10.38 & $-$6.62 & 0.669 & 16.62 & 2.62 & $-$0.93 & 95 & 3,255 \\
\quad GPT-OSS-120B & 15.95 & $-$1.05 & 1.011 & 14.06 & 5.18 & $-$0.28 & 131 & 3,788 \\
\quad Qwen3-235B-A22B & 11.66 & $-$5.33 & 0.690 & 19.95 & $-$0.71 & $-$1.74 & 71 & 2,280 \\
\quad Llama-3.3-70B-Instruct & 12.30 & $-$4.70 & 0.943 & 14.44 & 4.81 & $-$0.43 & 132 & 1,766 \\
\quad MiniMax-M3 & 11.34 & $-$5.66 & 0.688 & 19.31 & $-$0.07 & $-$1.00 & 137 & 3,905 \\
\quad GLM-5.3-Flash & 10.93 & $-$6.07 & 0.973 & 9.24 & 10.00 & $-$0.57 & 59 & 1,955 \\
\quad GPT-5.6-Terra & 7.82 & $-$9.18 & 0.773 & 11.27 & 7.98 & $-$0.74 & 66 & 1,955 \\
\quad Gemini-3-Flash & 13.38 & $-$3.62 & 0.835 & 16.23 & 3.02 & $-$0.90 & 48 & 1,565 \\
\addlinespace
\multicolumn{9}{@{}l}{\itshape Income (covered call)} \\
\quad DeepSeek-V4-Flash & 12.46 & $-$4.08 & 0.765 & 18.00 & 0.74 & $-$1.15 & 52 & 1,440 \\
\quad GPT-OSS-120B & 14.94 & $-$1.59 & 0.859 & 18.24 & 0.50 & $-$0.75 & 177 & 4,694 \\
\quad Qwen3-235B-A22B & 13.81 & $-$2.72 & 0.805 & 18.09 & 0.65 & $-$1.15 & 52 & 1,662 \\
\quad Llama-3.3-70B-Instruct & 14.93 & $-$1.60 & 0.832 & 18.94 & $-$0.20 & $-$0.97 & 37 & 926 \\
\quad MiniMax-M3 & 10.91 & $-$5.62 & 0.668 & 18.73 & 0.01 & $-$1.72 & 98 & 1,768 \\
\quad GLM-5.3-Flash & 13.54 & $-$2.99 & 0.827 & 17.51 & 1.23 & $-$0.95 & 123 & 1,607 \\
\quad GPT-5.6-Terra & 13.76 & $-$2.78 & 0.801 & 18.39 & 0.35 & $-$1.14 & 96 & 1,130 \\
\quad Gemini-3-Flash & 15.37 & $-$1.16 & 0.873 & 18.05 & 0.69 & $-$0.58 & 103 & 3,192 \\
\addlinespace
\multicolumn{9}{@{}l}{\textbf{(b) NVDA}\quad buy-and-hold 28.83\%, maximum drawdown 30.92\%} \\
\midrule
\multicolumn{9}{@{}l}{\itshape Hedge} \\
\quad DeepSeek-V4-Flash & 12.13 & $-$16.70 & 0.519 & 34.50 & $-$3.59 & $-$0.84 & 93 & 2,942 \\
\quad GPT-OSS-120B & 23.97 & $-$4.86 & 0.766 & 30.09 & 0.82 & $-$0.48 & 167 & 4,984 \\
\quad Qwen3-235B-A22B & 34.92 & $-$8.32 & 0.793 & 45.36 & $-$0.39 & $-$0.76 & 131 & 4,232 \\
\quad Llama-3.3-70B-Instruct & 19.89 & $-$16.14 & 0.646 & 30.97 & 7.08 & $-$1.22 & 100 & 1,693 \\
\quad MiniMax-M3 & 18.47 & $-$10.36 & 0.637 & 30.60 & 0.32 & $-$1.26 & 154 & 3,294 \\
\quad GLM-5.3-Flash & 12.34 & $-$16.49 & 0.543 & 24.27 & 6.65 & $-$1.16 & 95 & 2,597 \\
\quad GPT-5.6-Terra & 11.58 & $-$17.24 & 0.505 & 26.04 & 4.87 & $-$0.92 & 96 & 1,458 \\
\quad Gemini-3-Flash & 11.37 & $-$17.45 & 0.473 & 36.53 & $-$5.62 & $-$2.20 & 48 & 1,630 \\
\addlinespace
\multicolumn{9}{@{}l}{\itshape Income (covered call)} \\
\quad DeepSeek-V4-Flash & 32.75 & 3.92 & 0.966 & 26.25 & 4.66 & 0.20 & 42 & 1,338 \\
\quad GPT-OSS-120B & 26.56 & $-$2.27 & 0.807 & 31.84 & $-$0.93 & $-$0.33 & 165 & 4,364 \\
\quad Qwen3-235B-A22B & 31.35 & 2.52 & 0.890 & 30.18 & 0.74 & 0.32 & 94 & 2,870 \\
\quad Llama-3.3-70B-Instruct & 28.35 & $-$0.47 & 0.817 & 31.70 & $-$0.79 & 0.01 & 46 & 739 \\
\quad MiniMax-M3 & 20.87 & $-$7.96 & 0.681 & 31.55 & $-$0.63 & $-$1.02 & 73 & 1,272 \\
\quad GLM-5.3-Flash & 28.09 & $-$0.73 & 0.885 & 25.73 & 5.18 & $-$0.29 & 90 & 1,829 \\
\quad GPT-5.6-Terra & 25.08 & $-$3.75 & 0.769 & 30.28 & 0.63 & $-$0.70 & 115 & 1,180 \\
\quad Gemini-3-Flash & 22.29 & $-$6.53 & 0.721 & 29.37 & 1.55 & $-$0.88 & 64 & 2,085 \\
\addlinespace
\multicolumn{9}{@{}l}{\textbf{(c) AMD}\quad buy-and-hold 64.61\%, maximum drawdown 33.41\%} \\
\midrule
\multicolumn{9}{@{}l}{\itshape Hedge} \\
\quad DeepSeek-V4-Flash & 77.62 & 13.01 & 1.689 & 17.56 & 15.85 & 0.09 & 113 & 3,562 \\
\quad GPT-OSS-120B & 69.67 & 5.06 & 1.295 & 35.83 & $-$2.42 & 0.24 & 176 & 4,319 \\
\quad Qwen3-235B-A22B & 60.01 & $-$4.60 & 1.198 & 31.67 & 1.73 & $-$0.63 & 183 & 5,920 \\
\quad Llama-3.3-70B-Instruct & 133.49 & 4.26 & 1.362 & 52.37 & 10.78 & $-$0.51 & 104 & 1,798 \\
\quad MiniMax-M3 & 57.26 & $-$7.35 & 1.171 & 29.44 & 3.97 & $-$0.95 & 169 & 4,328 \\
\quad GLM-5.3-Flash & 49.67 & $-$14.95 & 1.137 & 24.72 & 8.68 & $-$0.81 & 81 & 2,448 \\
\quad GPT-5.6-Terra & 50.34 & $-$14.27 & 1.175 & 23.38 & 10.03 & $-$0.77 & 104 & 1,770 \\
\quad Gemini-3-Flash & 63.66 & $-$0.96 & 1.294 & 30.78 & 2.63 & $-$0.24 & 58 & 1,955 \\
\addlinespace
\multicolumn{9}{@{}l}{\itshape Income (covered call)} \\
\quad DeepSeek-V4-Flash & 68.41 & 3.80 & 1.361 & 27.90 & 5.51 & $-$0.01 & 50 & 1,326 \\
\quad GPT-OSS-120B & 45.92 & $-$18.70 & 0.985 & 32.56 & 0.84 & $-$1.57 & 169 & 3,814 \\
\quad Qwen3-235B-A22B & 52.73 & $-$11.88 & 1.068 & 35.58 & $-$2.17 & $-$0.73 & 80 & 2,490 \\
\quad Llama-3.3-70B-Instruct & 63.74 & $-$0.87 & 1.205 & 34.39 & $-$0.98 & $-$0.01 & 22 & 296 \\
\quad MiniMax-M3 & 46.59 & $-$18.02 & 1.030 & 29.62 & 3.78 & $-$1.20 & 62 & 1,416 \\
\quad GLM-5.3-Flash & 52.66 & $-$11.95 & 1.201 & 29.43 & 3.97 & $-$0.83 & 102 & 2,173 \\
\quad GPT-5.6-Terra & 38.85 & $-$25.76 & 0.928 & 30.77 & 2.64 & $-$1.25 & 102 & 1,416 \\
\quad Gemini-3-Flash & 81.11 & 16.50 & 1.497 & 30.53 & 2.88 & 0.78 & 57 & 1,858 \\
\addlinespace
\multicolumn{9}{@{}l}{\textbf{(d) GOOG}\quad buy-and-hold 53.84\%, maximum drawdown 24.87\%} \\
\midrule
\multicolumn{9}{@{}l}{\itshape Hedge} \\
\quad DeepSeek-V4-Flash & 49.96 & $-$3.89 & 1.903 & 18.22 & 6.64 & $-$0.45 & 84 & 3,255 \\
\quad GPT-OSS-120B & 47.95 & $-$5.89 & 1.702 & 23.32 & 1.55 & $-$0.93 & 164 & 4,958 \\
\quad Qwen3-235B-A22B & 49.91 & $-$3.93 & 1.756 & 23.70 & 1.17 & $-$0.91 & 175 & 5,660 \\
\quad Llama-3.3-70B-Instruct & 55.30 & $-$3.93 & 1.900 & 21.82 & 5.34 & $-$0.56 & 124 & 1,429 \\
\quad MiniMax-M3 & 46.24 & $-$7.60 & 1.722 & 20.61 & 4.25 & $-$1.09 & 153 & 4,087 \\
\quad GLM-5.3-Flash & 42.35 & $-$11.49 & 1.782 & 17.97 & 6.90 & $-$0.95 & 78 & 2,534 \\
\quad GPT-5.6-Terra & 37.38 & $-$16.46 & 1.719 & 16.73 & 8.14 & $-$1.11 & 67 & 1,939 \\
\quad Gemini-3-Flash & 48.88 & $-$4.97 & 1.830 & 19.95 & 4.92 & $-$0.61 & 32 & 1,110 \\
\addlinespace
\multicolumn{9}{@{}l}{\itshape Income (covered call)} \\
\quad DeepSeek-V4-Flash & 47.22 & $-$6.62 & 1.841 & 20.06 & 4.81 & $-$0.78 & 49 & 1,565 \\
\quad GPT-OSS-120B & 46.58 & $-$7.27 & 1.677 & 23.46 & 1.41 & $-$1.44 & 182 & 4,470 \\
\quad Qwen3-235B-A22B & 46.06 & $-$7.78 & 1.683 & 22.43 & 2.43 & $-$0.97 & 109 & 3,470 \\
\quad Llama-3.3-70B-Instruct & 55.19 & 1.35 & 1.792 & 25.10 & $-$0.24 & 0.47 & 32 & 423 \\
\quad MiniMax-M3 & 45.54 & $-$8.30 & 1.664 & 24.06 & 0.81 & $-$1.23 & 90 & 2,182 \\
\quad GLM-5.3-Flash & 40.48 & $-$13.36 & 1.691 & 19.82 & 5.05 & $-$1.36 & 98 & 1,801 \\
\quad GPT-5.6-Terra & 49.39 & $-$4.46 & 1.740 & 23.80 & 1.06 & $-$1.18 & 121 & 1,383 \\
\quad Gemini-3-Flash & 41.25 & $-$12.59 & 1.556 & 21.33 & 3.54 & $-$1.26 & 77 & 2,510 \\
\addlinespace
\multicolumn{9}{@{}l}{\textbf{(e) TSLA}\quad buy-and-hold 15.48\%, maximum drawdown 40.94\%} \\
\midrule
\multicolumn{9}{@{}l}{\itshape Hedge} \\
\quad DeepSeek-V4-Flash & $-$1.06 & $-$16.54 & 0.200 & 43.47 & $-$2.53 & $-$0.91 & 131 & 4,243 \\
\quad GPT-OSS-120B & 6.97 & $-$8.51 & 0.378 & 38.84 & 2.10 & $-$0.73 & 174 & 4,958 \\
\quad Qwen3-235B-A22B & 1.43 & $-$14.04 & 0.277 & 45.36 & $-$4.43 & $-$1.22 & 148 & 4,815 \\
\quad Llama-3.3-70B-Instruct & $-$0.23 & $-$18.80 & 0.234 & 41.86 & 6.33 & $-$1.19 & 133 & 1,842 \\
\quad MiniMax-M3 & 14.74 & $-$0.73 & 0.533 & 35.59 & 5.35 & $-$0.28 & 181 & 4,152 \\
\quad GLM-5.3-Flash & $-$0.65 & $-$16.13 & 0.172 & 31.51 & 9.43 & $-$1.07 & 107 & 2,329 \\
\quad GPT-5.6-Terra & 4.35 & $-$11.12 & 0.292 & 25.99 & 14.95 & $-$0.71 & 123 & 1,988 \\
\quad Gemini-3-Flash & 11.48 & $-$3.99 & 0.470 & 36.93 & 4.00 & $-$0.41 & 47 & 1,565 \\
\addlinespace
\multicolumn{9}{@{}l}{\itshape Income (covered call)} \\
\quad DeepSeek-V4-Flash & 20.13 & 4.66 & 0.640 & 37.42 & 3.52 & 0.09 & 36 & 1,012 \\
\quad GPT-OSS-120B & 8.40 & $-$7.07 & 0.420 & 44.86 & $-$3.93 & $-$0.42 & 165 & 4,392 \\
\quad Qwen3-235B-A22B & 21.38 & 5.90 & 0.637 & 42.15 & $-$1.21 & 0.57 & 74 & 2,259 \\
\quad Llama-3.3-70B-Instruct & 10.24 & $-$5.24 & 0.450 & 43.23 & $-$2.30 & $-$0.36 & 43 & 801 \\
\quad MiniMax-M3 & 8.61 & $-$6.87 & 0.417 & 40.58 & 0.36 & $-$0.69 & 90 & 1,913 \\
\quad GLM-5.3-Flash & 12.46 & $-$3.01 & 0.490 & 39.69 & 1.25 & $-$0.42 & 58 & 1,656 \\
\quad GPT-5.6-Terra & 10.94 & $-$4.54 & 0.459 & 41.56 & $-$0.62 & $-$0.54 & 112 & 1,338 \\
\quad Gemini-3-Flash & 16.39 & 0.92 & 0.553 & 42.20 & $-$1.26 & 0.12 & 74 & 2,540 \\
\end{longtable}
\normalsize


The following fill-level turnover and median-tenor description retains its
six-provider-model scope; the per-asset performance and stock-exposure
analyses include all eight models.

Trading frequency and tenor choice vary more across models than across mandates. Mean option fills per month range from $13.7$ for the least active model--mandate cell to $71.0$ for the most active. The direction of the mandate effect is not shared either: the Hedge mandate is the busier of the two for four of six models, by up to $3.6$ times, while GPT-OSS-120B and GLM-5.3-Flash trade both mandates at nearly the same rate. Median entry maturity spans $2.5$ to $28$ days, so the panel contains both weekly-rolling and monthly-tenor policies, and only $20.2\%$ of all overlay fills are shorter than a week. Five of six models sit inside the prescribed $7$--$45$ day (Income) and $7$--$60$ day (Hedge) bands at the median, and only Llama-3.3-70B-Instruct writes income legs shorter than a week on average. Tenor policy is therefore a model property in this record rather than a shared short-maturity bias, and the near-continuous rolling that the earlier panel showed is confined to the short-tenor models.

Exposure to the rolling penalty has to be read off the stock leg rather
than off assignment records, which the runs do not keep: the engine models
assignment and exercise but never emits an event for either, so no
assignment count can be recovered from the traces.
The stock-leg record gives the closest available
evidence, and it is nearly uniform. Every one of the $80$ overlay cases ends
the year holding exactly its target share count, and $76$ of them are below
target on at most $8$ of the $250$ trading days. That floor is an artefact
rather than behaviour: every account opens flat and buys the stock leg on the
first day, so one day per asset is the opening ramp. Only four cases depart
from it, and all four belong to one model.

The exception is Llama-3.3-70B-Instruct under the hedge mandate, whose
stock leg is below target on $58$, $144$, $164$ and $246$ of the $250$
days on AMD, GOOG, NVDA and TSLA, with a deepest shortfall on each equal
to the entire target position, $20{,}000$, $11{,}000$, $12{,}500$ and
$12{,}000$ shares respectively. Its hedged-equity return on those four
assets is therefore earned with the equity leg absent or incomplete for
much of the year rather than against the mandated exposure, which is the
context in which its $\Delta\text{Ret}\%$ and $\Delta\text{MDD}$ on those
assets should be read. The extraction records how many days the leg was
short and how deep the worst shortfall ran, but not the shape in between,
so the stronger claim that the leg was flat throughout is not supported.
Overlay exposure drift in this record is thus a property of one
(model, mandate) pair rather than of the panel.

\paragraph{Uncertainty on the overlay Sharpe ratios.}
Overlay evaluation yields one continuous equity path per cell rather than a set
of independent episodes, so the episode bootstrap used for the other two tasks
does not apply and daily overlay returns are serially dependent. We therefore
interval the Sharpe ratios of Table~\ref{tab:overlay_per_asset} with a
stationary block bootstrap, resampling blocks of geometrically distributed
length with a mean of ten trading days, which is longer than the Mon+Wed
rebalance cycle and so keeps whole cycles intact within a block.

The result is a caution about how the Sharpe column should be read. All $80$
cells carry a positive point estimate, from $0.17$ to $1.90$, which follows
from holding a long equity base through a year in which all five underlyings
rose. Only one interval excludes zero, DeepSeek-V4-Flash hedging GOOG at
$1.90$ with $[0.09, 3.81]$, and the median interval spans $3.72$ units of
Sharpe, far wider than any difference between models. One year of daily data
on five assets does not identify a risk-adjusted edge in either direction, and
we draw no cross-model overlay conclusion from these ratios; the active-return
and drawdown columns, which are differences against a matched benchmark on the
same path, carry the overlay findings instead.

\paragraph{Mechanical baselines.}
Outcome numbers alone cannot say whether an agent's overlay reflects option
reasoning or exposure that a fixed rule would have captured just as well. We
therefore add two deterministic references, a covered call and a protective
put, each writing or buying the contract nearest $0.30$ delta inside the
mandate's DTE band on every rebalance day. They run the same overlay
configurations as the agents, with the identical universe, capital, dates,
Mon+Wed rebalance cadence, risk rules and fill model; only the decision layer
differs, so the comparison isolates the decision.

Table~\ref{tab:overlay_baselines} gives the result, and the two mandates
answer the question differently. Under the covered call the agents are clearly
ahead, by $17.6$ percentage points of active return on average and on four of
the five assets. The rule's losses concentrate exactly where the underlying
rallied hardest, $-47.0$ on AMD and $-47.5$ on GOOG against underlyings up
$64.6\%$ and $53.8\%$, because writing a $0.30$-delta call on every rebalance
day sells precisely the upside that then arrives; the agents wrote less, and
in this market that was the better choice. Under the hedge mandate the same
comparison is close to a tie, $+0.02$ points on average and ahead on two of
five assets, and the rule reduces drawdown roughly twice as much, $+8.20$
against $+4.21$. The systematic put buyer pays for protection and gets it; the
agents pay less and are protected less.

Two qualifications belong with this table. Neither mandate's panel-average
active return is positive: beating a fixed overlay rule does not establish
an advantage over passive buy-and-hold. The panel mean under the hedge mandate
includes the Llama-3.3-70B-Instruct cells whose equity leg is documented above
as incomplete, which depresses it; the covered-call comparison, where the
margin is large, is unaffected by that.

\begin{table}[t]
\centering
\small
\setlength{\tabcolsep}{5pt}
\caption{Mechanical overlay baselines against the LLM panel. Each baseline runs
the paper's own overlay configuration with only the decision layer replaced by a
fixed $0.30$-delta rule, so universe, capital, dates, rebalance cadence, risk
rules and fill model are identical to the agent runs. Cells are active return
against the matched buy-and-hold in percentage points; \emph{$\Delta$Ret} and
\emph{$\Delta$MDD} average over the five assets, and the last row of each block
is the panel mean minus the rule on the same asset, with the count of assets on
which the panel is ahead.}
\label{tab:overlay_baselines}
\begin{tabular}{l rrrrr rr}
\toprule
 & QQQ & NVDA & AMD & GOOG & TSLA & $\Delta$Ret & $\Delta$MDD \\
\midrule
\multicolumn{5}{@{}l}{\textit{Hedging}} \\[1pt]
Protective put rule & -9.1 & -11.6 & +3.2 & -7.1 & -15.2 & -7.97 & +8.20 \\
LLM panel mean & -5.3 & -13.4 & -2.5 & -7.3 & -11.2 & -7.94 & +4.21 \\
\quad\textit{panel $-$ rule} & +3.8 & -1.8 & -5.7 & -0.1 & +3.9 & +0.02 & \multicolumn{1}{c}{2/5} \\
\midrule
\multicolumn{5}{@{}l}{\textit{Covered Call}} \\[1pt]
Covered call rule & -7.7 & -16.8 & -47.0 & -47.5 & +8.8 & -22.04 & +3.57 \\
LLM panel mean & -2.8 & -1.9 & -8.4 & -7.4 & -1.9 & -4.47 & +1.14 \\
\quad\textit{panel $-$ rule} & +4.9 & +14.9 & +38.7 & +40.1 & -10.7 & +17.56 & \multicolumn{1}{c}{4/5} \\
\bottomrule
\end{tabular}
\end{table}

\clearpage

\subsubsection{0DTE Intraday}

The month-by-month results in Table~\ref{tab:intraday_monthly_spy} show
dispersion without a stable ranking. Summed over the $102$ sessions,
GPT-OSS-120B, MiniMax-M3 and GLM-5.3-Flash end the year positive at
$+14.4$, $+14.4$ and $+11.0$ percentage points, while DeepSeek-V4-Flash,
Llama-3.3-70B-Instruct and Qwen3-235B-A22B end negative at $-74.6$,
$-60.6$ and $-37.1$ points. GPT-5.6-Terra and Gemini-3-Flash also end
negative, at $-18.4$ and $-11.3$ points, with five positive months each.
Daily win rates span $25.5\%$ to $47.1\%$, so the spread in annual totals also reflects tail months
rather than from hit rate: the best single month is GLM-5.3-Flash's
$+54.5$ points in April and the worst is GPT-OSS-120B's $-31.6$ points in
May. Only GPT-OSS-120B is positive in a majority of months, $7$ of $12$;
the others are positive in $3$ to $5$. No month is good for the whole
panel and none is bad for all of it, which is the opposite of the clean
temporal pattern the earlier five-model panel appeared to show.

\vspace{12pt}

\setlength{\LTleft}{0pt}
\setlength{\LTright}{0pt}
\renewcommand{\arraystretch}{1.05}
\setlength{\tabcolsep}{4pt}
\footnotesize
\begin{longtable}{@{\extracolsep{\fill}} l l r r r r r r r r @{}}
\caption{Per-month, per-model intraday 0-DTE SPY performance. Each session is an independent fresh-\$10k episode; PnL\% sums the daily realized return percentages inside the month and \textit{Overall} sums every covered session in 2025. Mean\%, Win\%, P/L, PF, Best\% and Worst\% are computed over the sessions inside each month.}\label{tab:intraday_monthly_spy} \\
\toprule
\textbf{Month} & \textbf{Model} & \textbf{Sess.} & \textbf{Win\%} & \textbf{PnL\%} & \textbf{Mean\%} & \textbf{P/L} & \textbf{PF} & \textbf{Best\%} & \textbf{Worst\%} \\
\midrule
\endfirsthead
\multicolumn{10}{l}{\footnotesize\itshape (continued from previous page)} \\
\toprule
\textbf{Month} & \textbf{Model} & \textbf{Sess.} & \textbf{Win\%} & \textbf{PnL\%} & \textbf{Mean\%} & \textbf{P/L} & \textbf{PF} & \textbf{Best\%} & \textbf{Worst\%} \\
\midrule
\endhead
\midrule
\multicolumn{10}{r}{\footnotesize\itshape continued on next page} \\
\endfoot
\bottomrule
\endlastfoot
\multirow{8}{*}{2025-01} & DeepSeek-V4-Flash & 9 & 33.3 & $-$10.97 & $-$1.22 & 0.45 & 0.23 & +2.07 & $-$4.59 \\*
 & GPT-OSS-120B & 9 & 33.3 & +13.55 & +1.50 & 3.14 & 1.57 & +21.87 & $-$7.93 \\*
 & Qwen3-235B-A22B & 9 & 33.3 & $-$5.29 & $-$0.59 & 0.94 & 0.47 & +3.03 & $-$2.75 \\*
 & Llama-3.3-70B-Instruct & 9 & 11.1 & $-$19.69 & $-$2.19 & 1.01 & 0.13 & +2.86 & $-$5.50 \\*
 & MiniMax-M3 & 9 & 44.4 & $-$1.87 & $-$0.21 & 0.91 & 0.73 & +2.49 & $-$3.72 \\*
 & GLM-5.3-Flash & 9 & 55.6 & +3.17 & +0.35 & 1.10 & 1.37 & +5.06 & $-$3.82 \\*
 & GPT-5.6-Terra & 9 & 33.3 & $-$1.10 & $-$0.12 & 1.66 & 0.83 & +3.89 & $-$3.12 \\*
 & Gemini-3-Flash & 9 & 55.6 & $-$0.61 & $-$0.07 & 0.74 & 0.92 & +4.40 & $-$3.84 \\
\midrule
\multirow{8}{*}{2025-02} & DeepSeek-V4-Flash & 8 & 25.0 & +6.16 & +0.77 & 4.56 & 1.52 & +16.17 & $-$3.01 \\*
 & GPT-OSS-120B & 8 & 37.5 & +13.73 & +1.72 & 3.09 & 1.86 & +22.55 & $-$4.27 \\*
 & Qwen3-235B-A22B & 8 & 12.5 & $-$16.51 & $-$2.06 & 0.87 & 0.12 & +2.33 & $-$7.42 \\*
 & Llama-3.3-70B-Instruct & 8 & 37.5 & +11.28 & +1.41 & 3.24 & 1.95 & +18.87 & $-$5.19 \\*
 & MiniMax-M3 & 8 & 37.5 & $-$2.14 & $-$0.27 & 1.24 & 0.74 & +3.38 & $-$3.17 \\*
 & GLM-5.3-Flash & 8 & 25.0 & $-$3.37 & $-$0.42 & 2.07 & 0.69 & +7.16 & $-$3.29 \\*
 & GPT-5.6-Terra & 8 & 37.5 & +6.85 & +0.86 & 2.84 & 1.70 & +14.28 & $-$3.64 \\*
 & Gemini-3-Flash & 8 & 37.5 & +2.40 & +0.30 & 2.05 & 1.23 & +6.26 & $-$5.45 \\
\midrule
\multirow{8}{*}{2025-03} & DeepSeek-V4-Flash & 8 & 25.0 & $-$7.29 & $-$0.91 & 1.95 & 0.65 & +9.81 & $-$6.74 \\*
 & GPT-OSS-120B & 8 & 37.5 & +25.93 & +3.24 & 4.17 & 2.50 & +20.12 & $-$5.70 \\*
 & Qwen3-235B-A22B & 8 & 50.0 & $-$2.58 & $-$0.32 & 0.77 & 0.77 & +4.34 & $-$5.10 \\*
 & Llama-3.3-70B-Instruct & 8 & 25.0 & $-$11.61 & $-$1.45 & 1.19 & 0.40 & +3.92 & $-$3.86 \\*
 & MiniMax-M3 & 8 & 25.0 & $-$1.05 & $-$0.13 & 2.66 & 0.89 & +4.33 & $-$3.60 \\*
 & GLM-5.3-Flash & 8 & 25.0 & $-$11.86 & $-$1.48 & 1.07 & 0.36 & +3.98 & $-$4.31 \\*
 & GPT-5.6-Terra & 8 & 50.0 & +8.87 & +1.11 & 1.59 & 1.59 & +14.02 & $-$4.79 \\*
 & Gemini-3-Flash & 8 & 50.0 & +8.30 & +1.04 & 2.16 & 2.16 & +6.31 & $-$2.82 \\
\midrule
\multirow{8}{*}{2025-04} & DeepSeek-V4-Flash & 8 & 12.5 & $-$9.67 & $-$1.21 & 4.02 & 0.57 & +13.02 & $-$3.57 \\*
 & GPT-OSS-120B & 8 & 25.0 & +15.10 & +1.89 & 4.91 & 1.64 & +23.05 & $-$5.48 \\*
 & Qwen3-235B-A22B & 8 & 25.0 & $-$3.09 & $-$0.39 & 2.27 & 0.76 & +8.46 & $-$3.87 \\*
 & Llama-3.3-70B-Instruct & 8 & 12.5 & $-$9.11 & $-$1.14 & 3.44 & 0.49 & +8.81 & $-$3.71 \\*
 & MiniMax-M3 & 8 & 75.0 & +25.51 & +3.19 & 2.62 & 7.85 & +16.82 & $-$3.18 \\*
 & GLM-5.3-Flash & 8 & 50.0 & +54.51 & +6.81 & 7.35 & 7.35 & +28.81 & $-$2.97 \\*
 & GPT-5.6-Terra & 8 & 50.0 & +17.71 & +2.21 & 2.38 & 2.38 & +12.50 & $-$3.59 \\*
 & Gemini-3-Flash & 8 & 75.0 & +11.51 & +1.44 & 1.29 & 3.87 & +10.11 & $-$3.21 \\
\midrule
\multirow{8}{*}{2025-05} & DeepSeek-V4-Flash & 9 & 22.2 & $-$7.85 & $-$0.87 & 2.10 & 0.60 & +7.99 & $-$3.17 \\*
 & GPT-OSS-120B & 9 & 11.1 & $-$31.58 & $-$3.51 & 0.56 & 0.07 & +2.37 & $-$6.87 \\*
 & Qwen3-235B-A22B & 9 & 22.2 & $-$8.59 & $-$0.95 & 1.13 & 0.32 & +2.10 & $-$3.47 \\*
 & Llama-3.3-70B-Instruct & 9 & 11.1 & $-$21.50 & $-$2.39 & 1.00 & 0.12 & +3.06 & $-$5.20 \\*
 & MiniMax-M3 & 9 & 0.0 & $-$12.95 & $-$1.44 & -- & 0.00 & $-$0.24 & $-$3.31 \\*
 & GLM-5.3-Flash & 9 & 22.2 & $-$5.28 & $-$0.59 & 2.30 & 0.66 & +5.21 & $-$3.01 \\*
 & GPT-5.6-Terra & 9 & 22.2 & $-$11.36 & $-$1.26 & 1.17 & 0.33 & +4.03 & $-$3.22 \\*
 & Gemini-3-Flash & 9 & 11.1 & $-$21.13 & $-$2.35 & 1.25 & 0.16 & +3.90 & $-$6.63 \\
\midrule
\multirow{8}{*}{2025-06} & DeepSeek-V4-Flash & 8 & 25.0 & $-$5.59 & $-$0.70 & 1.17 & 0.39 & +1.81 & $-$3.65 \\*
 & GPT-OSS-120B & 8 & 12.5 & $-$20.50 & $-$2.56 & 1.42 & 0.20 & +5.23 & $-$4.44 \\*
 & Qwen3-235B-A22B & 8 & 25.0 & $-$4.70 & $-$0.59 & 0.60 & 0.20 & +0.99 & $-$1.88 \\*
 & Llama-3.3-70B-Instruct & 8 & 25.0 & $-$9.29 & $-$1.16 & 1.15 & 0.38 & +5.13 & $-$4.69 \\*
 & MiniMax-M3 & 8 & 37.5 & +4.38 & +0.55 & 3.11 & 1.86 & +3.19 & $-$1.90 \\*
 & GLM-5.3-Flash & 8 & 37.5 & $-$7.30 & $-$0.91 & 0.65 & 0.39 & +2.25 & $-$3.03 \\*
 & GPT-5.6-Terra & 8 & 12.5 & $-$15.03 & $-$1.88 & 0.82 & 0.12 & +2.00 & $-$3.23 \\*
 & Gemini-3-Flash & 8 & 25.0 & $-$12.10 & $-$1.51 & 0.27 & 0.09 & +1.04 & $-$4.16 \\
\midrule
\multirow{8}{*}{2025-07} & DeepSeek-V4-Flash & 8 & 50.0 & $-$6.28 & $-$0.79 & 0.43 & 0.43 & +2.99 & $-$3.34 \\*
 & GPT-OSS-120B & 8 & 50.0 & +0.47 & +0.06 & 1.03 & 1.03 & +9.07 & $-$3.87 \\*
 & Qwen3-235B-A22B & 8 & 37.5 & +2.41 & +0.30 & 2.61 & 1.57 & +4.33 & $-$1.15 \\*
 & Llama-3.3-70B-Instruct & 8 & 50.0 & +18.16 & +2.27 & 3.16 & 3.16 & +9.51 & $-$3.31 \\*
 & MiniMax-M3 & 8 & 37.5 & +0.51 & +0.06 & 1.83 & 1.10 & +4.36 & $-$2.26 \\*
 & GLM-5.3-Flash & 8 & 50.0 & $-$1.76 & $-$0.22 & 0.75 & 0.75 & +2.26 & $-$2.81 \\*
 & GPT-5.6-Terra & 8 & 25.0 & $-$11.21 & $-$1.40 & 0.30 & 0.10 & +0.67 & $-$3.02 \\*
 & Gemini-3-Flash & 8 & 62.5 & +1.07 & +0.13 & 0.68 & 1.14 & +4.29 & $-$6.65 \\
\midrule
\multirow{8}{*}{2025-08} & DeepSeek-V4-Flash & 9 & 11.1 & $-$18.17 & $-$2.02 & 0.71 & 0.09 & +1.78 & $-$5.21 \\*
 & GPT-OSS-120B & 9 & 11.1 & $-$23.36 & $-$2.60 & 1.13 & 0.14 & +3.83 & $-$6.76 \\*
 & Qwen3-235B-A22B & 9 & 55.6 & +4.68 & +0.52 & 1.92 & 2.40 & +2.45 & $-$1.37 \\*
 & Llama-3.3-70B-Instruct & 9 & 33.3 & $-$8.18 & $-$0.91 & 1.13 & 0.56 & +5.35 & $-$6.31 \\*
 & MiniMax-M3 & 9 & 22.2 & $-$5.03 & $-$0.56 & 0.66 & 0.19 & +1.06 & $-$2.19 \\*
 & GLM-5.3-Flash & 9 & 33.3 & $-$9.07 & $-$1.01 & 0.69 & 0.34 & +4.29 & $-$3.06 \\*
 & GPT-5.6-Terra & 9 & 22.2 & $-$9.69 & $-$1.08 & 0.97 & 0.28 & +3.20 & $-$3.31 \\*
 & Gemini-3-Flash & 9 & 44.4 & $-$6.31 & $-$0.70 & 0.49 & 0.39 & +2.29 & $-$4.63 \\
\midrule
\multirow{8}{*}{2025-09} & DeepSeek-V4-Flash & 8 & 25.0 & $-$12.90 & $-$1.61 & 0.12 & 0.04 & +0.33 & $-$3.28 \\*
 & GPT-OSS-120B & 8 & 12.5 & $-$23.63 & $-$2.95 & 2.03 & 0.29 & +9.62 & $-$10.07 \\*
 & Qwen3-235B-A22B & 8 & 62.5 & +1.66 & +0.21 & 0.96 & 1.59 & +1.67 & $-$2.50 \\*
 & Llama-3.3-70B-Instruct & 8 & 25.0 & $-$12.80 & $-$1.60 & 0.54 & 0.18 & +2.72 & $-$4.75 \\*
 & MiniMax-M3 & 8 & 25.0 & $-$3.30 & $-$0.41 & 1.23 & 0.41 & +1.47 & $-$2.56 \\*
 & GLM-5.3-Flash & 8 & 25.0 & $-$11.27 & $-$1.41 & 0.76 & 0.25 & +2.62 & $-$3.92 \\*
 & GPT-5.6-Terra & 8 & 37.5 & +0.79 & +0.10 & 1.82 & 1.09 & +4.79 & $-$3.21 \\*
 & Gemini-3-Flash & 8 & 50.0 & $-$2.78 & $-$0.35 & 0.70 & 0.70 & +3.77 & $-$4.40 \\
\midrule
\multirow{8}{*}{2025-10} & DeepSeek-V4-Flash & 10 & 50.0 & +3.24 & +0.32 & 1.22 & 1.22 & +8.71 & $-$4.03 \\*
 & GPT-OSS-120B & 10 & 60.0 & +43.95 & +4.39 & 2.53 & 3.80 & +19.32 & $-$4.69 \\*
 & Qwen3-235B-A22B & 10 & 50.0 & $-$1.54 & $-$0.15 & 0.85 & 0.85 & +5.92 & $-$3.69 \\*
 & Llama-3.3-70B-Instruct & 10 & 20.0 & $-$2.25 & $-$0.23 & 3.61 & 0.90 & +13.20 & $-$3.59 \\*
 & MiniMax-M3 & 10 & 70.0 & +10.91 & +1.09 & 1.38 & 3.23 & +9.73 & $-$3.46 \\*
 & GLM-5.3-Flash & 10 & 40.0 & +5.89 & +0.59 & 2.10 & 1.40 & +9.74 & $-$3.38 \\*
 & GPT-5.6-Terra & 10 & 60.0 & +6.94 & +0.69 & 1.25 & 1.87 & +5.77 & $-$3.08 \\*
 & Gemini-3-Flash & 10 & 50.0 & $-$5.02 & $-$0.50 & 0.60 & 0.60 & +3.07 & $-$4.53 \\
\midrule
\multirow{8}{*}{2025-11} & DeepSeek-V4-Flash & 8 & 37.5 & +5.71 & +0.71 & 2.70 & 1.62 & +7.05 & $-$3.37 \\*
 & GPT-OSS-120B & 8 & 50.0 & +5.36 & +0.67 & 1.26 & 1.26 & +9.81 & $-$10.10 \\*
 & Qwen3-235B-A22B & 8 & 37.5 & +0.41 & +0.05 & 1.75 & 1.05 & +6.45 & $-$3.05 \\*
 & Llama-3.3-70B-Instruct & 8 & 25.0 & $-$7.46 & $-$0.93 & 1.48 & 0.49 & +6.89 & $-$3.99 \\*
 & MiniMax-M3 & 8 & 62.5 & +1.86 & +0.23 & 0.92 & 1.54 & +3.08 & $-$1.37 \\*
 & GLM-5.3-Flash & 8 & 12.5 & +4.93 & +0.62 & 9.12 & 1.30 & +21.24 & $-$3.28 \\*
 & GPT-5.6-Terra & 8 & 25.0 & $-$10.04 & $-$1.25 & 1.25 & 0.42 & +6.76 & $-$6.11 \\*
 & Gemini-3-Flash & 8 & 75.0 & +17.49 & +2.19 & 1.21 & 3.64 & +8.33 & $-$4.15 \\
\midrule
\multirow{8}{*}{2025-12} & DeepSeek-V4-Flash & 9 & 33.3 & $-$10.95 & $-$1.22 & 0.60 & 0.30 & +3.60 & $-$3.08 \\*
 & GPT-OSS-120B & 9 & 44.4 & $-$4.65 & $-$0.52 & 0.96 & 0.77 & +6.32 & $-$7.15 \\*
 & Qwen3-235B-A22B & 9 & 22.2 & $-$3.92 & $-$0.43 & 2.46 & 0.70 & +7.54 & $-$3.21 \\*
 & Llama-3.3-70B-Instruct & 9 & 33.3 & +11.85 & +1.32 & 3.79 & 1.89 & +17.52 & $-$4.37 \\*
 & MiniMax-M3 & 9 & 33.3 & $-$2.47 & $-$0.28 & 1.51 & 0.76 & +4.99 & $-$3.71 \\*
 & GLM-5.3-Flash & 9 & 22.2 & $-$7.64 & $-$0.85 & 1.36 & 0.39 & +3.17 & $-$4.12 \\*
 & GPT-5.6-Terra & 9 & 55.6 & $-$1.12 & $-$0.12 & 0.70 & 0.87 & +2.32 & $-$3.58 \\*
 & Gemini-3-Flash & 9 & 33.3 & $-$4.11 & $-$0.46 & 1.30 & 0.65 & +5.03 & $-$2.91 \\
\midrule
\multirow{8}{*}{\textbf{Overall}} & DeepSeek-V4-Flash & 102 & 29.4 & $-$74.56 & $-$0.73 & 1.42 & 0.59 & +16.17 & $-$6.74 \\*
 & GPT-OSS-120B & 102 & 32.4 & +14.39 & +0.14 & 2.20 & 1.05 & +23.05 & $-$10.10 \\*
 & Qwen3-235B-A22B & 102 & 36.3 & $-$37.05 & $-$0.36 & 1.19 & 0.67 & +8.46 & $-$7.42 \\*
 & Llama-3.3-70B-Instruct & 102 & 25.5 & $-$60.59 & $-$0.59 & 2.06 & 0.70 & +18.87 & $-$6.31 \\*
 & MiniMax-M3 & 102 & 39.2 & +14.35 & +0.14 & 1.82 & 1.18 & +16.82 & $-$3.72 \\*
 & GLM-5.3-Flash & 102 & 33.3 & +10.96 & +0.11 & 2.14 & 1.07 & +28.81 & $-$4.31 \\*
 & GPT-5.6-Terra & 102 & 36.3 & $-$18.39 & $-$0.18 & 1.54 & 0.87 & +14.28 & $-$6.11 \\*
 & Gemini-3-Flash & 102 & 47.1 & $-$11.29 & $-$0.11 & 1.02 & 0.91 & +10.11 & $-$6.65 \\
\end{longtable}
\normalsize

Table~\ref{tab:intraday_strategy_spy} disaggregates the results by the
strategy bucket assigned at the first fill. Long calls are the clearest
loss centre, $-21.4\text{k}$ dollars pooled across the eight-model panel, against
$+5.5\text{k}$ for long puts, $+3.6\text{k}$ for short puts and
$-3.9\text{k}$ for short calls. The put-over-call asymmetry of the earlier
panel therefore survives in sign but not in magnitude, and no bucket
dominates for every model: long puts are the best bucket for GPT-OSS-120B
and GLM-5.3-Flash as well as both additional models, short puts for DeepSeek-V4-Flash and Qwen3-235B-A22B,
short calls for Llama-3.3-70B-Instruct and long calls for MiniMax-M3. Any
reading of a single structural bias in the 2025 SPY 0DTE tape has to
account for that disagreement.

The \emph{Worst\%} column carries the familiar denominator artefact of
short premium. Because the per-trade return denominator is the sale
proceeds, an unhedged short whose closing buyback exceeds the original
premium produces returns well below $-100\%$; such figures are arithmetic
artefacts of small premia closed at large prices on an adverse move, not
literal portfolio losses of that size. They do mark the buckets in which a
single mistimed trade can dominate a model's PnL, and they argue for
treating the short buckets' headline win rates with caution.

\begin{figure}[h]
    \centering
    \includegraphics[width=0.99\linewidth]{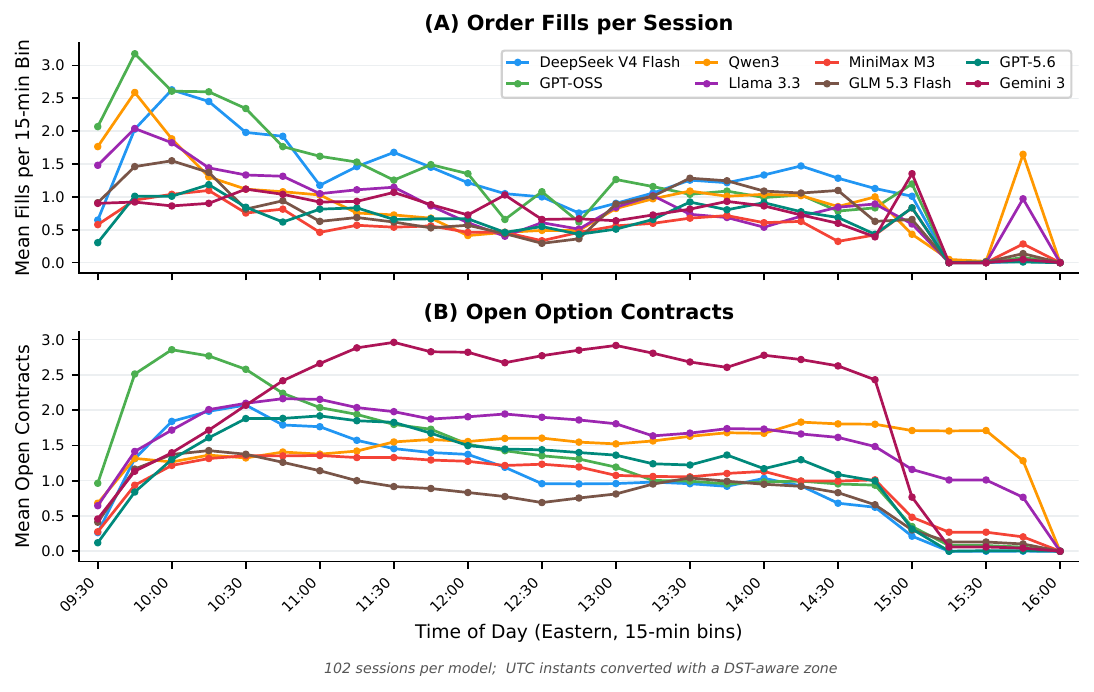}
    \caption{Per-model intraday activity profile for 0DTE SPY, in Eastern time. \textbf{(A)} Mean number of order fills per session in each 15-minute bin. \textbf{(B)} Mean number of open option contracts, averaged over every per-step portfolio snapshot falling in the bin. Recorded timestamps are UTC instants and the sessions track daylight saving, 14:30Z under EST and 13:30Z under EDT; converting with a DST-aware zone places all 102 sessions at 09:30--16:00 ET, and no fill falls outside that window for any model. A fixed offset would misplace the 66 EDT sessions by one hour.}
    \label{fig:intraday_activity_profile}
\end{figure}

\setlength{\LTleft}{0pt}
\setlength{\LTright}{0pt}
\renewcommand{\arraystretch}{1.05}
\setlength{\tabcolsep}{1.5pt}
\footnotesize
\begin{longtable}{@{\extracolsep{\fill}} l l r r r r r r r r r @{}}
\caption{Per-strategy-bucket, per-model intraday 0-DTE SPY performance. Each contract traded on each session is one trade, and the bucket is set by the first fill's side and the option right (Long Call = first BUY of a call, Short Put = first SELL of a put, and so on). PnL is FIFO-matched realized PnL net of fees, with any residual at end of day settled to zero at expiry, as in the recorded episodes. Per-trade return uses the entry-side notional as denominator, so \emph{Worst\%} can fall well below $-100$\% for short trades whose closing buyback dwarfs the premium received. Buy and Sell are total notionals (qty $\times$ price $\times$ 100) in thousands of USD.}\label{tab:intraday_strategy_spy} \\
\toprule
\textbf{Strategy} & \textbf{Model} & \textbf{Trades} & \textbf{Win\%} & \textbf{PnL (k\$)} & \textbf{P/L} & \textbf{PF} & \textbf{Best\%} & \textbf{Worst\%} & \textbf{Buy (k\$)} & \textbf{Sell (k\$)} \\
\midrule
\endfirsthead
\multicolumn{11}{l}{\footnotesize\itshape (continued from previous page)} \\
\toprule
\textbf{Strategy} & \textbf{Model} & \textbf{Trades} & \textbf{Win\%} & \textbf{PnL (k\$)} & \textbf{P/L} & \textbf{PF} & \textbf{Best\%} & \textbf{Worst\%} & \textbf{Buy (k\$)} & \textbf{Sell (k\$)} \\
\midrule
\endhead
\midrule
\multicolumn{11}{r}{\footnotesize\itshape continued on next page} \\
\endfoot
\bottomrule
\endlastfoot
\multirow{8}{*}{Long Call} & DeepSeek-V4-Flash & 384 & 23.4 & $-$5.9 & 1.98 & 0.61 & +100.2 & $-$78.4 & 83.3 & 78.7 \\*
 & GPT-OSS-120B & 296 & 30.1 & $-$5.1 & 1.80 & 0.78 & +130.8 & $-$101.0 & 99.9 & 96.0 \\*
 & Qwen3-235B-A22B & 211 & 36.0 & $-$2.1 & 1.34 & 0.75 & +203.7 & $-$223.9 & 47.3 & 45.9 \\*
 & Llama-3.3-70B-Instruct & 210 & 20.9 & $-$5.8 & 1.89 & 0.50 & +162.6 & $-$107.5 & 43.7 & 38.4 \\*
 & MiniMax-M3 & 200 & 37.0 & +1.3 & 2.04 & 1.20 & +499.3 & $-$100.3 & 31.2 & 32.8 \\*
 & GLM-5.3-Flash & 282 & 29.8 & $-$0.8 & 2.23 & 0.94 & +272.2 & $-$69.8 & 66.3 & 66.4 \\*
 & GPT-5.6-Terra & 231 & 35.5 & $-$2.4 & 1.42 & 0.78 & +227.7 & $-$73.4 & 46.4 & 44.5 \\*
 & Gemini-3-Flash & 240 & 50.0 & $-$0.7 & 0.96 & 0.96 & +144.2 & $-$96.1 & 68.1 & 68.0 \\
\midrule
\multirow{8}{*}{Long Put} & DeepSeek-V4-Flash & 363 & 36.1 & $-$1.6 & 1.60 & 0.90 & +292.0 & $-$68.1 & 96.7 & 96.2 \\*
 & GPT-OSS-120B & 268 & 36.9 & +2.8 & 1.92 & 1.12 & +605.5 & $-$90.9 & 91.6 & 95.4 \\*
 & Qwen3-235B-A22B & 306 & 38.6 & $-$1.4 & 1.45 & 0.91 & +443.2 & $-$510.2 & 101.5 & 100.8 \\*
 & Llama-3.3-70B-Instruct & 373 & 33.2 & +0.1 & 2.02 & 1.01 & +1887.0 & $-$104.3 & 120.6 & 121.8 \\*
 & MiniMax-M3 & 271 & 38.4 & +1.3 & 1.79 & 1.11 & +365.2 & $-$100.2 & 46.4 & 48.1 \\*
 & GLM-5.3-Flash & 293 & 35.5 & +2.3 & 2.17 & 1.19 & +201.9 & $-$101.2 & 58.1 & 61.0 \\*
 & GPT-5.6-Terra & 286 & 39.5 & +0.8 & 1.61 & 1.05 & +238.9 & $-$79.1 & 66.8 & 68.3 \\*
 & Gemini-3-Flash & 199 & 42.7 & +1.2 & 1.50 & 1.12 & +216.7 & $-$104.0 & 39.3 & 40.9 \\
\midrule
\multirow{8}{*}{Short Call} & DeepSeek-V4-Flash & 32 & 46.9 & $-$0.2 & 0.76 & 0.67 & +58.9 & $-$684.4 & 3.3 & 3.2 \\*
 & GPT-OSS-120B & 89 & 66.3 & +1.7 & 1.05 & 2.06 & +3826.5 & $-$541.9 & 11.2 & 13.3 \\*
 & Qwen3-235B-A22B & 72 & 52.8 & $-$0.5 & 0.53 & 0.59 & +90.2 & $-$1694.5 & 7.0 & 6.7 \\*
 & Llama-3.3-70B-Instruct & 17 & 58.8 & +0.2 & 1.46 & 2.08 & +89.1 & $-$49.8 & 1.1 & 1.4 \\*
 & MiniMax-M3 & 39 & 59.0 & $-$1.7 & 0.19 & 0.28 & +85.6 & $-$967.9 & 4.5 & 2.8 \\*
 & GLM-5.3-Flash & 17 & 52.9 & $-$0.5 & 0.17 & 0.20 & +85.3 & $-$330.4 & 1.8 & 1.3 \\*
 & GPT-5.6-Terra & 8 & 50.0 & $-$0.6 & 0.36 & 0.36 & +70.6 & $-$384.6 & 2.5 & 1.8 \\*
 & Gemini-3-Flash & 120 & 52.5 & $-$2.2 & 0.47 & 0.52 & +228.4 & $-$1061.0 & 13.3 & 11.4 \\
\midrule
\multirow{8}{*}{Short Put} & DeepSeek-V4-Flash & 16 & 75.0 & +0.2 & 0.65 & 1.95 & +573.7 & $-$246.9 & 1.2 & 1.5 \\*
 & GPT-OSS-120B & 92 & 64.1 & +2.0 & 1.03 & 1.84 & +2742.1 & $-$1264.2 & 15.6 & 17.9 \\*
 & Qwen3-235B-A22B & 110 & 53.6 & +0.3 & 0.92 & 1.07 & +135.6 & $-$254.7 & 39.5 & 40.0 \\*
 & Llama-3.3-70B-Instruct & 67 & 53.7 & $-$0.6 & 0.74 & 0.86 & +150.8 & $-$541.8 & 23.0 & 22.5 \\*
 & MiniMax-M3 & 70 & 67.1 & +0.6 & 0.74 & 1.51 & +585.6 & $-$288.5 & 4.7 & 5.4 \\*
 & GLM-5.3-Flash & 21 & 66.7 & +0.1 & 0.64 & 1.28 & +78.8 & $-$254.9 & 1.6 & 1.7 \\*
 & GPT-5.6-Terra & 11 & 81.8 & +0.4 & 6.46 & 29.08 & +45.0 & $-$69.2 & 0.8 & 1.2 \\*
 & Gemini-3-Flash & 110 & 69.1 & +0.6 & 0.55 & 1.23 & +1053.8 & $-$698.3 & 12.3 & 13.1 \\
\end{longtable}
\normalsize

Figure~\ref{fig:intraday_activity_profile} disentangles trading activity
(top panel, fills per 15-minute bin) from inventory (bottom panel, mean
open contracts), and the two are not redundant. Fills per session range
from $14.7$ for MiniMax-M3 to $33.6$ for GPT-OSS-120B, and every model
front-loads the session except one: $29.1\%$ to $43.3\%$ of fills land
before $11{:}00$~ET, and the peak bin is $09{:}45$ or $10{:}00$ for five of
eight models and $10{:}15$ for two more. Gemini-3-Flash is the exception, with
its busiest bin at $15{:}00$ and the lowest morning share in the panel. For the other models, inventory peaks later; peak inventory spans $1.36$ to $2.96$ mean open
contracts, so the morning burst is round-tripping rather than
accumulation. Late-session activity is thin, $3.2\%$ to $8.7\%$ of fills
after $15{:}00$~ET, and most models are flat well before the close, though
the exceptions carry contracts into the forced close in the final bin.
All times are Eastern, converted from the recorded UTC instants with a
daylight-saving aware zone; a fixed offset would misplace the $66$ EDT
sessions of the $102$ by one hour and push part of the morning burst into
pre-open bins.

\begin{table}[H]
\centering
\footnotesize
\caption{Per-model Greek and PnL attribution for intraday 0-DTE SPY, over the 102 selected sessions per model. Columns decompose the mean per-session PnL into first-order Greek contributions ($\delta$-PnL $\approx \bar{\Delta}\,\Delta S$, $\gamma$-PnL $\approx \tfrac{1}{2}\bar{\Gamma}(\Delta S)^2$, $\theta$-PnL $\approx \bar{\Theta}\,\Delta t$, $\nu$-PnL $\approx \bar{\nu}\,\Delta\sigma$), the second-order cross terms Vanna and Volga, and an unexplained residual. The residual absorbs the discontinuous payoff at expiry, which a first-order expansion cannot represent. Attribution is at model level: the recorded runs do not carry a per-strategy-bucket Greek split.}
\label{tab:intraday_greek_decomp_spy}
\renewcommand{\arraystretch}{1.05}
\setlength{\tabcolsep}{4pt}
\begin{adjustbox}{width=\linewidth}
\begin{tabular}{l r r r r r r r r r}
\toprule
\multirow{2}{*}{\textbf{Model}} & \multirow{2}{*}{\textbf{Cases}} & \multirow{2}{*}{\textbf{Net PnL (\$)}} & \multicolumn{7}{c}{\textit{BSM attribution, mean \$ per session}} \\
\cmidrule(l){4-10}
 & & & \textbf{$\delta$} & \textbf{$\gamma$} & \textbf{$\theta$} & \textbf{$\nu$} & \textbf{Vanna} & \textbf{Volga} & \textbf{Resid} \\
\midrule
DeepSeek-V4-Flash & 102 & -7,456 & $-$19 & +123 & $-$163 & +17 & $-$2 & +2 & $-$32 \\
GPT-OSS-120B & 102 & 1,439 & +73 & +211 & $-$271 & +32 & $-$3 & +4 & $-$32 \\
Qwen3-235B-A22B & 102 & -3,705 & $-$9 & +86 & $-$133 & +21 & $-$5 & +21 & $-$18 \\
Llama-3.3-70B-Instruct & 102 & -6,059 & $-$36 & +120 & $-$186 & +14 & $-$2 & +47 & $-$16 \\
MiniMax-M3 & 102 & 1,435 & +25 & +76 & $-$98 & +16 & $-$2 & +5 & $-$7 \\
GLM-5.3-Flash & 102 & 1,096 & +38 & +108 & $-$146 & +25 & $-$2 & +3 & $-$16 \\
GPT-5.6-Terra & 102 & -1,839 & +1 & +109 & $-$131 & +12 & $-$0 & +5 & $-$13 \\
Gemini-3-Flash & 102 & -1,130 & +11 & +93 & $-$121 & +18 & +3 & +1 & $-$15 \\
\bottomrule
\end{tabular}
\end{adjustbox}
\end{table}

Table~\ref{tab:intraday_greek_decomp_spy} decomposes the mean per-session
PnL into the linearized contributions of $\delta$, $\gamma$, $\theta$ and
$\nu$, the second-order cross terms vanna and volga, and a residual
absorbing what the expansion cannot explain. The picture is mechanically
similar across the panel rather than model-distinguishing.

Theta is negative for every model, $-98$ to $-271$ dollars per session,
and vega is small and positive, $+12$ to $+32$ dollars, consistent with
short-dated long-premium exposure carried for minutes to hours rather than
across a volatility event. Gamma is positive for every model, $+76$ to
$+211$ dollars, as a long-premium book near expiry requires. Delta
contributions straddle zero, $-36$ to $+73$ dollars per session, so
direction selection at fill time is not what separates the models.

The residual is negative for every model, $-7$ to $-32$ dollars per
session. Its origin is largely mechanical: contracts held to expiry settle
at zero or at intrinsic value while every Greek also goes to zero, so the
discrete jump from entry premium to settlement cannot be reproduced by a
first-order expansion and lands in the residual, signed by the side of the
trade. The decomposition therefore describes where risk was carried and
what it cost, and does not establish that any of it was mispriced.

A first-order expansion is a linearization, so the natural question is how
much of the realized mark change it actually accounts for. We answer it by
repricing exactly rather than expanding. For every consecutive step pair of
every held position we revalue the option sequentially in time, then spot,
then volatility, and treat the remainder as residual; because this reprices
rather than expands, it remains valid across jumps.
Table~\ref{tab:repricing_intraday} reports the result. The exact repricing
accounts for $80$ to $93\%$ of the absolute mark change at step level and $76$
to $88\%$ after signed aggregation within a session, leaving a residual share
of $0.06$ to $0.15$. Spot is the dominant term at $0.71$ to $0.77$ of
attributed magnitude, volatility contributes $0.12$ to $0.14$ and time $0.03$
to $0.04$, matching the sign structure the linearized table shows. The 0DTE
attribution in Table~\ref{tab:intraday_greek_decomp_spy} therefore rests on a
decomposition that an exact method reproduces, and the conclusions drawn from
it in this track can be read as quantitative rather than merely directional.

\begin{table}[t]
\centering
\small
\setlength{\tabcolsep}{5pt}
\caption{Full-repricing P\&L-explain for the 0DTE track. Each consecutive step
pair of each held position is revalued exactly, sequentially in time, spot and
volatility, and the remainder is the residual; because this reprices rather
than expands, it stays valid across jumps. \emph{Moved} is the share of step
pairs whose recorded mark changed, and the explained fractions are computed
over those pairs, since a pair with no mark change contributes nothing to
episode PnL. \emph{Explained} is $1-\sum|\text{residual}|/\sum|\Delta\text{mark}|$
at step level, and \emph{Session} the same after signed aggregation within a
session. The last four columns are each term's share of total absolute
attributed magnitude.}
\label{tab:repricing_intraday}
\begin{tabular}{lrrrrrrrr}
\toprule
 & & & \multicolumn{2}{c}{Explained} & \multicolumn{4}{c}{Share of Attribution} \\
\cmidrule(lr){4-5} \cmidrule(l){6-9}
Model & Pairs & Moved & Step & Session & Time & Spot & Vol & Resid \\
\midrule
DeepSeek-V4-Flash & 40,159 & 92\% & 0.92 & 0.87 & 0.04 & 0.76 & 0.13 & 0.06 \\
GPT-OSS-120B & 53,180 & 91\% & 0.92 & 0.88 & 0.04 & 0.76 & 0.14 & 0.07 \\
Qwen3-235B-A22B & 56,443 & 84\% & 0.80 & 0.76 & 0.03 & 0.71 & 0.12 & 0.15 \\
Llama-3.3-70B-Instruct & 63,000 & 88\% & 0.85 & 0.86 & 0.03 & 0.72 & 0.14 & 0.12 \\
MiniMax-M3 & 39,182 & 90\% & 0.89 & 0.84 & 0.04 & 0.74 & 0.13 & 0.09 \\
GLM-5.3-Flash & 32,328 & 93\% & 0.91 & 0.88 & 0.04 & 0.76 & 0.12 & 0.08 \\
GPT-5.6-Terra & 31,442 & 93\% & 0.93 & 0.88 & 0.04 & 0.77 & 0.13 & 0.06 \\
Gemini-3-Flash & 54,577 & 87\% & 0.88 & 0.83 & 0.04 & 0.74 & 0.13 & 0.09 \\
\bottomrule
\end{tabular}
\end{table}

The same check cannot be run on the Earnings track, for a reason worth
stating precisely because it applies to the first-order table there as well.
In the six-provider-model mark-refresh audit, single-name option marks refresh in only $8$ to $9\%$ of minute steps, against
$84$ to $93\%$ for SPY 0DTE, so in $83\%$ of earnings step pairs the recorded
mark is unchanged while the underlying has moved. Any step-level decomposition
must then book the entire modelled change as residual, and exact repricing
inherits this defect rather than curing it: restricted to the pairs whose mark
did move, it explains only about half the absolute change, and signed
aggregation to the day or the episode does not improve that, which indicates a
measurement limit rather than mean-reverting quote noise. We therefore report
no step-level attribution for Earnings beyond the descriptive first-order
table, and read that table as carried exposure only.

\clearpage
\subsubsection{Earnings Bet}

\begin{figure}[h]
    \centering
    \includegraphics[width=0.99\linewidth]{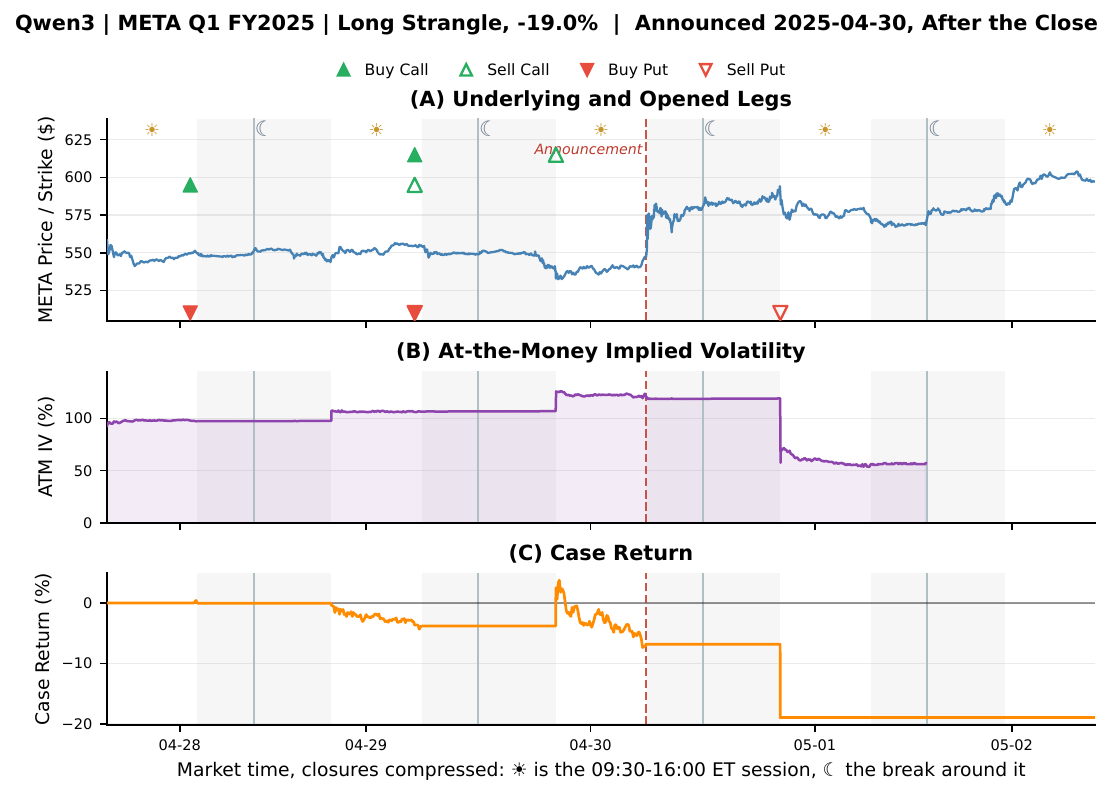}
    \caption{A worked Earnings Bet case: Qwen3-235B-A22B on META Q1 FY2025, which reported after the close on 30 April. \textbf{(A)} The underlying, with a marker at each opened leg. \textbf{(B)} At-the-money implied volatility, the volume-weighted mean over bars within $5\%$ of spot on the expiry the run actually traded. \textbf{(C)} Case return. The dashed line marks the announcement, so everything to its right is the reaction. Markers sit at each leg's strike, and hollow markers are sells. The axis is market time with the closures compressed to a narrow band, so no line is drawn across hours in which no bar exists. A sun marks each $09{:}30$--$16{:}00$~ET exchange session and a moon the shaded break around it, which holds the pre-market, after-hours and overnight steps. The volatility series stops at the end of the session before expiry, because implied volatility is not stably solvable at zero days to expiry: the same contracts print values above $200\%$ purely from the collapse in time to expiry. The return curve is piecewise because equity is marked at decision and fill points rather than every minute.}
    \label{fig:earnings_sample_case}
\end{figure}

Figure~\ref{fig:earnings_sample_case} works through one case end to end, to
make the mechanism behind the aggregate Greek attribution concrete.
Qwen3-235B-A22B opens a long strangle on META two sessions before the
announcement, buying the \$595 call and the \$510 put, and adds to both legs
the next day. Implied volatility behaves as the earnings literature expects:
it climbs from $95\%$ to a $125\%$ peak on the announcement session, then
halves to about $57\%$ once the report is out. The underlying does move, from
\$540 to \$580 overnight, but the strangle still loses $18.97\%$, because the
volatility the agent paid for collapses faster than the move pays off. This is
the per-case shape of the theta and residual terms that dominate
Table~\ref{tab:earnings_fills_greeks} across the panel.

\begin{figure}[t]
    \centering
    \includegraphics[width=0.99\linewidth]{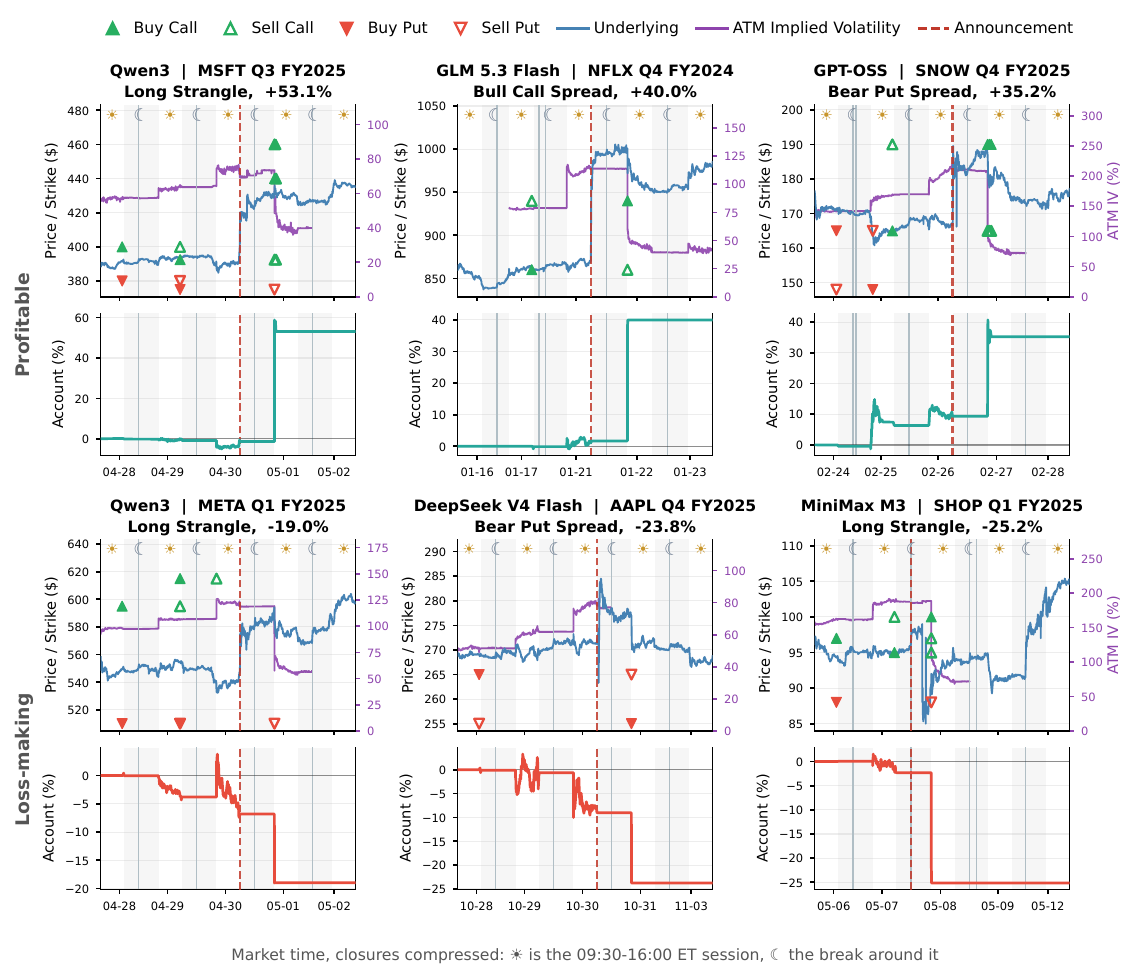}
    \caption{Six further Earnings Bet cases, three profitable (top) and three loss-making (bottom). For each case the upper panel carries the underlying, a marker at every fill placed at that leg's strike, and at-the-money implied volatility in purple; the lower panel carries the account. Hollow markers are sells. The dashed line marks the announcement. Axis conventions follow Figure~\ref{fig:earnings_sample_case}. All six opened net-debit structures, and the volatility path has the same shape throughout, so the rows are separated by the position still held when the report landed rather than by the structure chosen.}
    \label{fig:earnings_sample_grid}
\end{figure}

Figure~\ref{fig:earnings_sample_grid} sets six further cases side by side, three profitable and three not. All six opened net-debit structures and the implied-volatility path has the same shape in every panel, so the split is not between buying and selling volatility. What separates the rows is the position still on the book when the report landed. The three gains each carried upside into an up move: MSFT held both strangle legs into a $8.7\%$ rise, NFLX and SNOW held bull call spreads into $7.1\%$ and $6.4\%$. The three losses failed differently. META had sold both call legs the previous session and carried only the \$510 put into a $9.1\%$ rise, so the largest move of the six produced the worst leg selection. AAPL held a debit put spread through a net $-0.9\%$, leaving both legs out of the money. SHOP was already down $21.4\%$ of its eventual $25.2\%$ before the report, having extended a strangle over the two preceding sessions. The structure label is therefore a weak summary: it covers both rows, and what decided the outcome is what the agent still held at the announcement.

Table~\ref{tab:earnings_early_exit} compares three exit policies across all eight models. In $673$ of $755$ cases, $89.1\%$, the book is already flat at the $09{:}35$~ET policy time. The comparison is therefore conditional on the $82$ cases with an open position. Books are valued as cash plus signed option inventory at as-of marks; a SHORT quantity is already negative and is not negated a second time. This corrects the earlier exit-repricing helper for all models without changing any recorded end-of-case return.

On those $82$ cases, the $09{:}35$ valuation exceeds the recorded end-of-case return by a mean of $1.60$ and a median of $0.81$ percentage points, and does so in $65.9\%$ of cases. The conditional mean difference is positive for all eight models, from $0.30$ for Gemini-3-Flash to $2.92$ for MiniMax-M3, but each model contributes only $5$ to $17$ non-flat cases. At $12{:}30$~ET, model-level mean returns are within $0.01$ points of the end-of-case figure. The policy comparison holds decisions fixed and excludes additional liquidation costs; it does not establish an optimal exit rule. Terminal repricing reproduces the recorded final equity in every case, and no policy valuation needs a missing-mark fallback.

\vspace{-12pt}
\begin{table}[!htbp]
\centering
\caption{Earnings-case performance under three exit policies, computed by repricing every open leg at the option close for the policy's timestamp. \emph{T+1 09:35 ET} closes the book five minutes after the first regular-session open following the release, \emph{T+1 12:30 ET} closes mid-session that same day, and \emph{End of case} is the simulator's default. \textbf{Flat} counts the cases in which the book is flat at the 09:35 policy time, so the three policies coincide and the comparison is vacuous; the policy columns are therefore driven by the remaining minority of cases.}
\label{tab:earnings_early_exit}
\renewcommand{\arraystretch}{1.05}
\setlength{\tabcolsep}{4pt}
\footnotesize
\begin{adjustbox}{width=\linewidth}
\begin{tabular}{l r r r r r r r r r r r r r r}
\toprule
\multirow{2}{*}{\textbf{Model}} & \multirow{2}{*}{\textbf{Cases}} & \multirow{2}{*}{\textbf{Flat}} & \multicolumn{4}{c}{\textbf{T+1 09:35 ET}} & \multicolumn{4}{c}{\textbf{T+1 12:30 ET}} & \multicolumn{4}{c}{\textbf{End of Case}} \\
\cmidrule(lr){4-7} \cmidrule(lr){8-11} \cmidrule(lr){12-15}
 & & & \textbf{Mean\%} & \textbf{Med\%} & \textbf{Win\%} & \textbf{PF} & \textbf{Mean\%} & \textbf{Med\%} & \textbf{Win\%} & \textbf{PF} & \textbf{Mean\%} & \textbf{Med\%} & \textbf{Win\%} & \textbf{PF} \\
\midrule
DeepSeek-V4-Flash & 94 & 89 & $-$0.49 & $-$1.28 & 45.7 & 0.88 & $-$0.59 & $-$1.28 & 45.7 & 0.86 & $-$0.59 & $-$1.28 & 45.7 & 0.86 \\
GPT-OSS-120B & 95 & 83 & $-$0.10 & $-$0.41 & 45.3 & 0.97 & $-$0.20 & $-$0.53 & 44.2 & 0.94 & $-$0.20 & $-$0.53 & 44.2 & 0.94 \\
Qwen3-235B-A22B & 94 & 77 & $-$0.76 & $-$1.20 & 30.9 & 0.82 & $-$1.22 & $-$1.20 & 30.9 & 0.73 & $-$1.23 & $-$1.20 & 30.9 & 0.72 \\
Llama-3.3-70B-Instruct & 90 & 73 & $-$0.28 & $-$0.76 & 35.6 & 0.91 & $-$0.37 & $-$0.78 & 34.4 & 0.89 & $-$0.37 & $-$0.78 & 34.4 & 0.89 \\
MiniMax-M3 & 95 & 87 & $-$1.56 & $-$1.61 & 37.9 & 0.60 & $-$1.81 & $-$1.61 & 38.9 & 0.54 & $-$1.81 & $-$1.61 & 38.9 & 0.54 \\
GLM-5.3-Flash & 95 & 84 & $-$0.25 & $-$0.90 & 47.4 & 0.93 & $-$0.48 & $-$0.92 & 46.3 & 0.86 & $-$0.48 & $-$0.92 & 46.3 & 0.86 \\
GPT-5.6-Terra & 96 & 91 & 0.54 & $-$0.60 & 42.7 & 1.22 & 0.40 & $-$0.68 & 41.7 & 1.16 & 0.40 & $-$0.68 & 41.7 & 1.16 \\
Gemini-3-Flash & 96 & 89 & $-$0.66 & $-$0.17 & 37.5 & 0.83 & $-$0.68 & $-$0.17 & 37.5 & 0.83 & $-$0.68 & $-$0.17 & 37.5 & 0.83 \\
\bottomrule
\addlinespace
\multicolumn{15}{p{\dimexpr\linewidth-2\tabcolsep\relax}}{\scriptsize All eight models are repriced using signed contract quantities. The terminal repricing is checked against recorded final equity to within one cent per case. The policies value the positions held at the stated time and do not simulate behavioural feedback or additional liquidation costs.}

\end{tabular}
\end{adjustbox}
\end{table}

\begin{table}[!htbp]
\centering
\caption{Trading profile of each model across all earnings cases. \textbf{Size}: option fills per case and average premium notional deployed per case. \textbf{Tenor}: contract-quantity-weighted mean days to expiry on opening fills and FIFO holding period in days. \textbf{Direction}: share of recorded option-fill quantity in call legs. \textbf{Structure mix}: share of classified opening-intent groups, summing to approximately 100\% per model after rounding. Closing intents are excluded; these groups do not necessarily correspond to executed multi-leg fills.}
\label{tab:model_profile}
\renewcommand{\arraystretch}{1.1}
\setlength{\tabcolsep}{4pt}
\footnotesize
\begin{adjustbox}{width=\linewidth}
\begin{tabular}{l r rr rr r r r r r r r}
\toprule
\multirow{2}{*}{\textbf{Model}} & \multirow{2}{*}{\textbf{Cases}} & \multicolumn{2}{c}{\textbf{Size}} & \multicolumn{2}{c}{\textbf{Tenor}} & \multirow{2}{*}{\textbf{Call\%}} & \multicolumn{6}{c}{\textbf{Structure Mix (\%)}} \\
\cmidrule(lr){3-4} \cmidrule(lr){5-6} \cmidrule(l){8-13}
 & & \textbf{Fills} & \textbf{Prem.\ (k\$)} & \textbf{DTE} & \textbf{Hold} & & \textbf{Vert./2-leg} & \textbf{Single} & \textbf{Straddle} & \textbf{Strangle} & \textbf{Condor} & \textbf{Other} \\
\midrule
DeepSeek-V4-Flash & 94 & 6.6 & 35.2 & 4.0 & 2.2 & 40.1 & 91.8 & 5.1 & 0.3 & 2.0 & 0.0 & 0.8 \\
GPT-OSS-120B & 95 & 7.2 & 31.7 & 4.1 & 2.2 & 44.7 & 85.5 & 5.9 & 1.5 & 2.2 & 0.0 & 4.9 \\
Qwen3-235B-A22B & 94 & 8.5 & 21.7 & 3.6 & 1.5 & 42.5 & 60.8 & 10.0 & 2.8 & 18.5 & 0.0 & 7.9 \\
Llama-3.3-70B-Instruct & 90 & 7.5 & 20.9 & 4.0 & 2.0 & 59.0 & 68.4 & 14.7 & 4.9 & 3.2 & 0.0 & 8.8 \\
MiniMax-M3 & 95 & 6.0 & 29.2 & 4.1 & 2.3 & 25.5 & 93.0 & 2.9 & 1.0 & 2.1 & 0.0 & 1.0 \\
GLM-5.3-Flash & 95 & 7.5 & 29.6 & 4.0 & 2.1 & 32.7 & 92.3 & 0.7 & 2.5 & 0.9 & 0.0 & 3.6 \\
GPT-5.6-Terra & 96 & 5.6 & 22.4 & 4.0 & 2.2 & 18.3 & 91.8 & 3.4 & 1.4 & 2.1 & 0.0 & 1.4 \\
Gemini-3-Flash & 96 & 4.9 & 25.2 & 4.0 & 2.1 & 76.0 & 74.8 & 3.9 & 3.9 & 17.3 & 0.0 & 0.0 \\
\bottomrule
\addlinespace
\multicolumn{13}{p{\dimexpr\linewidth-2\tabcolsep\relax}}{\scriptsize Structure shares apply the same opening-intent classification to all eight models. Multiple groups can be opened in one case, so these percentages have a different denominator from the per-case structure table.}

\end{tabular}
\end{adjustbox}
\end{table}

The eight agents express an earnings view through similar structures but at
very different size, as summarized in Table~\ref{tab:model_profile}.
Vertical and two-leg positions dominate for every model, $60.8\%$ to
$93.0\%$ of classified opening-intent groups, and outright straddles and strangles
are a minority everywhere, so the sharp stylistic split of the earlier
five-model panel does not reproduce. Directional tilt does vary: call legs
span $18.3\%$ of recorded quantity for GPT-5.6-Terra to $76.0\%$ for
Gemini-3-Flash. The structure shares use the same opening-intent classifier
for all eight models, exclude closing intents, and describe intent groups
rather than asserting that every group became an executed multi-leg position.

Position sizing is far more uniform than before. Average premium per case
spans $20.9\text{k}$ to $35.2\text{k}$ dollars, a $1.7$-fold rather than
eight-fold spread, and entry maturity and holding period are
similar across the panel at $3.6$ to $4.1$ days and $1.5$ to $2.3$ days.
Neither structure choice nor sizing orders the per-model outcomes of
Table~\ref{tab:per_ticker_metrics}, which is itself informative: on this
record the earnings task does not reward a particular structure or a
particular notional, and the cross-model dispersion in outcome is not
explained by either. That is a weaker and more specific claim than
attributing the dispersion to sizing, which the current panel does not
support.

\begin{small}
\setlength{\tabcolsep}{4pt}
\begin{longtable}{@{\extracolsep{\fill}} l l r r r r r r r @{}}
\caption{Per-ticker earnings metrics across all models, aggregated over the quarterly earnings cases per ticker. Mean and median are per-case returns, MDD is the mean maximum drawdown within a case, Fills counts executed option-fill records.}\label{tab:per_ticker_metrics} \\
\toprule
\textbf{Ticker} & \textbf{Model} & \textbf{Cases} & \textbf{Mean\%} & \textbf{Med\%} & \textbf{Win\%} & \textbf{PF} & \textbf{MDD\%} & \textbf{Fills} \\
\midrule
\endfirsthead
\multicolumn{9}{l}{\small\itshape (continued from previous page)} \\
\toprule
\textbf{Ticker} & \textbf{Model} & \textbf{Cases} & \textbf{Mean\%} & \textbf{Med\%} & \textbf{Win\%} & \textbf{PF} & \textbf{MDD\%} & \textbf{Fills} \\
\midrule
\endhead
\midrule
\multicolumn{9}{r}{\small\itshape continued on next page} \\
\endfoot
\bottomrule
\endlastfoot
\multirow{8}{*}{AAPL} & DeepSeek-V4-Flash & 4 & $-$6.03 & $-$7.24 & 50.0 & 0.38 & 14.10 & 28 \\*
 & GPT-OSS-120B & 4 & $-$3.08 & $-$2.60 & 25.0 & 0.12 & 4.39 & 32 \\*
 & Qwen3-235B-A22B & 4 & $-$3.19 & $-$3.07 & 0.0 & 0.00 & 3.67 & 28 \\*
 & Llama-3.3-70B-Instruct & 4 & $-$1.56 & $-$1.45 & 50.0 & 0.54 & 5.21 & 27 \\*
 & MiniMax-M3 & 4 & $-$4.57 & $-$7.49 & 25.0 & 0.33 & 9.11 & 32 \\*
 & GLM-5.3-Flash & 4 & $-$1.04 & $-$1.53 & 50.0 & 0.72 & 6.29 & 38 \\*
 & GPT-5.6-Terra & 4 & 1.05 & $-$2.31 & 50.0 & 1.41 & 5.60 & 30 \\*
 & Gemini-3-Flash & 4 & $-$5.52 & $-$4.92 & 0.0 & 0.00 & 6.71 & 20 \\
\midrule
\multirow{8}{*}{AMD} & DeepSeek-V4-Flash & 4 & 8.19 & 7.53 & 100.0 & -- & 5.81 & 34 \\*
 & GPT-OSS-120B & 4 & $-$1.10 & 0.17 & 50.0 & 0.59 & 5.22 & 32 \\*
 & Qwen3-235B-A22B & 4 & 2.12 & 2.14 & 75.0 & 8.78 & 1.79 & 24 \\*
 & Llama-3.3-70B-Instruct & 4 & $-$6.81 & $-$4.82 & 0.0 & 0.00 & 8.00 & 30 \\*
 & MiniMax-M3 & 4 & $-$1.38 & $-$1.28 & 25.0 & 0.53 & 6.23 & 24 \\*
 & GLM-5.3-Flash & 4 & 2.76 & 2.50 & 75.0 & 34.69 & 3.35 & 36 \\*
 & GPT-5.6-Terra & 4 & 3.58 & 3.20 & 100.0 & -- & 3.21 & 24 \\*
 & Gemini-3-Flash & 4 & $-$10.74 & $-$4.51 & 0.0 & 0.00 & 13.29 & 24 \\
\midrule
\multirow{8}{*}{AMZN} & DeepSeek-V4-Flash & 3 & $-$1.89 & $-$1.75 & 33.3 & 0.60 & 6.47 & 24 \\*
 & GPT-OSS-120B & 4 & $-$0.00 & $-$0.55 & 50.0 & 1.00 & 4.40 & 32 \\*
 & Qwen3-235B-A22B & 3 & $-$15.07 & $-$13.19 & 0.0 & 0.00 & 17.46 & 24 \\*
 & Llama-3.3-70B-Instruct & 3 & $-$3.82 & $-$4.44 & 33.3 & 0.50 & 9.53 & 41 \\*
 & MiniMax-M3 & 4 & 3.31 & 2.48 & 75.0 & 5.35 & 2.63 & 20 \\*
 & GLM-5.3-Flash & 4 & 0.87 & 1.08 & 75.0 & 3.06 & 1.75 & 26 \\*
 & GPT-5.6-Terra & 4 & 0.25 & 0.16 & 50.0 & 1.62 & 2.08 & 28 \\*
 & Gemini-3-Flash & 4 & 6.87 & 3.22 & 75.0 & 2.01 & 11.52 & 26 \\
\midrule
\multirow{8}{*}{AVGO} & DeepSeek-V4-Flash & 4 & $-$6.74 & $-$5.69 & 0.0 & 0.00 & 9.00 & 32 \\*
 & GPT-OSS-120B & 4 & $-$0.03 & $-$4.81 & 25.0 & 0.99 & 6.78 & 24 \\*
 & Qwen3-235B-A22B & 4 & 7.13 & 2.40 & 50.0 & 2.67 & 10.38 & 36 \\*
 & Llama-3.3-70B-Instruct & 4 & $-$4.62 & $-$5.30 & 25.0 & 0.32 & 8.96 & 26 \\*
 & MiniMax-M3 & 4 & $-$11.11 & $-$8.98 & 0.0 & 0.00 & 12.90 & 19 \\*
 & GLM-5.3-Flash & 4 & $-$12.27 & $-$11.11 & 0.0 & 0.00 & 14.46 & 28 \\*
 & GPT-5.6-Terra & 4 & $-$5.83 & $-$4.21 & 0.0 & 0.00 & 7.28 & 26 \\*
 & Gemini-3-Flash & 4 & $-$3.35 & $-$4.04 & 50.0 & 0.48 & 10.17 & 24 \\
\midrule
\multirow{8}{*}{CELH} & DeepSeek-V4-Flash & 4 & $-$3.89 & $-$8.56 & 25.0 & 0.51 & 9.30 & 24 \\*
 & GPT-OSS-120B & 4 & 5.08 & 5.14 & 50.0 & 5.26 & 2.27 & 22 \\*
 & Qwen3-235B-A22B & 4 & 4.90 & 4.01 & 50.0 & 4.05 & 3.23 & 31 \\*
 & Llama-3.3-70B-Instruct & 3 & $-$0.41 & $-$0.43 & 0.0 & 0.00 & 0.79 & 8 \\*
 & MiniMax-M3 & 4 & 3.55 & 2.09 & 50.0 & 4.75 & 1.53 & 20 \\*
 & GLM-5.3-Flash & 4 & $-$2.09 & $-$2.75 & 25.0 & 0.29 & 3.86 & 22 \\*
 & GPT-5.6-Terra & 4 & 1.15 & 0.29 & 50.0 & 3.60 & 0.76 & 12 \\*
 & Gemini-3-Flash & 4 & 4.10 & $-$1.60 & 25.0 & 2.50 & 4.63 & 16 \\
\midrule
\multirow{8}{*}{COIN} & DeepSeek-V4-Flash & 4 & 0.47 & 0.16 & 50.0 & 1.16 & 6.14 & 26 \\*
 & GPT-OSS-120B & 4 & 9.81 & 10.42 & 75.0 & 6.82 & 7.37 & 27 \\*
 & Qwen3-235B-A22B & 4 & $-$10.93 & $-$11.89 & 0.0 & 0.00 & 15.83 & 28 \\*
 & Llama-3.3-70B-Instruct & 4 & 0.55 & $-$0.18 & 50.0 & 1.32 & 5.23 & 30 \\*
 & MiniMax-M3 & 4 & 2.75 & 0.36 & 50.0 & 2.00 & 7.81 & 23 \\*
 & GLM-5.3-Flash & 4 & $-$7.09 & $-$8.22 & 25.0 & 0.16 & 12.05 & 20 \\*
 & GPT-5.6-Terra & 4 & 5.85 & 1.62 & 50.0 & 4.58 & 7.27 & 20 \\*
 & Gemini-3-Flash & 4 & $-$11.42 & $-$12.34 & 0.0 & 0.00 & 18.41 & 16 \\
\midrule
\multirow{8}{*}{CRM} & DeepSeek-V4-Flash & 4 & 5.96 & 5.81 & 50.0 & 4.08 & 6.19 & 24 \\*
 & GPT-OSS-120B & 4 & $-$1.70 & $-$1.69 & 25.0 & 0.67 & 9.93 & 18 \\*
 & Qwen3-235B-A22B & 4 & $-$10.29 & $-$10.40 & 25.0 & 0.01 & 12.73 & 33 \\*
 & Llama-3.3-70B-Instruct & 4 & $-$4.22 & $-$5.65 & 25.0 & 0.26 & 8.98 & 32 \\*
 & MiniMax-M3 & 4 & $-$3.18 & $-$2.49 & 50.0 & 0.40 & 11.21 & 16 \\*
 & GLM-5.3-Flash & 4 & 2.33 & 3.06 & 75.0 & 2.60 & 3.81 & 24 \\*
 & GPT-5.6-Terra & 4 & 0.86 & 0.86 & 50.0 & 1.59 & 3.09 & 12 \\*
 & Gemini-3-Flash & 4 & $-$14.13 & $-$13.31 & 0.0 & 0.00 & 16.38 & 18 \\
\midrule
\multirow{8}{*}{DIS} & DeepSeek-V4-Flash & 4 & $-$4.82 & $-$5.96 & 25.0 & 0.16 & 8.10 & 24 \\*
 & GPT-OSS-120B & 4 & $-$2.11 & $-$1.19 & 25.0 & 0.42 & 5.89 & 37 \\*
 & Qwen3-235B-A22B & 4 & $-$1.22 & $-$2.20 & 50.0 & 0.61 & 4.48 & 41 \\*
 & Llama-3.3-70B-Instruct & 3 & $-$0.44 & $-$0.40 & 33.3 & 0.18 & 0.88 & 14 \\*
 & MiniMax-M3 & 4 & $-$5.16 & $-$4.96 & 0.0 & 0.00 & 6.76 & 32 \\*
 & GLM-5.3-Flash & 4 & $-$5.22 & $-$6.19 & 25.0 & 0.30 & 9.19 & 43 \\*
 & GPT-5.6-Terra & 4 & $-$0.61 & $-$1.40 & 25.0 & 0.66 & 2.36 & 20 \\*
 & Gemini-3-Flash & 4 & $-$4.25 & $-$3.81 & 25.0 & 0.11 & 5.74 & 20 \\
\midrule
\multirow{8}{*}{HOOD} & DeepSeek-V4-Flash & 4 & $-$0.96 & $-$2.74 & 50.0 & 0.85 & 10.24 & 36 \\*
 & GPT-OSS-120B & 4 & $-$2.09 & $-$1.97 & 25.0 & 0.20 & 4.74 & 34 \\*
 & Qwen3-235B-A22B & 4 & 0.49 & $-$1.15 & 25.0 & 1.21 & 4.94 & 37 \\*
 & Llama-3.3-70B-Instruct & 4 & $-$1.48 & $-$3.23 & 25.0 & 0.49 & 3.99 & 31 \\*
 & MiniMax-M3 & 4 & 1.31 & $-$1.46 & 25.0 & 2.01 & 3.94 & 28 \\*
 & GLM-5.3-Flash & 4 & $-$1.95 & $-$2.59 & 25.0 & 0.13 & 6.11 & 37 \\*
 & GPT-5.6-Terra & 4 & $-$1.74 & $-$1.37 & 0.0 & 0.00 & 3.90 & 24 \\*
 & Gemini-3-Flash & 4 & $-$0.40 & $-$0.87 & 50.0 & 0.68 & 3.06 & 20 \\
\midrule
\multirow{8}{*}{INTC} & DeepSeek-V4-Flash & 4 & $-$2.97 & $-$1.58 & 25.0 & 0.07 & 4.15 & 28 \\*
 & GPT-OSS-120B & 4 & 0.74 & 1.05 & 75.0 & 6.61 & 1.28 & 32 \\*
 & Qwen3-235B-A22B & 4 & $-$1.70 & $-$1.48 & 25.0 & 0.01 & 2.08 & 28 \\*
 & Llama-3.3-70B-Instruct & 4 & $-$0.09 & $-$0.05 & 50.0 & 0.67 & 0.58 & 37 \\*
 & MiniMax-M3 & 4 & $-$1.06 & $-$1.35 & 25.0 & 0.16 & 1.68 & 28 \\*
 & GLM-5.3-Flash & 4 & $-$1.62 & $-$1.79 & 25.0 & 0.08 & 2.65 & 54 \\*
 & GPT-5.6-Terra & 4 & $-$0.90 & $-$0.86 & 25.0 & 0.12 & 1.24 & 32 \\*
 & Gemini-3-Flash & 4 & 0.88 & 0.98 & 100.0 & -- & 0.68 & 20 \\
\midrule
\multirow{8}{*}{META} & DeepSeek-V4-Flash & 4 & 0.21 & 5.40 & 75.0 & 1.05 & 6.87 & 24 \\*
 & GPT-OSS-120B & 4 & 0.91 & 3.46 & 75.0 & 1.43 & 4.11 & 16 \\*
 & Qwen3-235B-A22B & 4 & 8.35 & 0.00 & 25.0 & 2.76 & 7.02 & 17 \\*
 & Llama-3.3-70B-Instruct & 4 & 1.57 & 3.71 & 50.0 & 1.27 & 9.19 & 16 \\*
 & MiniMax-M3 & 4 & $-$6.22 & $-$5.11 & 25.0 & 0.26 & 10.05 & 20 \\*
 & GLM-5.3-Flash & 4 & 2.08 & 6.33 & 75.0 & 1.43 & 8.73 & 27 \\*
 & GPT-5.6-Terra & 4 & 1.48 & 0.69 & 50.0 & 1.27 & 9.47 & 20 \\*
 & Gemini-3-Flash & 4 & 5.25 & 10.31 & 75.0 & 2.17 & 8.78 & 20 \\
\midrule
\multirow{8}{*}{MSFT} & DeepSeek-V4-Flash & 4 & 6.93 & $-$0.57 & 50.0 & 3.63 & 6.05 & 28 \\*
 & GPT-OSS-120B & 4 & 3.83 & 3.06 & 75.0 & 3.38 & 7.11 & 29 \\*
 & Qwen3-235B-A22B & 4 & 12.89 & 4.93 & 50.0 & 5.53 & 5.51 & 25 \\*
 & Llama-3.3-70B-Instruct & 4 & 16.46 & 22.31 & 75.0 & 14.35 & 7.87 & 35 \\*
 & MiniMax-M3 & 4 & $-$2.95 & $-$2.32 & 25.0 & 0.26 & 7.47 & 28 \\*
 & GLM-5.3-Flash & 4 & 3.63 & 1.58 & 50.0 & 8.68 & 7.11 & 28 \\*
 & GPT-5.6-Terra & 4 & $-$4.24 & $-$4.10 & 25.0 & 0.20 & 8.70 & 29 \\*
 & Gemini-3-Flash & 4 & 11.94 & 11.29 & 75.0 & -- & 2.30 & 24 \\
\midrule
\multirow{8}{*}{MU} & DeepSeek-V4-Flash & 4 & $-$9.48 & $-$9.71 & 25.0 & 0.11 & 14.03 & 30 \\*
 & GPT-OSS-120B & 4 & $-$0.70 & $-$1.68 & 50.0 & 0.85 & 7.53 & 28 \\*
 & Qwen3-235B-A22B & 4 & $-$3.13 & $-$1.19 & 50.0 & 0.36 & 8.40 & 26 \\*
 & Llama-3.3-70B-Instruct & 4 & $-$0.92 & $-$2.44 & 25.0 & 0.66 & 5.80 & 32 \\*
 & MiniMax-M3 & 4 & $-$1.97 & $-$0.30 & 50.0 & 0.49 & 5.91 & 24 \\*
 & GLM-5.3-Flash & 4 & 1.96 & 1.24 & 50.0 & 2.10 & 4.16 & 24 \\*
 & GPT-5.6-Terra & 4 & 1.77 & 0.02 & 50.0 & 1.76 & 4.57 & 23 \\*
 & Gemini-3-Flash & 4 & 0.82 & $-$2.32 & 25.0 & 1.31 & 7.02 & 26 \\
\midrule
\multirow{8}{*}{NFLX} & DeepSeek-V4-Flash & 3 & 0.61 & $-$4.37 & 33.3 & 1.10 & 9.14 & 16 \\*
 & GPT-OSS-120B & 3 & $-$0.49 & $-$6.34 & 33.3 & 0.91 & 6.90 & 16 \\*
 & Qwen3-235B-A22B & 3 & 0.00 & 0.00 & 0.0 & -- & 0.00 & 0 \\*
 & Llama-3.3-70B-Instruct & 3 & 11.30 & 0.00 & 33.3 & 9.68 & 3.16 & 20 \\*
 & MiniMax-M3 & 3 & $-$3.47 & $-$6.34 & 33.3 & 0.49 & 8.41 & 16 \\*
 & GLM-5.3-Flash & 3 & 5.22 & $-$7.76 & 33.3 & 1.64 & 11.05 & 16 \\*
 & GPT-5.6-Terra & 4 & 6.12 & 4.46 & 50.0 & 6.49 & 2.32 & 18 \\*
 & Gemini-3-Flash & 4 & 0.00 & 0.00 & 0.0 & -- & 0.00 & 0 \\
\midrule
\multirow{8}{*}{NKE} & DeepSeek-V4-Flash & 4 & $-$0.29 & 2.01 & 75.0 & 0.91 & 6.02 & 26 \\*
 & GPT-OSS-120B & 4 & $-$2.93 & $-$0.68 & 25.0 & 0.06 & 4.26 & 34 \\*
 & Qwen3-235B-A22B & 4 & 6.47 & 7.74 & 75.0 & 17.98 & 1.78 & 47 \\*
 & Llama-3.3-70B-Instruct & 3 & 0.61 & 0.95 & 66.7 & 1.80 & 1.35 & 29 \\*
 & MiniMax-M3 & 4 & 0.24 & 2.80 & 75.0 & 1.11 & 4.09 & 24 \\*
 & GLM-5.3-Flash & 4 & 4.54 & 4.97 & 100.0 & -- & 1.78 & 33 \\*
 & GPT-5.6-Terra & 4 & 0.90 & 1.47 & 50.0 & 2.22 & 1.46 & 20 \\*
 & Gemini-3-Flash & 4 & $-$1.77 & $-$1.33 & 25.0 & 0.27 & 3.40 & 20 \\
\midrule
\multirow{8}{*}{NVDA} & DeepSeek-V4-Flash & 4 & 0.27 & $-$1.54 & 50.0 & 1.12 & 5.08 & 34 \\*
 & GPT-OSS-120B & 4 & $-$5.67 & $-$7.14 & 25.0 & 0.05 & 7.88 & 34 \\*
 & Qwen3-235B-A22B & 4 & $-$1.20 & $-$1.26 & 0.0 & 0.00 & 2.72 & 27 \\*
 & Llama-3.3-70B-Instruct & 4 & $-$5.05 & $-$4.16 & 0.0 & 0.00 & 7.27 & 28 \\*
 & MiniMax-M3 & 4 & $-$7.82 & $-$7.79 & 0.0 & 0.00 & 9.41 & 28 \\*
 & GLM-5.3-Flash & 4 & $-$3.06 & $-$3.62 & 25.0 & 0.14 & 4.87 & 32 \\*
 & GPT-5.6-Terra & 4 & $-$5.33 & $-$5.56 & 0.0 & 0.00 & 7.24 & 31 \\*
 & Gemini-3-Flash & 4 & 7.32 & 7.72 & 50.0 & 4.64 & 5.37 & 20 \\
\midrule
\multirow{8}{*}{PLTR} & DeepSeek-V4-Flash & 4 & 0.66 & $-$2.03 & 50.0 & 1.13 & 7.89 & 16 \\*
 & GPT-OSS-120B & 4 & 0.36 & 2.41 & 75.0 & 1.06 & 7.29 & 40 \\*
 & Qwen3-235B-A22B & 4 & $-$0.44 & $-$2.14 & 50.0 & 0.94 & 9.96 & 56 \\*
 & Llama-3.3-70B-Instruct & 4 & $-$2.45 & $-$4.96 & 25.0 & 0.57 & 7.41 & 22 \\*
 & MiniMax-M3 & 4 & 1.15 & 0.64 & 50.0 & 1.53 & 3.60 & 16 \\*
 & GLM-5.3-Flash & 4 & 10.55 & 8.75 & 75.0 & 5.26 & 5.36 & 20 \\*
 & GPT-5.6-Terra & 4 & 1.17 & $-$0.45 & 50.0 & 1.29 & 5.76 & 16 \\*
 & Gemini-3-Flash & 4 & $-$1.36 & $-$6.73 & 25.0 & 0.80 & 8.94 & 16 \\
\midrule
\multirow{8}{*}{PYPL} & DeepSeek-V4-Flash & 4 & $-$0.71 & $-$0.19 & 50.0 & 0.55 & 3.01 & 22 \\*
 & GPT-OSS-120B & 4 & $-$5.74 & $-$5.96 & 0.0 & 0.00 & 6.36 & 35 \\*
 & Qwen3-235B-A22B & 4 & $-$2.35 & $-$3.08 & 25.0 & 0.21 & 4.54 & 58 \\*
 & Llama-3.3-70B-Instruct & 4 & $-$1.18 & $-$1.55 & 50.0 & 0.63 & 4.49 & 40 \\*
 & MiniMax-M3 & 4 & $-$2.03 & $-$0.74 & 50.0 & 0.45 & 5.22 & 28 \\*
 & GLM-5.3-Flash & 4 & $-$2.45 & $-$3.65 & 25.0 & 0.39 & 6.37 & 37 \\*
 & GPT-5.6-Terra & 4 & $-$1.41 & $-$0.64 & 50.0 & 0.54 & 4.37 & 26 \\*
 & Gemini-3-Flash & 4 & $-$1.25 & $-$0.44 & 0.0 & 0.00 & 2.22 & 22 \\
\midrule
\multirow{8}{*}{RKLB} & DeepSeek-V4-Flash & 4 & 4.39 & 2.47 & 75.0 & 11.16 & 4.94 & 22 \\*
 & GPT-OSS-120B & 4 & $-$0.27 & $-$1.44 & 50.0 & 0.85 & 3.66 & 16 \\*
 & Qwen3-235B-A22B & 4 & $-$1.91 & $-$1.81 & 0.0 & 0.00 & 2.47 & 17 \\*
 & Llama-3.3-70B-Instruct & 3 & $-$2.19 & $-$2.49 & 0.0 & 0.00 & 2.98 & 10 \\*
 & MiniMax-M3 & 4 & 0.00 & $-$0.06 & 50.0 & 1.00 & 6.03 & 25 \\*
 & GLM-5.3-Flash & 4 & 0.90 & $-$0.70 & 50.0 & 1.73 & 2.55 & 24 \\*
 & GPT-5.6-Terra & 4 & $-$0.86 & $-$0.37 & 25.0 & 0.11 & 1.51 & 12 \\*
 & Gemini-3-Flash & 4 & $-$2.87 & $-$1.97 & 25.0 & 0.03 & 4.21 & 24 \\
\midrule
\multirow{8}{*}{SHOP} & DeepSeek-V4-Flash & 4 & $-$4.76 & $-$4.34 & 25.0 & 0.20 & 11.48 & 22 \\*
 & GPT-OSS-120B & 4 & $-$5.55 & $-$4.59 & 25.0 & 0.19 & 10.39 & 29 \\*
 & Qwen3-235B-A22B & 4 & $-$8.34 & $-$10.30 & 25.0 & 0.23 & 11.56 & 30 \\*
 & Llama-3.3-70B-Instruct & 4 & 2.94 & $-$1.22 & 25.0 & 3.08 & 2.32 & 25 \\*
 & MiniMax-M3 & 4 & $-$10.91 & $-$9.42 & 25.0 & 0.01 & 13.17 & 20 \\*
 & GLM-5.3-Flash & 4 & $-$5.51 & $-$7.92 & 25.0 & 0.35 & 9.52 & 20 \\*
 & GPT-5.6-Terra & 4 & $-$4.36 & $-$9.09 & 25.0 & 0.46 & 8.38 & 16 \\*
 & Gemini-3-Flash & 4 & $-$2.63 & 0.03 & 50.0 & 0.12 & 5.92 & 18 \\
\midrule
\multirow{8}{*}{SMCI} & DeepSeek-V4-Flash & 4 & $-$0.02 & 0.07 & 50.0 & 0.97 & 3.77 & 28 \\*
 & GPT-OSS-120B & 4 & 3.16 & 2.12 & 100.0 & -- & 2.32 & 32 \\*
 & Qwen3-235B-A22B & 4 & 0.41 & 0.59 & 50.0 & 1.34 & 4.63 & 92 \\*
 & Llama-3.3-70B-Instruct & 4 & 1.55 & 0.34 & 50.0 & 3.59 & 2.18 & 37 \\*
 & MiniMax-M3 & 4 & 1.78 & 2.10 & 50.0 & 2.86 & 4.98 & 26 \\*
 & GLM-5.3-Flash & 4 & 3.76 & 1.82 & 75.0 & 28.66 & 2.19 & 44 \\*
 & GPT-5.6-Terra & 4 & 0.75 & 1.04 & 75.0 & 2.11 & 2.13 & 31 \\*
 & Gemini-3-Flash & 4 & 1.67 & 1.31 & 100.0 & -- & 1.81 & 20 \\
\midrule
\multirow{8}{*}{SNOW} & DeepSeek-V4-Flash & 4 & 3.81 & 0.95 & 50.0 & 1.90 & 8.11 & 28 \\*
 & GPT-OSS-120B & 4 & 6.86 & $-$1.31 & 50.0 & 3.84 & 5.43 & 28 \\*
 & Qwen3-235B-A22B & 4 & $-$6.64 & $-$6.77 & 50.0 & 0.13 & 11.28 & 34 \\*
 & Llama-3.3-70B-Instruct & 4 & 0.37 & 1.49 & 75.0 & 1.09 & 7.40 & 40 \\*
 & MiniMax-M3 & 4 & 5.92 & 5.75 & 75.0 & 5.59 & 5.68 & 24 \\*
 & GLM-5.3-Flash & 4 & $-$5.10 & $-$3.77 & 50.0 & 0.22 & 11.08 & 22 \\*
 & GPT-5.6-Terra & 4 & 5.24 & 11.89 & 75.0 & 2.27 & 8.90 & 20 \\*
 & Gemini-3-Flash & 4 & 1.95 & 6.45 & 75.0 & 1.47 & 6.63 & 20 \\
\midrule
\multirow{8}{*}{SOFI} & DeepSeek-V4-Flash & 4 & $-$1.57 & $-$2.44 & 25.0 & 0.37 & 3.46 & 24 \\*
 & GPT-OSS-120B & 4 & $-$1.79 & $-$0.97 & 25.0 & 0.07 & 2.62 & 31 \\*
 & Qwen3-235B-A22B & 4 & $-$1.47 & $-$1.68 & 0.0 & 0.00 & 1.87 & 43 \\*
 & Llama-3.3-70B-Instruct & 4 & $-$0.52 & $-$0.68 & 25.0 & 0.32 & 1.64 & 52 \\*
 & MiniMax-M3 & 4 & $-$2.64 & $-$3.81 & 25.0 & 0.12 & 3.51 & 19 \\*
 & GLM-5.3-Flash & 4 & $-$1.08 & $-$1.48 & 25.0 & 0.27 & 1.71 & 35 \\*
 & GPT-5.6-Terra & 4 & $-$0.83 & $-$1.24 & 25.0 & 0.18 & 1.23 & 20 \\*
 & Gemini-3-Flash & 4 & $-$0.23 & $-$1.19 & 25.0 & 0.77 & 1.46 & 26 \\
\midrule
\multirow{8}{*}{TSLA} & DeepSeek-V4-Flash & 4 & $-$1.45 & $-$1.97 & 50.0 & 0.61 & 9.02 & 16 \\*
 & GPT-OSS-120B & 4 & $-$2.43 & $-$5.57 & 25.0 & 0.48 & 6.14 & 24 \\*
 & Qwen3-235B-A22B & 4 & $-$7.51 & $-$6.56 & 25.0 & 0.08 & 8.99 & 21 \\*
 & Llama-3.3-70B-Instruct & 4 & $-$6.55 & $-$10.09 & 25.0 & 0.21 & 11.44 & 16 \\*
 & MiniMax-M3 & 4 & 0.62 & 1.92 & 75.0 & 1.40 & 5.52 & 32 \\*
 & GLM-5.3-Flash & 4 & $-$0.29 & 0.08 & 50.0 & 0.85 & 7.04 & 27 \\*
 & GPT-5.6-Terra & 4 & 5.58 & 1.36 & 50.0 & 7.35 & 3.09 & 23 \\*
 & Gemini-3-Flash & 4 & 2.72 & 0.00 & 25.0 & -- & 0.35 & 8 \\
\end{longtable}
\end{small}

Table~\ref{tab:earnings_strategy} decomposes performance by the structure
opened, and what it shows first is a directional choice about volatility
rather than a ranking. Net-debit structures account for $645$ of the $755$
cases, $85.4\%$: the agents buy premium into the announcement almost to the
exclusion of selling it, with credit spreads and naked short legs together
only $76$ cases, $10.1\%$. That is the side of the trade that the volatility
crush works against, and it is the same side every case in
Figure~\ref{fig:earnings_sample_grid} is on.

The most used structure is the bear put spread at $320$ cases, and its
mean return is negative for six of eight models, per-model means running
from $-3.37\%$ to $+0.44\%$ for an equal-model average of $-0.85\%$.
The bull call spread, $192$ cases, is positive for five of eight models,
with an equal-model average of $+0.84\%$ over a per-model range of
$-2.02\%$ to $+5.44\%$. Outright straddles and strangles, $118$ cases
between them, average $-4.19\%$ and $-4.52\%$ across models. The remaining
structures have small per-model samples and do not support a ranking.
Structure choice therefore does not by itself separate
the models, but the common choice across all of them, paying for premium
before a scheduled volatility contraction, is consistent with the theta and
residual terms that dominate Table~\ref{tab:earnings_fills_greeks}.

On Greek attribution, Table~\ref{tab:earnings_fills_greeks} shows theta
large and negative for every model, $-625$ to $-3{,}108$ dollars per
case over the holding window, and vega positive but small next to it,
$+11$ to $+658$ dollars. Delta spans $-65$ to $+1{,}166$ dollars,
with Gemini-3-Flash the only negative cell. The residual is highly variable, from
$-31{,}340$ to $+1{,}685$ dollars per case, which is a larger scale than
the realized PnL for most models. Earnings windows combine a large spot
jump with an IV collapse, and both violate the small-move assumption
behind a first-order expansion, so the attribution should be read as a
description of carried exposure and not as a profit explanation for
premium-heavy positions.

\setlength{\LTleft}{0pt}
\setlength{\LTright}{0pt}
\renewcommand{\arraystretch}{1.05}
\setlength{\tabcolsep}{4pt}
\footnotesize
\begin{longtable}{@{\extracolsep{\fill}} l l r r r r r r @{}}
\caption{Earnings performance by the structure opened in each case. Cases counts the cases in which the model opened that structure and Share its percentage of that model's classified cases. Buckets a model used only a handful of times are reported for completeness but do not support a ranking.}\label{tab:earnings_strategy} \\
\toprule
\textbf{Structure} & \textbf{Model} & \textbf{Cases} & \textbf{Share\%} & \textbf{Mean\%} & \textbf{Med\%} & \textbf{Win\%} & \textbf{PF} \\
\midrule
\endfirsthead
\multicolumn{8}{l}{\footnotesize\itshape (continued from previous page)} \\
\toprule
\textbf{Structure} & \textbf{Model} & \textbf{Cases} & \textbf{Share\%} & \textbf{Mean\%} & \textbf{Med\%} & \textbf{Win\%} & \textbf{PF} \\
\midrule
\endhead
\midrule
\multicolumn{8}{r}{\footnotesize\itshape continued on next page} \\
\endfoot
\bottomrule
\endlastfoot
\multirow{7}{*}{Bear Call Spread} & DeepSeek-V4-Flash & 3 & 3.2 & $-$4.05 & $-$4.37 & 0.0 & 0.00 \\*
 & GPT-OSS-120B & 5 & 5.3 & 3.49 & 1.68 & 60.0 & 2.55 \\*
 & Qwen3-235B-A22B & 1 & 1.1 & 52.37 & 52.37 & 100.0 & -- \\*
 & Llama-3.3-70B-Instruct & 3 & 3.3 & $-$10.19 & $-$9.91 & 0.0 & 0.00 \\*
 & MiniMax-M3 & 3 & 3.2 & $-$1.45 & $-$6.34 & 33.3 & 0.71 \\*
 & GLM-5.3-Flash & 4 & 4.2 & $-$2.13 & $-$4.66 & 25.0 & 0.56 \\*
 & GPT-5.6-Terra & 1 & 1.0 & 24.40 & 24.40 & 100.0 & -- \\
\midrule
\multirow{8}{*}{Bear Put Spread} & DeepSeek-V4-Flash & 48 & 51.1 & $-$0.28 & $-$0.19 & 50.0 & 0.92 \\*
 & GPT-OSS-120B & 41 & 43.2 & $-$0.31 & $-$1.35 & 41.5 & 0.89 \\*
 & Qwen3-235B-A22B & 23 & 24.5 & $-$3.37 & $-$1.23 & 34.8 & 0.33 \\*
 & Llama-3.3-70B-Instruct & 18 & 20.0 & $-$0.10 & $-$0.73 & 38.9 & 0.95 \\*
 & MiniMax-M3 & 58 & 61.1 & $-$1.91 & $-$1.68 & 36.2 & 0.50 \\*
 & GLM-5.3-Flash & 52 & 54.7 & 0.44 & 0.75 & 51.9 & 1.21 \\*
 & GPT-5.6-Terra & 65 & 67.7 & 0.04 & $-$1.11 & 41.5 & 1.02 \\*
 & Gemini-3-Flash & 15 & 15.6 & $-$1.29 & $-$0.53 & 33.3 & 0.49 \\
\midrule
\multirow{8}{*}{Bull Call Spread} & DeepSeek-V4-Flash & 22 & 23.4 & $-$0.99 & $-$2.28 & 45.5 & 0.81 \\*
 & GPT-OSS-120B & 23 & 24.2 & $-$2.02 & $-$3.98 & 34.8 & 0.54 \\*
 & Qwen3-235B-A22B & 11 & 11.7 & 5.44 & 0.37 & 54.5 & 2.61 \\*
 & Llama-3.3-70B-Instruct & 19 & 21.1 & 0.36 & $-$1.73 & 36.8 & 1.06 \\*
 & MiniMax-M3 & 21 & 22.1 & $-$0.55 & $-$1.13 & 42.9 & 0.81 \\*
 & GLM-5.3-Flash & 24 & 25.3 & 1.18 & $-$0.25 & 50.0 & 1.27 \\*
 & GPT-5.6-Terra & 23 & 24.0 & 1.79 & 0.43 & 52.2 & 1.54 \\*
 & Gemini-3-Flash & 49 & 51.0 & 1.55 & 0.07 & 51.0 & 1.36 \\
\midrule
\multirow{6}{*}{Bull Put Spread} & DeepSeek-V4-Flash & 5 & 5.3 & 2.69 & $-$1.73 & 40.0 & 1.73 \\*
 & GPT-OSS-120B & 4 & 4.2 & 7.74 & 1.07 & 75.0 & 5.84 \\*
 & Qwen3-235B-A22B & 3 & 3.2 & 3.89 & 3.53 & 66.7 & 7.79 \\*
 & Llama-3.3-70B-Instruct & 5 & 5.6 & 1.24 & $-$0.13 & 40.0 & 5.87 \\*
 & MiniMax-M3 & 3 & 3.2 & 2.66 & 1.94 & 66.7 & 1.71 \\*
 & GLM-5.3-Flash & 4 & 4.2 & $-$7.37 & $-$7.82 & 25.0 & 0.06 \\
\midrule
\multirow{3}{*}{Long Call} & DeepSeek-V4-Flash & 1 & 1.1 & 14.34 & 14.34 & 100.0 & -- \\*
 & Llama-3.3-70B-Instruct & 3 & 3.3 & 3.38 & $-$1.22 & 33.3 & 4.40 \\*
 & Gemini-3-Flash & 3 & 3.1 & $-$2.87 & $-$4.10 & 33.3 & 0.14 \\
\midrule
\multirow{4}{*}{Long Put} & DeepSeek-V4-Flash & 1 & 1.1 & 7.51 & 7.51 & 100.0 & -- \\*
 & GPT-OSS-120B & 2 & 2.1 & 2.03 & 2.03 & 100.0 & -- \\*
 & Llama-3.3-70B-Instruct & 4 & 4.4 & 5.31 & 3.88 & 50.0 & 2.39 \\*
 & Gemini-3-Flash & 1 & 1.0 & $-$19.73 & $-$19.73 & 0.0 & 0.00 \\
\midrule
\multirow{5}{*}{None} & GPT-OSS-120B & 1 & 1.1 & 0.00 & 0.00 & 0.0 & -- \\*
 & Qwen3-235B-A22B & 4 & 4.3 & 0.00 & 0.00 & 0.0 & -- \\*
 & Llama-3.3-70B-Instruct & 1 & 1.1 & 0.00 & 0.00 & 0.0 & -- \\*
 & GPT-5.6-Terra & 2 & 2.1 & 0.00 & 0.00 & 0.0 & -- \\*
 & Gemini-3-Flash & 10 & 10.4 & 0.00 & 0.00 & 0.0 & -- \\
\midrule
\multirow{6}{*}{Other Multi-Leg} & DeepSeek-V4-Flash & 2 & 2.1 & 2.97 & 2.97 & 100.0 & -- \\*
 & GPT-OSS-120B & 7 & 7.4 & $-$2.28 & $-$3.38 & 28.6 & 0.40 \\*
 & Qwen3-235B-A22B & 1 & 1.1 & 2.54 & 2.54 & 100.0 & -- \\*
 & Llama-3.3-70B-Instruct & 4 & 4.4 & $-$2.30 & $-$3.79 & 25.0 & 0.41 \\*
 & GLM-5.3-Flash & 1 & 1.1 & $-$17.48 & $-$17.48 & 0.0 & 0.00 \\*
 & GPT-5.6-Terra & 1 & 1.0 & 0.00 & 0.00 & 0.0 & -- \\
\midrule
\multirow{4}{*}{Short Call} & DeepSeek-V4-Flash & 2 & 2.1 & $-$4.33 & $-$4.33 & 0.0 & 0.00 \\*
 & GPT-OSS-120B & 3 & 3.2 & $-$6.38 & $-$2.90 & 33.3 & 0.23 \\*
 & Llama-3.3-70B-Instruct & 2 & 2.2 & 7.39 & 7.39 & 50.0 & 50.92 \\*
 & MiniMax-M3 & 2 & 2.1 & $-$3.14 & $-$3.14 & 50.0 & 0.21 \\
\midrule
\multirow{6}{*}{Short Put} & DeepSeek-V4-Flash & 4 & 4.3 & $-$3.03 & $-$11.79 & 25.0 & 0.74 \\*
 & GPT-OSS-120B & 3 & 3.2 & $-$0.49 & $-$1.53 & 33.3 & 0.87 \\*
 & Qwen3-235B-A22B & 4 & 4.3 & $-$2.47 & 0.59 & 50.0 & 0.32 \\*
 & Llama-3.3-70B-Instruct & 9 & 10.0 & $-$0.34 & $-$0.31 & 44.4 & 0.80 \\*
 & MiniMax-M3 & 2 & 2.1 & $-$0.45 & $-$0.45 & 50.0 & 0.73 \\*
 & GLM-5.3-Flash & 1 & 1.1 & $-$1.48 & $-$1.48 & 0.0 & 0.00 \\
\midrule
\multirow{8}{*}{Straddle} & DeepSeek-V4-Flash & 1 & 1.1 & $-$15.65 & $-$15.65 & 0.0 & 0.00 \\*
 & GPT-OSS-120B & 4 & 4.2 & 5.31 & 10.41 & 75.0 & 2.17 \\*
 & Qwen3-235B-A22B & 9 & 9.6 & $-$5.00 & $-$4.76 & 22.2 & 0.10 \\*
 & Llama-3.3-70B-Instruct & 14 & 15.6 & $-$3.19 & $-$4.08 & 21.4 & 0.34 \\*
 & MiniMax-M3 & 3 & 3.2 & $-$1.31 & 2.14 & 66.7 & 0.64 \\*
 & GLM-5.3-Flash & 7 & 7.4 & $-$4.21 & $-$0.92 & 42.9 & 0.36 \\*
 & GPT-5.6-Terra & 2 & 2.1 & $-$8.02 & $-$8.02 & 0.0 & 0.00 \\*
 & Gemini-3-Flash & 2 & 2.1 & $-$1.44 & $-$1.44 & 50.0 & 0.25 \\
\midrule
\multirow{8}{*}{Strangle} & DeepSeek-V4-Flash & 5 & 5.3 & $-$2.48 & $-$0.46 & 40.0 & 0.24 \\*
 & GPT-OSS-120B & 2 & 2.1 & 1.28 & 1.28 & 100.0 & -- \\*
 & Qwen3-235B-A22B & 38 & 40.4 & $-$2.88 & $-$1.78 & 18.4 & 0.45 \\*
 & Llama-3.3-70B-Instruct & 8 & 8.9 & $-$0.34 & 0.00 & 37.5 & 0.76 \\*
 & MiniMax-M3 & 3 & 3.2 & $-$13.98 & $-$15.27 & 0.0 & 0.00 \\*
 & GLM-5.3-Flash & 2 & 2.1 & $-$5.19 & $-$5.19 & 0.0 & 0.00 \\*
 & GPT-5.6-Terra & 2 & 2.1 & $-$6.85 & $-$6.85 & 0.0 & 0.00 \\*
 & Gemini-3-Flash & 16 & 16.7 & $-$5.69 & $-$3.93 & 25.0 & 0.07 \\
\end{longtable}
\normalsize

\begin{table}[H]
\centering
\footnotesize
\caption{Per-model Greek and PnL attribution for the Earnings Bet task, over the valid cases per model. Columns decompose the mean per-case PnL into first-order Greek contributions, the second-order cross terms Vanna and Volga, and an unexplained residual, over the holding window of each case. Theta is large and negative for every model, consistent with long-premium exposure held across the announcement, while the residual is highly variable across models. Attribution is at model level: the recorded runs do not carry a per-structure Greek split.}
\label{tab:earnings_fills_greeks}
\renewcommand{\arraystretch}{1.05}
\setlength{\tabcolsep}{4pt}
\begin{adjustbox}{width=\linewidth}
\begin{tabular}{l r r r r r r r r r}
\toprule
\multirow{2}{*}{\textbf{Model}} & \multirow{2}{*}{\textbf{Cases}} & \multirow{2}{*}{\textbf{Net PnL (\$)}} & \multicolumn{7}{c}{\textit{BSM attribution, mean \$ per case}} \\
\cmidrule(l){4-10}
 & & & \textbf{$\delta$} & \textbf{$\gamma$} & \textbf{$\theta$} & \textbf{$\nu$} & \textbf{Vanna} & \textbf{Volga} & \textbf{Resid} \\
\midrule
DeepSeek-V4-Flash & 94 & -27,549 & +553 & +9659 & $-$1642 & +11 & +20 & $-$159 & $-$8737 \\
GPT-OSS-120B & 95 & -9,681 & +659 & +20157 & $-$1503 & +260 & $-$8 & $-$25 & $-$19642 \\
Qwen3-235B-A22B & 94 & -57,682 & +239 & +32689 & $-$3108 & +658 & +0 & +248 & $-$31340 \\
Llama-3.3-70B-Instruct & 90 & -16,428 & +750 & +271 & $-$1681 & +202 & +13 & $-$17 & +279 \\
MiniMax-M3 & 95 & -85,944 & +257 & $-$864 & $-$1408 & +176 & +28 & $-$136 & +1043 \\
GLM-5.3-Flash & 95 & -22,923 & +1166 & +13988 & $-$1178 & +319 & +9 & $-$83 & $-$14463 \\
GPT-5.6-Terra & 96 & 19,285 & +609 & $-$1465 & $-$625 & +97 & +21 & $-$122 & +1685 \\
Gemini-3-Flash & 96 & -32,869 & $-$65 & +6852 & $-$1973 & +369 & $-$3 & $-$96 & $-$5427 \\
\bottomrule
\end{tabular}
\end{adjustbox}
\end{table}

The residual is negative for five of eight models; Llama-3.3-70B-Instruct,
MiniMax-M3 and GPT-5.6-Terra have positive residuals. It is not the largest
absolute component in every row. These differences caution against treating
the linearized terms as a universal profit explanation or inferring
attribution quality from a structure label. The six-provider mark-refresh
audit above also identifies a measurement limitation: a step-level method
cannot recover price changes that were never recorded.

\subsection{Ablation and Pairwise Detail}
\label{appendix:ablation_detail}

Table~\ref{tab:ablations_by_task} expands the context ablations of
Section~\ref{sec:ablations} to every tested profile--task--model cell.
Table~\ref{tab:main_pairwise} reports all 28 model pairs per episodic task.
Only MiniMax-M3 versus GPT-5.6-Terra in Earnings is significant after
Holm correction ($p=0.046$); no 0DTE pair is significant.

\begin{table}[h]
\centering
\small
\setlength{\tabcolsep}{5pt}
\caption{Execution-layer ablation of the inventory check on the closing path.
Every recorded episode is replayed with the decision stream held fixed and
inversions clamped on any contract that received a duplicate closing intent,
which is the one defect this guard addresses. \emph{Eps} counts episodes in
which the guard binds at all; \emph{Mean} and \emph{Worst} are the mean and
minimum episode return, with and without it, in percent.}
\label{tab:verifier_ablation}
\begin{tabular}{lrr rr rr}
\toprule
 & & & \multicolumn{2}{c}{Mean Return} & \multicolumn{2}{c}{Worst Episode} \\
\cmidrule(lr){4-5} \cmidrule(l){6-7}
Model & $n$ & Eps & Off & Guard & Off & Guard \\
\midrule
\multicolumn{7}{@{}l}{\textit{Earnings}} \\[1pt]
DeepSeek-V4-Flash & 94 & 1 & $-0.6$ & $-0.5$ & $-23.8$ & $-23.8$ \\
GPT-OSS-120B & 95 & 0 & $-0.2$ & $-0.2$ & $-22.0$ & $-22.0$ \\
Qwen3-235B-A22B & 94 & 5 & $-1.2$ & $-1.2$ & $-22.8$ & $-22.8$ \\
Llama-3.3-70B-Instruct & 90 & 1 & $-0.4$ & $-0.4$ & $-23.2$ & $-23.2$ \\
MiniMax-M3 & 95 & 2 & $-1.8$ & $-1.5$ & $-25.2$ & $-25.2$ \\
GLM-5.3-Flash & 95 & 0 & $-0.5$ & $-0.5$ & $-20.8$ & $-20.8$ \\
GPT-5.6-Terra & 96 & 0 & $+0.4$ & $+0.4$ & $-16.6$ & $-16.6$ \\
Gemini-3-Flash & 96 & 0 & $-0.7$ & $-0.7$ & $-30.5$ & $-30.5$ \\
\midrule
\multicolumn{7}{@{}l}{\textit{0DTE Intraday}} \\[1pt]
DeepSeek-V4-Flash & 102 & 2 & $-0.7$ & $-0.7$ & $-6.7$ & $-6.7$ \\
GPT-OSS-120B & 102 & 17 & $+0.1$ & $-0.0$ & $-10.1$ & $-10.1$ \\
Qwen3-235B-A22B & 102 & 43 & $-0.4$ & $-0.2$ & $-7.4$ & $-4.6$ \\
Llama-3.3-70B-Instruct & 102 & 9 & $-0.6$ & $-0.6$ & $-6.3$ & $-6.3$ \\
MiniMax-M3 & 102 & 8 & $+0.1$ & $+0.1$ & $-3.7$ & $-3.7$ \\
GLM-5.3-Flash & 102 & 4 & $+0.1$ & $+0.1$ & $-4.3$ & $-4.3$ \\
GPT-5.6-Terra & 102 & 0 & $-0.2$ & $-0.2$ & $-6.1$ & $-6.1$ \\
Gemini-3-Flash & 102 & 10 & $-0.1$ & $-0.1$ & $-6.6$ & $-6.6$ \\
\bottomrule
\end{tabular}
\end{table}

\paragraph{Ablating an execution-layer component.}
We replay recorded episodes with decisions fixed, blocking inventory
inversions caused by duplicate closing intents. Blocked quantity changes
are reversed at recorded marks and their fees refunded.
Table~\ref{tab:verifier_ablation} shows mean-return changes of at most
$0.30$ percentage points. The worst episode is unchanged in 15 of 16
model--task cells; Qwen3-235B-A22B in 0DTE improves from $-7.4\%$ to
$-4.6\%$. No negative model mean becomes positive. This replay measures
the guard's mechanical effect, without modeling agent responses to rejected
orders.

\begingroup
\setlength{\LTleft}{0pt}
\setlength{\LTright}{0pt}
\renewcommand{\arraystretch}{1.12}
\setlength{\tabcolsep}{1.5pt}
\footnotesize
\begin{longtable}{@{\extracolsep{\fill}} l l r r c r r l l r @{}}
\caption{Context ablations by profile, task and model. $\Delta$Ret, its 95\% bootstrap CI, and Med are paired return changes in percentage points. NT and MDD show baseline $\to$ ablation (\%). Worse is the share of cases with lower returns. $p_{\mathrm{Holm}}$ is adjusted within each profile--task family. Overlay cells have only two paired cases.}\label{tab:ablations_by_task} \\
\toprule
\textbf{Task} & \textbf{Model} & \textbf{$n$} & \textbf{$\Delta$Ret} & \textbf{95\% CI} & \textbf{Med} & \textbf{Worse\%} & \textbf{NT\%} & \textbf{MDD\%} & \textbf{$p_{\mathrm{Holm}}$} \\
\midrule
\endfirsthead
\multicolumn{10}{l}{\scriptsize\itshape (continued from previous page)} \\
\toprule
\textbf{Task} & \textbf{Model} & \textbf{$n$} & \textbf{$\Delta$Ret} & \textbf{95\% CI} & \textbf{Med} & \textbf{Worse\%} & \textbf{NT\%} & \textbf{MDD\%} & \textbf{$p_{\mathrm{Holm}}$} \\
\midrule
\endhead
\midrule
\multicolumn{10}{r}{\scriptsize\itshape continued on next page} \\
\endfoot
\bottomrule
\endlastfoot
\multicolumn{10}{@{}l}{\textbf{Fixed Chain}} \\*
\multirow{3}{*}{0DTE} & DeepSeek-V4-Flash & 24 & $-$0.09 & [$-$0.86, 0.59] & 0.01 & 50.0 & 0.0 $\to$ 0.0 & 4.4 $\to$ 4.3 & 0.899 \\*
 & GPT-OSS-120B & 24 & $-$2.16 & [$-$4.36, $-$0.43] & $-$0.73 & 62.5 & 0.0 $\to$ 0.0 & 6.3 $\to$ 5.7 & 0.085 \\*
 & Qwen3-235B-A22B & 24 & $-$0.44 & [$-$1.53, 0.64] & $-$0.22 & 54.2 & 0.0 $\to$ 0.0 & 2.6 $\to$ 2.7 & 0.899 \\
\multirow{3}{*}{Covered Call} & DeepSeek-V4-Flash & 2 & $-$3.68 & [$-$9.83, 2.46] & $-$3.68 & 50.0 & 0.0 $\to$ 0.0 & 22.1 $\to$ 24.6 & 1.000 \\*
 & GPT-OSS-120B & 2 & $-$4.98 & [$-$10.29, 0.33] & $-$4.98 & 50.0 & 0.0 $\to$ 0.0 & 25.0 $\to$ 23.5 & 1.000 \\*
 & Qwen3-235B-A22B & 2 & $-$2.15 & [$-$6.48, 2.18] & $-$2.15 & 50.0 & 0.0 $\to$ 0.0 & 24.1 $\to$ 23.7 & 1.000 \\
\multirow{3}{*}{Earnings} & DeepSeek-V4-Flash & 24 & 1.70 & [$-$2.24, 6.10] & 1.37 & 37.5 & 0.0 $\to$ 0.0 & 7.1 $\to$ 6.5 & 1.000 \\*
 & GPT-OSS-120B & 24 & 2.67 & [$-$1.88, 8.40] & $-$0.73 & 54.2 & 0.0 $\to$ 0.0 & 6.2 $\to$ 6.5 & 1.000 \\*
 & Qwen3-235B-A22B & 24 & $-$0.07 & [$-$6.44, 5.95] & 0.91 & 41.7 & 4.2 $\to$ 8.3 & 7.8 $\to$ 7.7 & 1.000 \\
\multirow{3}{*}{Hedging} & DeepSeek-V4-Flash & 2 & 2.81 & [$-$0.30, 5.92] & 2.81 & 50.0 & 0.0 $\to$ 0.0 & 25.6 $\to$ 24.1 & 1.000 \\*
 & GPT-OSS-120B & 2 & $-$4.93 & [$-$7.41, $-$2.44] & $-$4.93 & 100.0 & 0.0 $\to$ 0.0 & 22.1 $\to$ 24.0 & 1.000 \\*
 & Qwen3-235B-A22B & 2 & $-$9.77 & [$-$18.15, $-$1.39] & $-$9.77 & 100.0 & 0.0 $\to$ 0.0 & 32.7 $\to$ 25.3 & 1.000 \\
\midrule
\multicolumn{10}{@{}l}{\textbf{No Portfolio State}} \\*
\multirow{3}{*}{0DTE} & DeepSeek-V4-Flash & 24 & 2.64 & [$-$3.26, 9.39] & $-$2.64 & 62.5 & 0.0 $\to$ 0.0 & 4.4 $\to$ 11.3 & 0.904 \\*
 & GPT-OSS-120B & 24 & 4.20 & [$-$1.79, 10.91] & $-$1.32 & 54.2 & 0.0 $\to$ 0.0 & 6.3 $\to$ 13.1 & 0.698 \\*
 & Qwen3-235B-A22B & 24 & 0.25 & [$-$5.09, 6.88] & $-$6.00 & 70.8 & 0.0 $\to$ 0.0 & 2.6 $\to$ 12.7 & 0.943 \\
\multirow{3}{*}{Covered Call} & DeepSeek-V4-Flash & 2 & $-$4.70 & [$-$11.12, 1.72] & $-$4.70 & 50.0 & 0.0 $\to$ 0.0 & 22.1 $\to$ 22.6 & 1.000 \\*
 & GPT-OSS-120B & 2 & 1.50 & [$-$3.05, 6.05] & 1.50 & 50.0 & 0.0 $\to$ 0.0 & 25.0 $\to$ 23.8 & 1.000 \\*
 & Qwen3-235B-A22B & 2 & 0.59 & [$-$5.51, 6.69] & 0.59 & 50.0 & 0.0 $\to$ 0.0 & 24.1 $\to$ 21.6 & 1.000 \\
\multirow{3}{*}{Earnings} & DeepSeek-V4-Flash & 24 & 0.98 & [$-$2.10, 4.20] & 0.90 & 45.8 & 0.0 $\to$ 0.0 & 7.1 $\to$ 5.3 & 1.000 \\*
 & GPT-OSS-120B & 24 & $-$1.05 & [$-$4.58, 2.32] & 0.00 & 45.8 & 0.0 $\to$ 0.0 & 6.2 $\to$ 7.7 & 1.000 \\*
 & Qwen3-235B-A22B & 24 & $-$4.42 & [$-$10.29, 0.52] & $-$1.08 & 58.3 & 4.2 $\to$ 16.7 & 7.8 $\to$ 10.3 & 0.413 \\
\multirow{3}{*}{Hedging} & DeepSeek-V4-Flash & 2 & $-$2.04 & [$-$2.38, $-$1.69] & $-$2.04 & 100.0 & 0.0 $\to$ 0.0 & 25.6 $\to$ 26.8 & 1.000 \\*
 & GPT-OSS-120B & 2 & $-$0.83 & [$-$5.97, 4.31] & $-$0.83 & 50.0 & 0.0 $\to$ 0.0 & 22.1 $\to$ 22.2 & 1.000 \\*
 & Qwen3-235B-A22B & 2 & 2.02 & [$-$2.38, 6.42] & 2.02 & 50.0 & 0.0 $\to$ 0.0 & 32.7 $\to$ 26.6 & 1.000 \\
\midrule
\multicolumn{10}{@{}l}{\textbf{No Task Rules}} \\*
\multirow{3}{*}{0DTE} & DeepSeek-V4-Flash & 24 & $-$0.28 & [$-$1.48, 0.76] & $-$0.20 & 58.3 & 0.0 $\to$ 0.0 & 4.4 $\to$ 3.9 & 1.000 \\*
 & GPT-OSS-120B & 24 & 0.15 & [$-$1.58, 2.29] & $-$0.63 & 70.8 & 0.0 $\to$ 0.0 & 6.3 $\to$ 6.9 & 1.000 \\*
 & Qwen3-235B-A22B & 24 & 1.06 & [0.02, 2.14] & 0.71 & 37.5 & 0.0 $\to$ 0.0 & 2.6 $\to$ 2.6 & 0.211 \\
\multirow{3}{*}{Covered Call} & DeepSeek-V4-Flash & 2 & $-$3.67 & [$-$12.13, 4.79] & $-$3.67 & 50.0 & 0.0 $\to$ 0.0 & 22.1 $\to$ 24.3 & 1.000 \\*
 & GPT-OSS-120B & 2 & $-$0.53 & [$-$1.48, 0.42] & $-$0.53 & 50.0 & 0.0 $\to$ 0.0 & 25.0 $\to$ 25.6 & 1.000 \\*
 & Qwen3-235B-A22B & 2 & $-$1.66 & [$-$5.63, 2.31] & $-$1.66 & 50.0 & 0.0 $\to$ 0.0 & 24.1 $\to$ 25.5 & 1.000 \\
\multirow{3}{*}{Earnings} & DeepSeek-V4-Flash & 24 & 0.54 & [$-$2.46, 3.42] & 1.26 & 41.7 & 0.0 $\to$ 16.7 & 7.1 $\to$ 4.3 & 1.000 \\*
 & GPT-OSS-120B & 24 & $-$0.09 & [$-$3.50, 4.06] & $-$1.25 & 62.5 & 0.0 $\to$ 0.0 & 6.2 $\to$ 5.6 & 1.000 \\*
 & Qwen3-235B-A22B & 24 & $-$1.54 & [$-$6.30, 1.99] & 0.00 & 41.7 & 4.2 $\to$ 12.5 & 7.8 $\to$ 7.6 & 1.000 \\
\multirow{3}{*}{Hedging} & DeepSeek-V4-Flash & 2 & 0.85 & [0.36, 1.33] & 0.85 & 0.0 & 0.0 $\to$ 0.0 & 25.6 $\to$ 28.7 & 1.000 \\*
 & GPT-OSS-120B & 2 & $-$2.01 & [$-$2.74, $-$1.29] & $-$2.01 & 100.0 & 0.0 $\to$ 0.0 & 22.1 $\to$ 23.7 & 1.000 \\*
 & Qwen3-235B-A22B & 2 & $-$2.84 & [$-$9.72, 4.04] & $-$2.84 & 50.0 & 0.0 $\to$ 0.0 & 32.7 $\to$ 25.7 & 1.000 \\
\midrule
\multicolumn{10}{@{}l}{\textbf{No News}} \\*
\multirow{3}{*}{Earnings} & DeepSeek-V4-Flash & 24 & $-$2.46 & [$-$7.44, 2.87] & $-$2.43 & 62.5 & 0.0 $\to$ 0.0 & 7.1 $\to$ 10.6 & 1.000 \\*
 & GPT-OSS-120B & 24 & 0.82 & [$-$5.96, 7.41] & 0.07 & 45.8 & 0.0 $\to$ 4.2 & 6.2 $\to$ 7.6 & 1.000 \\*
 & Qwen3-235B-A22B & 24 & 0.72 & [$-$5.46, 5.40] & 1.46 & 29.2 & 4.2 $\to$ 8.3 & 7.8 $\to$ 5.8 & 1.000 \\
\midrule
\multicolumn{10}{@{}l}{\textbf{No IV}} \\*
\multirow{3}{*}{Earnings} & DeepSeek-V4-Flash & 24 & $-$4.17 & [$-$7.95, $-$0.54] & $-$3.86 & 66.7 & 0.0 $\to$ 0.0 & 7.1 $\to$ 10.4 & 0.083 \\*
 & GPT-OSS-120B & 24 & $-$5.79 & [$-$10.23, $-$1.39] & $-$3.62 & 70.8 & 0.0 $\to$ 0.0 & 6.2 $\to$ 8.6 & 0.059 \\*
 & Qwen3-235B-A22B & 24 & 1.50 & [$-$5.45, 7.51] & 2.62 & 33.3 & 4.2 $\to$ 16.7 & 7.8 $\to$ 5.8 & 0.680 \\
\midrule
\multicolumn{10}{@{}l}{\textbf{No Greeks / Multi-TF}} \\*
\multirow{3}{*}{0DTE} & DeepSeek-V4-Flash & 24 & 0.77 & [$-$0.64, 1.96] & 2.41 & 29.2 & 0.0 $\to$ 95.8 & 4.4 $\to$ 0.2 & 0.835 \\*
 & GPT-OSS-120B & 24 & $-$0.35 & [$-$3.36, 2.34] & 1.91 & 33.3 & 0.0 $\to$ 0.0 & 6.3 $\to$ 2.7 & 1.000 \\*
 & Qwen3-235B-A22B & 24 & 0.33 & [$-$0.88, 1.34] & 0.86 & 20.8 & 0.0 $\to$ 45.8 & 2.6 $\to$ 0.2 & 1.000 \\
\end{longtable}
\endgroup

\begingroup
\setlength{\LTleft}{0pt}
\setlength{\LTright}{0pt}
\renewcommand{\arraystretch}{1.12}
\setlength{\tabcolsep}{2.5pt}
\footnotesize
\begin{longtable}{@{\extracolsep{\fill}} l l r r c r r r @{}}
\caption{Pairwise comparisons on shared episodes. $\Delta$Ret and Med are mean and median return differences (A $-$ B; percentage points). CIs use paired bootstrap resampling. $p$ uses 50{,}000 two-sided sign flips; Holm correction covers all 28 pairs within each task.}\label{tab:main_pairwise} \\
\toprule
\textbf{Model A} & \textbf{Model B} & \textbf{$n$} & \textbf{$\Delta$Ret} & \textbf{95\% CI} & \textbf{Med} & \textbf{$p$} & \textbf{$p_{\mathrm{Holm}}$} \\
\midrule
\endfirsthead
\multicolumn{8}{l}{\footnotesize\itshape (continued from previous page)} \\
\toprule
\textbf{Model A} & \textbf{Model B} & \textbf{$n$} & \textbf{$\Delta$Ret} & \textbf{95\% CI} & \textbf{Med} & \textbf{$p$} & \textbf{$p_{\mathrm{Holm}}$} \\
\midrule
\endhead
\midrule
\multicolumn{8}{r}{\footnotesize\itshape continued on next page} \\
\endfoot
\bottomrule
\endlastfoot
\multicolumn{8}{@{}l}{\textbf{Earnings}} \\*
\multirow{7}{*}{DeepSeek-V4-Flash} & GPT-OSS-120B & 94 & $-$0.28 & [$-$2.27, 1.64] & 0.05 & 0.781 & 1.000 \\*
 & Qwen3-235B-A22B & 94 & 0.64 & [$-$2.33, 3.57] & 0.30 & 0.673 & 1.000 \\*
 & Llama-3.3-70B-Instruct & 90 & $-$0.05 & [$-$2.41, 2.35] & 0.51 & 0.967 & 1.000 \\*
 & MiniMax-M3 & 94 & 1.28 & [$-$0.71, 3.34] & 0.45 & 0.226 & 1.000 \\*
 & GLM-5.3-Flash & 94 & $-$0.08 & [$-$2.11, 1.99] & $-$0.13 & 0.941 & 1.000 \\*
 & GPT-5.6-Terra & 94 & $-$0.98 & [$-$3.05, 1.17] & $-$0.81 & 0.369 & 1.000 \\*
 & Gemini-3-Flash & 94 & 0.63 & [$-$1.90, 3.23] & 0.58 & 0.630 & 1.000 \\*
\addlinespace[2pt]
\multirow{6}{*}{GPT-OSS-120B} & Qwen3-235B-A22B & 94 & 0.92 & [$-$2.05, 3.76] & 0.46 & 0.538 & 1.000 \\*
 & Llama-3.3-70B-Instruct & 90 & 0.02 & [$-$1.85, 1.95] & 0.13 & 0.988 & 1.000 \\*
 & MiniMax-M3 & 95 & 1.61 & [0.05, 3.27] & 0.47 & 0.055 & 1.000 \\*
 & GLM-5.3-Flash & 95 & 0.28 & [$-$1.75, 2.54] & $-$0.24 & 0.807 & 1.000 \\*
 & GPT-5.6-Terra & 95 & $-$0.61 & [$-$2.32, 1.11] & $-$0.75 & 0.494 & 1.000 \\*
 & Gemini-3-Flash & 95 & 0.49 & [$-$1.90, 2.91] & 0.43 & 0.702 & 1.000 \\*
\addlinespace[2pt]
\multirow{5}{*}{Qwen3-235B-A22B} & Llama-3.3-70B-Instruct & 90 & $-$0.88 & [$-$3.37, 1.65] & $-$0.13 & 0.498 & 1.000 \\*
 & MiniMax-M3 & 94 & 0.64 & [$-$2.01, 3.43] & 0.68 & 0.656 & 1.000 \\*
 & GLM-5.3-Flash & 94 & $-$0.72 & [$-$3.32, 2.02] & $-$0.74 & 0.607 & 1.000 \\*
 & GPT-5.6-Terra & 94 & $-$1.63 & [$-$4.14, 1.04] & $-$0.83 & 0.230 & 1.000 \\*
 & Gemini-3-Flash & 94 & $-$0.02 & [$-$2.38, 2.44] & 0.00 & 0.991 & 1.000 \\*
\addlinespace[2pt]
\multirow{4}{*}{Llama-3.3-70B-Instruct} & MiniMax-M3 & 90 & 1.58 & [$-$0.53, 3.67] & 1.30 & 0.144 & 1.000 \\*
 & GLM-5.3-Flash & 90 & $-$0.01 & [$-$1.90, 1.94] & $-$0.56 & 0.988 & 1.000 \\*
 & GPT-5.6-Terra & 90 & $-$0.69 & [$-$2.73, 1.30] & $-$0.17 & 0.514 & 1.000 \\*
 & Gemini-3-Flash & 90 & 0.72 & [$-$1.38, 2.89] & $-$0.05 & 0.514 & 1.000 \\*
\addlinespace[2pt]
\multirow{3}{*}{MiniMax-M3} & GLM-5.3-Flash & 95 & $-$1.33 & [$-$3.27, 0.64] & 0.00 & 0.183 & 1.000 \\*
 & GPT-5.6-Terra & 95 & $-$2.22 & [$-$3.60, $-$0.86] & $-$0.64 & 0.002 & 0.046 \\*
 & Gemini-3-Flash & 95 & $-$1.12 & [$-$3.61, 1.36] & 0.38 & 0.384 & 1.000 \\*
\addlinespace[2pt]
\multirow{2}{*}{GLM-5.3-Flash} & GPT-5.6-Terra & 95 & $-$0.89 & [$-$2.74, 0.88] & $-$0.17 & 0.339 & 1.000 \\*
 & Gemini-3-Flash & 95 & 0.21 & [$-$2.29, 2.69] & 0.26 & 0.870 & 1.000 \\*
\addlinespace[2pt]
GPT-5.6-Terra & Gemini-3-Flash & 96 & 1.09 & [$-$1.39, 3.51] & 0.51 & 0.392 & 1.000 \\
\midrule
\multicolumn{8}{@{}l}{\textbf{0DTE}} \\*
\multirow{7}{*}{DeepSeek-V4-Flash} & GPT-OSS-120B & 102 & $-$0.87 & [$-$2.12, 0.24] & 0.75 & 0.152 & 1.000 \\*
 & Qwen3-235B-A22B & 102 & $-$0.37 & [$-$1.19, 0.54] & $-$0.65 & 0.406 & 1.000 \\*
 & Llama-3.3-70B-Instruct & 102 & $-$0.14 & [$-$1.11, 0.83] & $-$0.23 & 0.785 & 1.000 \\*
 & MiniMax-M3 & 102 & $-$0.87 & [$-$1.65, $-$0.12] & $-$1.02 & 0.025 & 0.712 \\*
 & GLM-5.3-Flash & 102 & $-$0.84 & [$-$1.87, 0.05] & $-$0.33 & 0.089 & 1.000 \\*
 & GPT-5.6-Terra & 102 & $-$0.55 & [$-$1.28, 0.14] & $-$0.24 & 0.132 & 1.000 \\*
 & Gemini-3-Flash & 102 & $-$0.62 & [$-$1.39, 0.17] & $-$0.47 & 0.128 & 1.000 \\*
\addlinespace[2pt]
\multirow{6}{*}{GPT-OSS-120B} & Qwen3-235B-A22B & 102 & 0.50 & [$-$0.82, 1.94] & $-$1.61 & 0.483 & 1.000 \\*
 & Llama-3.3-70B-Instruct & 102 & 0.74 & [$-$0.59, 2.13] & $-$1.00 & 0.296 & 1.000 \\*
 & MiniMax-M3 & 102 & 0.00 & [$-$1.12, 1.18] & $-$2.18 & 1.000 & 1.000 \\*
 & GLM-5.3-Flash & 102 & 0.03 & [$-$1.08, 1.21] & $-$1.05 & 0.954 & 1.000 \\*
 & GPT-5.6-Terra & 102 & 0.32 & [$-$0.71, 1.44] & $-$1.27 & 0.567 & 1.000 \\*
 & Gemini-3-Flash & 102 & 0.25 & [$-$0.93, 1.50] & $-$1.39 & 0.684 & 1.000 \\*
\addlinespace[2pt]
\multirow{5}{*}{Qwen3-235B-A22B} & Llama-3.3-70B-Instruct & 102 & 0.23 & [$-$0.63, 0.99] & 0.70 & 0.598 & 1.000 \\*
 & MiniMax-M3 & 102 & $-$0.50 & [$-$1.03, 0.03] & $-$0.21 & 0.066 & 1.000 \\*
 & GLM-5.3-Flash & 102 & $-$0.47 & [$-$1.48, 0.42] & 0.10 & 0.352 & 1.000 \\*
 & GPT-5.6-Terra & 102 & $-$0.18 & [$-$1.01, 0.60] & 0.38 & 0.658 & 1.000 \\*
 & Gemini-3-Flash & 102 & $-$0.25 & [$-$0.94, 0.42] & $-$0.06 & 0.479 & 1.000 \\*
\addlinespace[2pt]
\multirow{4}{*}{Llama-3.3-70B-Instruct} & MiniMax-M3 & 102 & $-$0.73 & [$-$1.50, 0.07] & $-$1.46 & 0.068 & 1.000 \\*
 & GLM-5.3-Flash & 102 & $-$0.70 & [$-$1.91, 0.48] & $-$0.32 & 0.255 & 1.000 \\*
 & GPT-5.6-Terra & 102 & $-$0.41 & [$-$1.36, 0.55] & $-$0.49 & 0.389 & 1.000 \\*
 & Gemini-3-Flash & 102 & $-$0.48 & [$-$1.38, 0.44] & $-$0.89 & 0.305 & 1.000 \\*
\addlinespace[2pt]
\multirow{3}{*}{MiniMax-M3} & GLM-5.3-Flash & 102 & 0.03 & [$-$0.80, 0.77] & 0.59 & 0.939 & 1.000 \\*
 & GPT-5.6-Terra & 102 & 0.32 & [$-$0.30, 0.92] & 0.44 & 0.304 & 1.000 \\*
 & Gemini-3-Flash & 102 & 0.25 & [$-$0.33, 0.80] & 0.13 & 0.395 & 1.000 \\*
\addlinespace[2pt]
\multirow{2}{*}{GLM-5.3-Flash} & GPT-5.6-Terra & 102 & 0.29 & [$-$0.53, 1.18] & 0.14 & 0.524 & 1.000 \\*
 & Gemini-3-Flash & 102 & 0.22 & [$-$0.65, 1.18] & $-$0.30 & 0.652 & 1.000 \\*
\addlinespace[2pt]
GPT-5.6-Terra & Gemini-3-Flash & 102 & $-$0.07 & [$-$0.77, 0.63] & $-$0.32 & 0.846 & 1.000 \\
\end{longtable}
\endgroup

\subsection{Token Usage in Evaluation}

The eight-model panel consumes $888.54$~M tokens, $89.67$~M to $127.63$~M per model, dominated by input rather than output. The input-to-output ratio spans about $2{:}1$ for DeepSeek-V4-Flash to $78{:}1$ for Gemini-3-Flash. 0DTE is the largest footprint for every model at $60.34$--$93.53$~M tokens over $102$ sessions, with Overlay smaller and Earnings an order of magnitude smaller. Per-call input spans roughly $4.6$--$20.3$~K tokens by scenario, so call counts, input context and output length all contribute to cost differences.

\clearpage
\begin{table}[h]
\centering
\footnotesize
\renewcommand{\arraystretch}{1.25}
\caption{Token usage by model and scenario. Totals are in millions of tokens; per-call averages are in raw tokens, which in millions would round to zero. Counted from the recorded LLM calls of the selected runs; the per-model totals reproduce the official release's own token accounting exactly. The panel consumes 888.54~M tokens in total.}
\label{tab:token-merged}
\begin{adjustbox}{width=0.9\textwidth}
\begin{tabular}{ll r rrr rr}
\toprule
& & & \multicolumn{3}{c}{\textbf{Total Usage (M)}} & \multicolumn{2}{c}{\textbf{Per Call (tokens)}} \\
\cmidrule(lr){4-6} \cmidrule(l){7-8}
\textbf{Model} & \textbf{Scenario} & \textbf{Calls} & \textbf{In} & \textbf{Out} & \textbf{Total} & \textbf{In} & \textbf{Out} \\
\midrule
\multirow{4}{*}{DeepSeek-V4-Flash} & Overlay & 1,344 & 21.05 & 8.00 & 29.05 & 15,661 & 5,951 \\*
 & Earnings Bet & 404 & 2.21 & 2.84 & 5.05 & 5,478 & 7,025 \\*
 & 0DTE Intraday & 9,082 & 62.55 & 30.98 & 93.53 & 6,888 & 3,411 \\*
 & \textit{All Scenarios} & \textit{10,830} & \textit{85.81} & \textit{41.81} & \textit{127.63} & \textit{7,924} & \textit{3,861} \\
\midrule
\multirow{4}{*}{GPT-OSS-120B} & Overlay & 1,743 & 25.77 & 2.67 & 28.44 & 14,786 & 1,533 \\*
 & Earnings Bet & 434 & 2.10 & 0.60 & 2.70 & 4,833 & 1,387 \\*
 & 0DTE Intraday & 10,398 & 68.48 & 10.82 & 79.30 & 6,586 & 1,041 \\*
 & \textit{All Scenarios} & \textit{12,575} & \textit{96.35} & \textit{14.10} & \textit{110.44} & \textit{7,662} & \textit{1,121} \\
\midrule
\multirow{4}{*}{Qwen3-235B-A22B} & Overlay & 1,700 & 33.49 & 0.45 & 33.94 & 19,698 & 265 \\*
 & Earnings Bet & 546 & 2.92 & 0.16 & 3.09 & 5,352 & 300 \\*
 & 0DTE Intraday & 8,586 & 71.70 & 1.65 & 73.35 & 8,350 & 192 \\*
 & \textit{All Scenarios} & \textit{10,832} & \textit{108.11} & \textit{2.26} & \textit{110.37} & \textit{9,980} & \textit{209} \\
\midrule
\multirow{4}{*}{Llama-3.3-70B-Instruct} & Overlay & 1,640 & 24.47 & 0.41 & 24.87 & 14,918 & 247 \\*
 & Earnings Bet & 568 & 2.59 & 0.12 & 2.72 & 4,564 & 220 \\*
 & 0DTE Intraday & 10,010 & 66.38 & 1.18 & 67.57 & 6,632 & 118 \\*
 & \textit{All Scenarios} & \textit{12,218} & \textit{93.44} & \textit{1.71} & \textit{95.15} & \textit{7,648} & \textit{140} \\
\midrule
\multirow{4}{*}{MiniMax-M3} & Overlay & 1,590 & 23.74 & 7.43 & 31.16 & 14,928 & 4,671 \\*
 & Earnings Bet & 429 & 2.19 & 2.00 & 4.19 & 5,113 & 4,662 \\*
 & 0DTE Intraday & 9,691 & 65.69 & 19.01 & 84.71 & 6,779 & 1,962 \\*
 & \textit{All Scenarios} & \textit{11,710} & \textit{91.62} & \textit{28.44} & \textit{120.06} & \textit{7,824} & \textit{2,429} \\
\midrule
\multirow{4}{*}{GLM-5.3-Flash} & Overlay & 1,518 & 23.48 & 11.29 & 34.77 & 15,471 & 7,435 \\*
 & Earnings Bet & 425 & 2.16 & 3.93 & 6.09 & 5,073 & 9,255 \\*
 & 0DTE Intraday & 8,398 & 58.18 & 20.43 & 78.61 & 6,928 & 2,433 \\*
 & \textit{All Scenarios} & \textit{10,341} & \textit{83.82} & \textit{35.65} & \textit{119.47} & \textit{8,106} & \textit{3,448} \\
\midrule
\multirow{4}{*}{GPT-5.6-Terra} & Overlay & 1,781 & 25.94 & 1.18 & 27.11 & 14,563 & 660 \\*
 & Earnings Bet & 404 & 1.98 & 0.24 & 2.22 & 4,902 & 585 \\*
 & 0DTE Intraday & 9,016 & 58.32 & 2.02 & 60.34 & 6,469 & 224 \\*
 & \textit{All Scenarios} & \textit{11,201} & \textit{86.24} & \textit{3.43} & \textit{89.67} & \textit{7,699} & \textit{306} \\
\midrule
\multirow{4}{*}{Gemini-3-Flash} & Overlay & 1,364 & 27.69 & 0.27 & 27.96 & 20,301 & 197 \\*
 & Earnings Bet & 413 & 2.53 & 0.09 & 2.62 & 6,123 & 217 \\*
 & 0DTE Intraday & 9,735 & 84.07 & 1.10 & 85.17 & 8,635 & 113 \\*
 & \textit{All Scenarios} & \textit{11,512} & \textit{114.28} & \textit{1.46} & \textit{115.75} & \textit{9,927} & \textit{127} \\
\bottomrule
\end{tabular}
\end{adjustbox}
\end{table}

\newpage
\subsection{Visualizations}

\subsubsection{Portfolio Return for Overlay Scenarios}

\begin{figure}[H]
    \centering
    \includegraphics[width=0.93\linewidth]{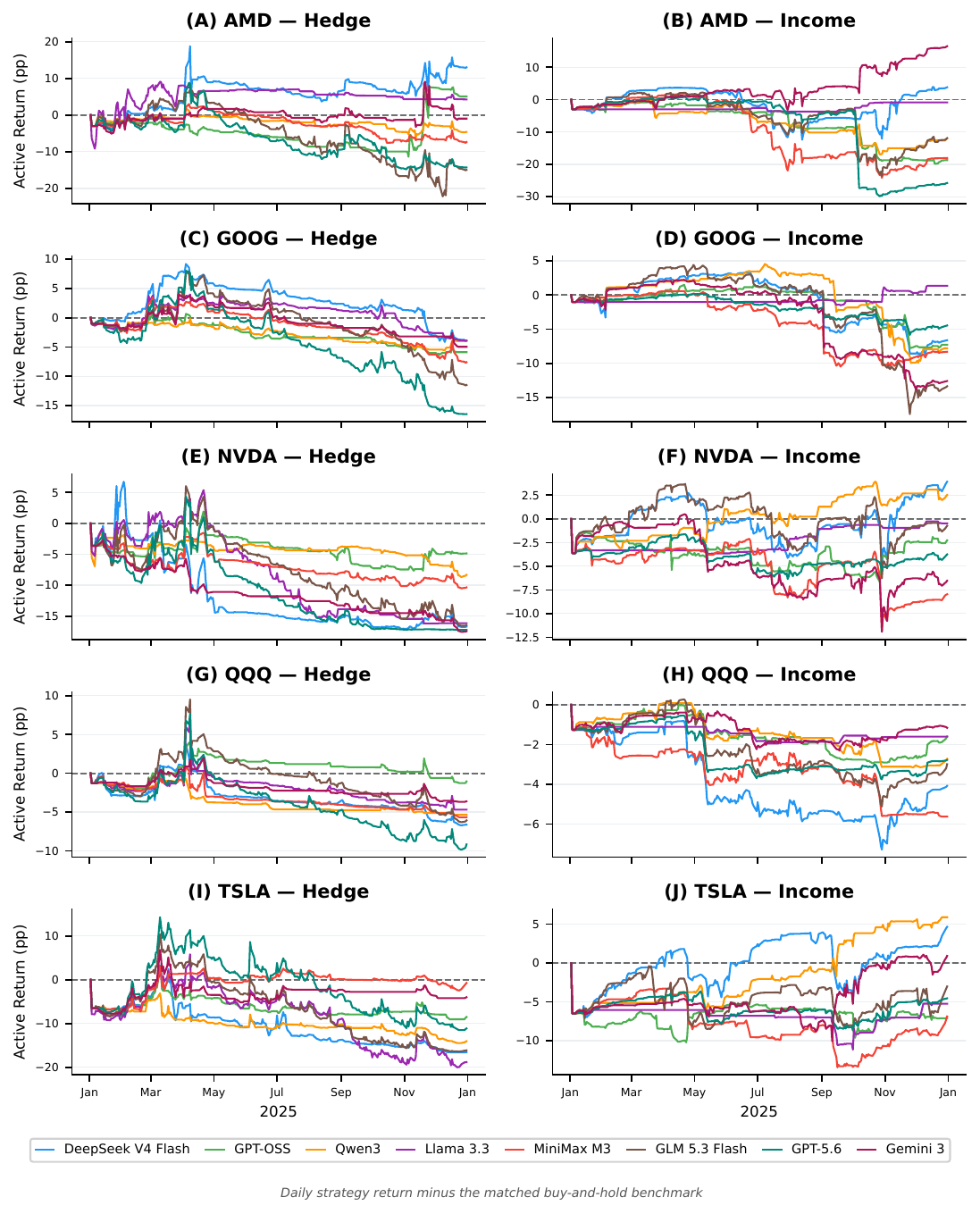}
    \caption{Daily active return of the option overlay against the matched buy-and-hold benchmark, for all five assets (rows) and both mandates, Hedge (left) and Income (right). Each curve is one model. The benchmark holds the run's own share count from the first close and leaves the remaining cash idle, which reproduces the released per-case benchmark return exactly. Flat segments are stretches in which a model carried no option position, so its portfolio tracked the benchmark.}
    \label{fig:overlay_active_return_grid}
\end{figure}

We visualize agent performance across the six provider-served models under
the overlay mandates. Figure~\ref{fig:overlay_active_return_grid} reports the full-year
active-return path for every asset and mode.


\subsubsection{Trading Behavior Across Mandates}

\begin{figure}[H]
    \centering
    \includegraphics[width=0.99\linewidth]{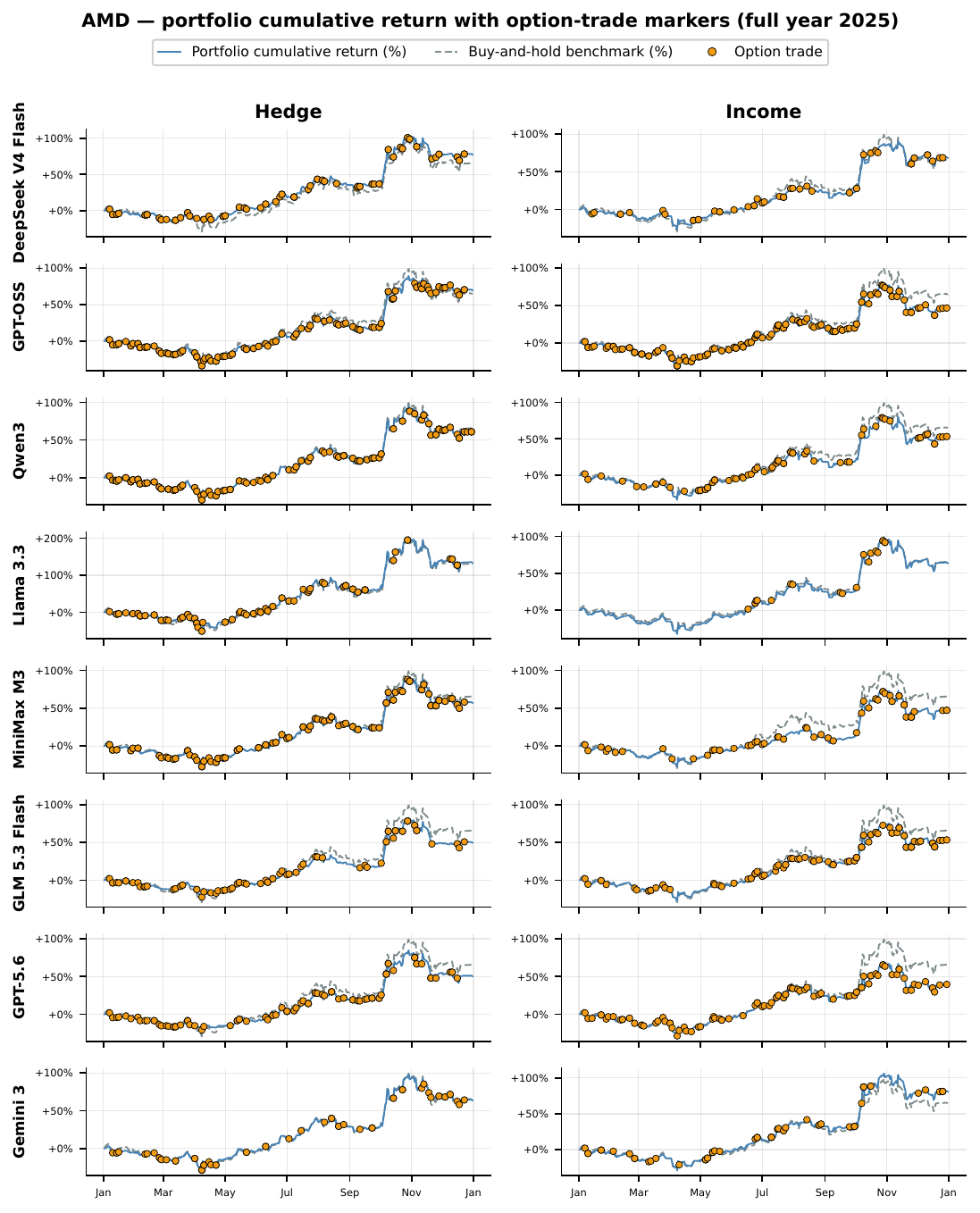}
    \caption{Portfolio cumulative return and option-trade activity for \textbf{AMD} across the six LLM agents under the \emph{hedge} (left) and \emph{income} (right) overlay mandates, 2025.}
    \label{fig:amd-portfolio-trades-2025}
\end{figure}

\begin{figure}[H]
    \centering
    \includegraphics[width=0.99\linewidth]{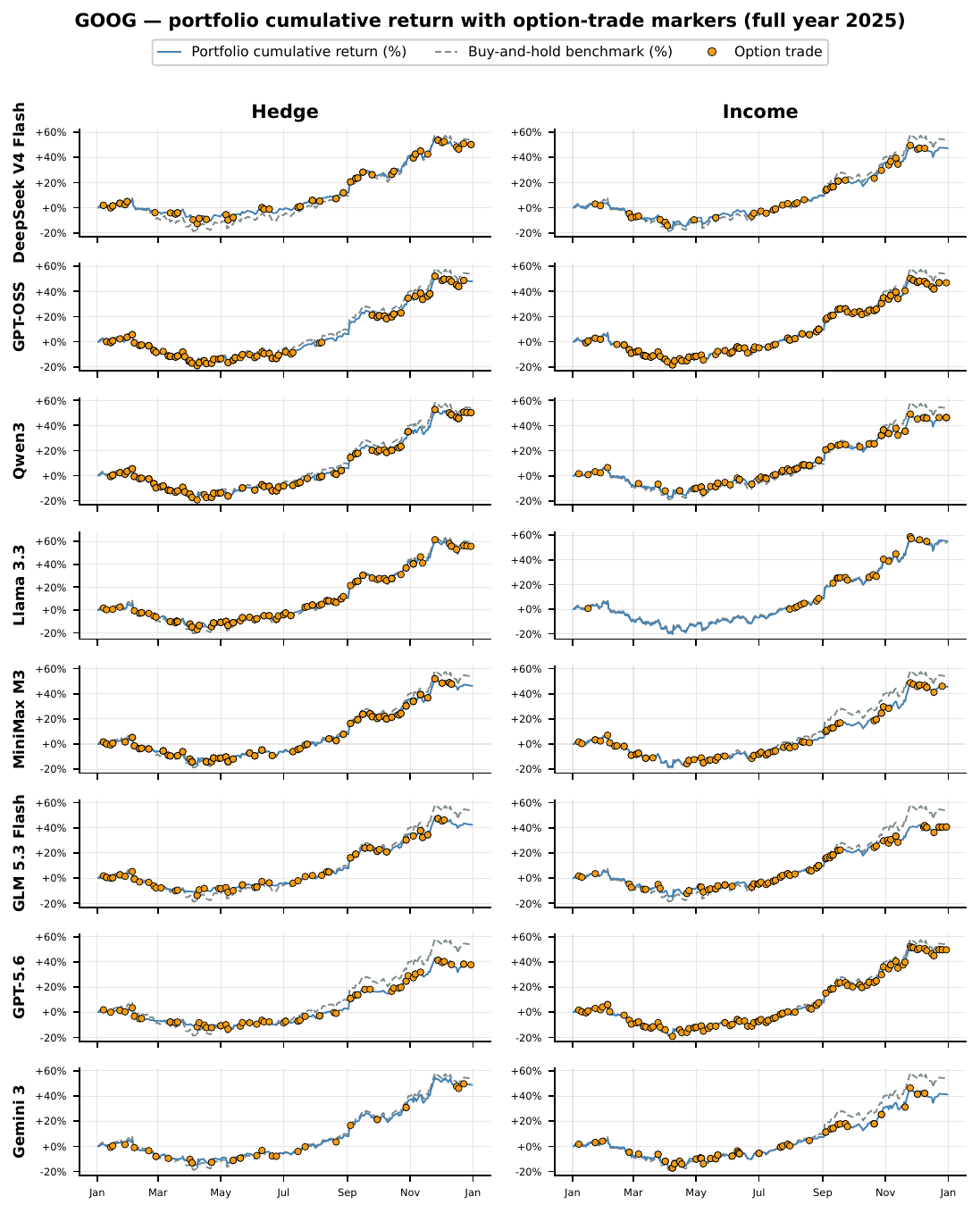}
    \caption{Portfolio cumulative return and option-trade activity for \textbf{GOOG}, as in Figure~\ref{fig:amd-portfolio-trades-2025}.}
    \label{fig:goog-portfolio-trades-2025}
\end{figure}

\begin{figure}[H]
    \centering
    \includegraphics[width=0.99\linewidth]{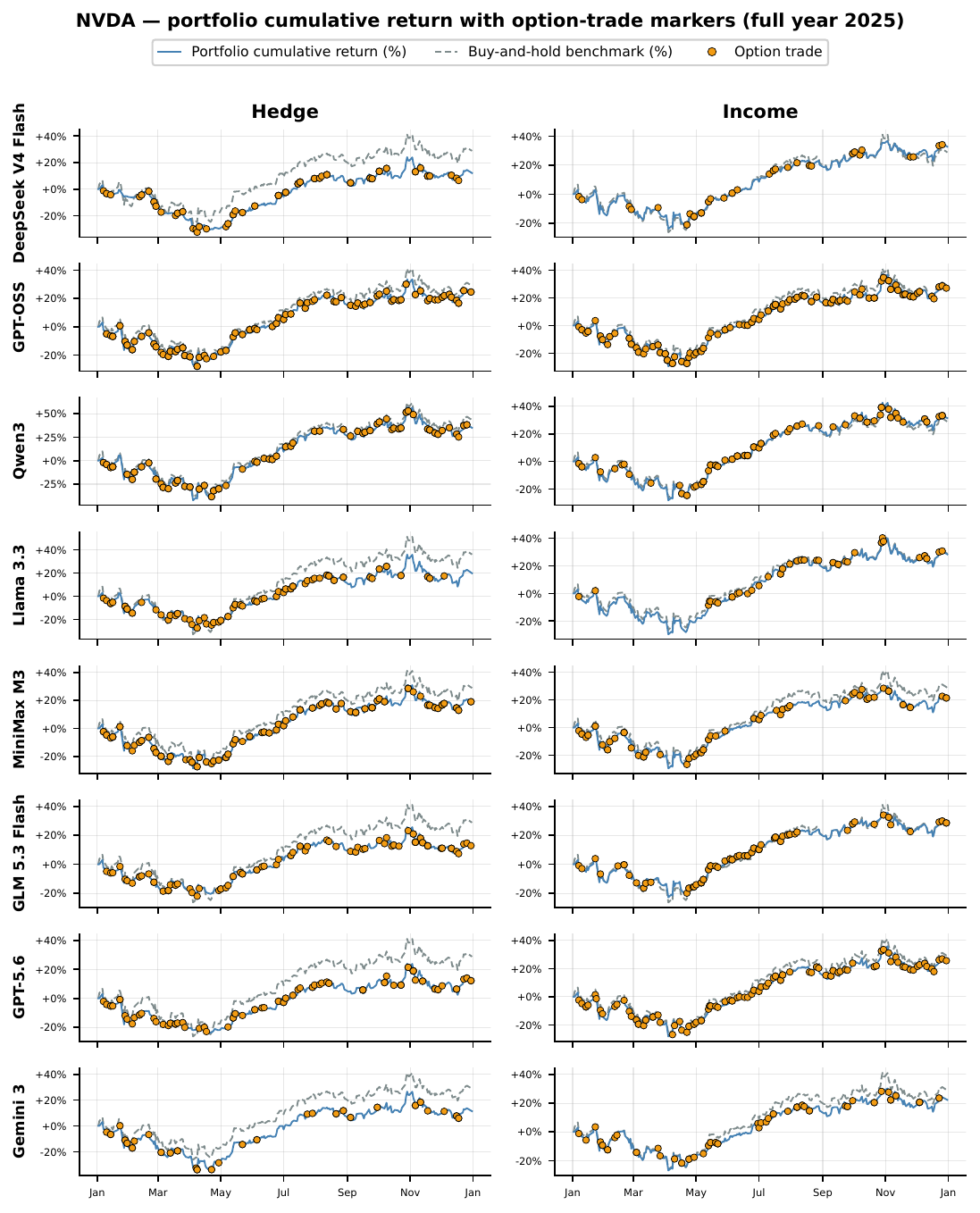}
    \caption{Portfolio cumulative return and option-trade activity for \textbf{NVDA}, as in Figure~\ref{fig:amd-portfolio-trades-2025}.}
    \label{fig:nvda-portfolio-trades-2025}
\end{figure}

\begin{figure}[H]
    \centering
    \includegraphics[width=0.99\linewidth]{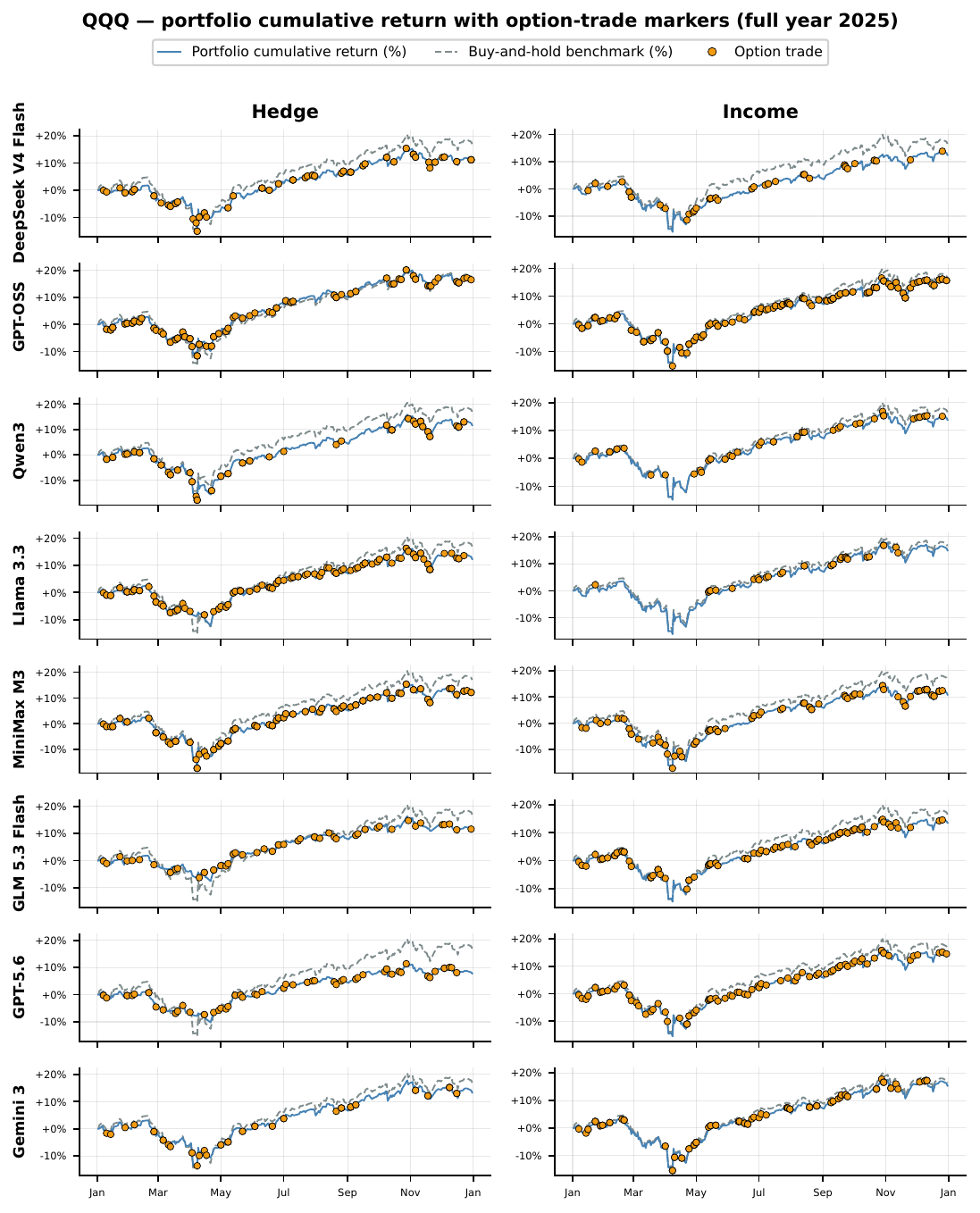}
    \caption{Portfolio cumulative return and option-trade activity for \textbf{QQQ}, as in Figure~\ref{fig:amd-portfolio-trades-2025}.}
    \label{fig:qqq-portfolio-trades-2025}
\end{figure}

\begin{figure}[H]
    \centering
    \includegraphics[width=0.99\linewidth]{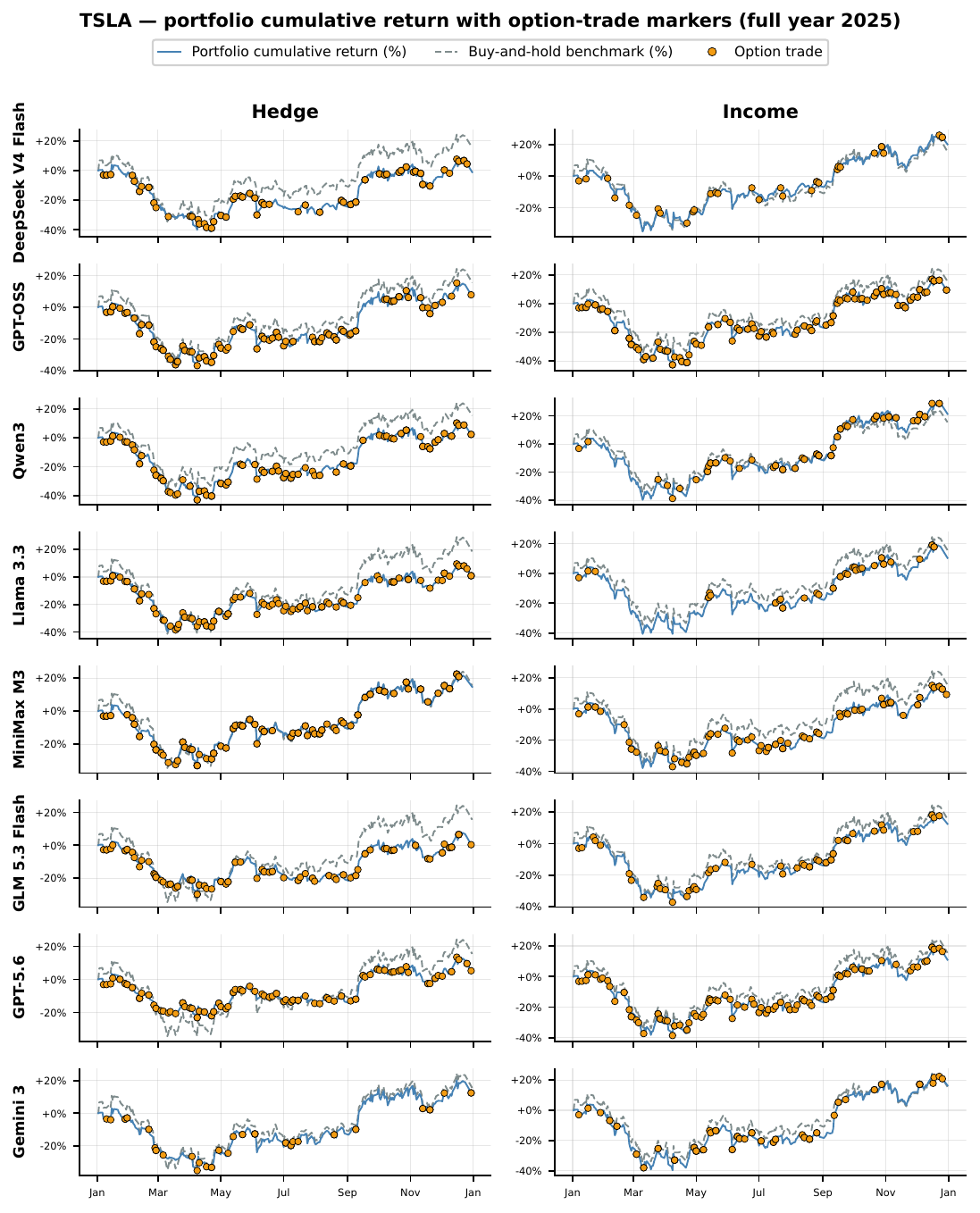}
    \caption{Portfolio cumulative return and option-trade activity for \textbf{TSLA}, as in Figure~\ref{fig:amd-portfolio-trades-2025}.}
    \label{fig:tsla-portfolio-trades-2025}
\end{figure}

Figures~\ref{fig:amd-portfolio-trades-2025}--\ref{fig:tsla-portfolio-trades-2025} show portfolio cumulative return and option-trade activity for the five underlyings across the six agents over 2025. Rows are agents, columns are the two mandates. Each panel plots cumulative return from the first equity snapshot of January, with a marker at every executed option trade and a dashed matched buy-and-hold line holding the run's own target share count, so the gap between the lines is the overlay's active return. The stock leg is fixed, so markers reflect overlay activity only. Every panel in this release carries option trades.

\subsubsection{Samples for Intraday Scenarios}

\begin{figure}[H]
    \centering
    \begin{subfigure}[b]{0.9\linewidth}
        \centering
        \includegraphics[width=\linewidth]{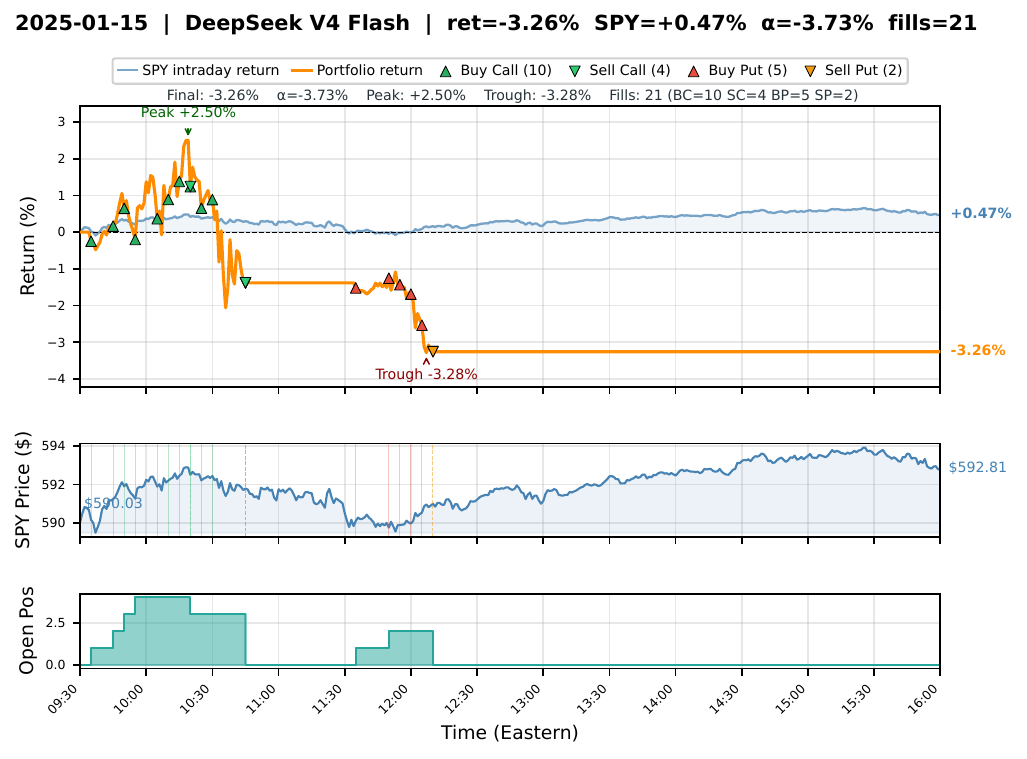}
    \end{subfigure}

    \vspace{4pt}

    \begin{subfigure}[b]{0.9\linewidth}
        \centering
        \includegraphics[width=\linewidth]{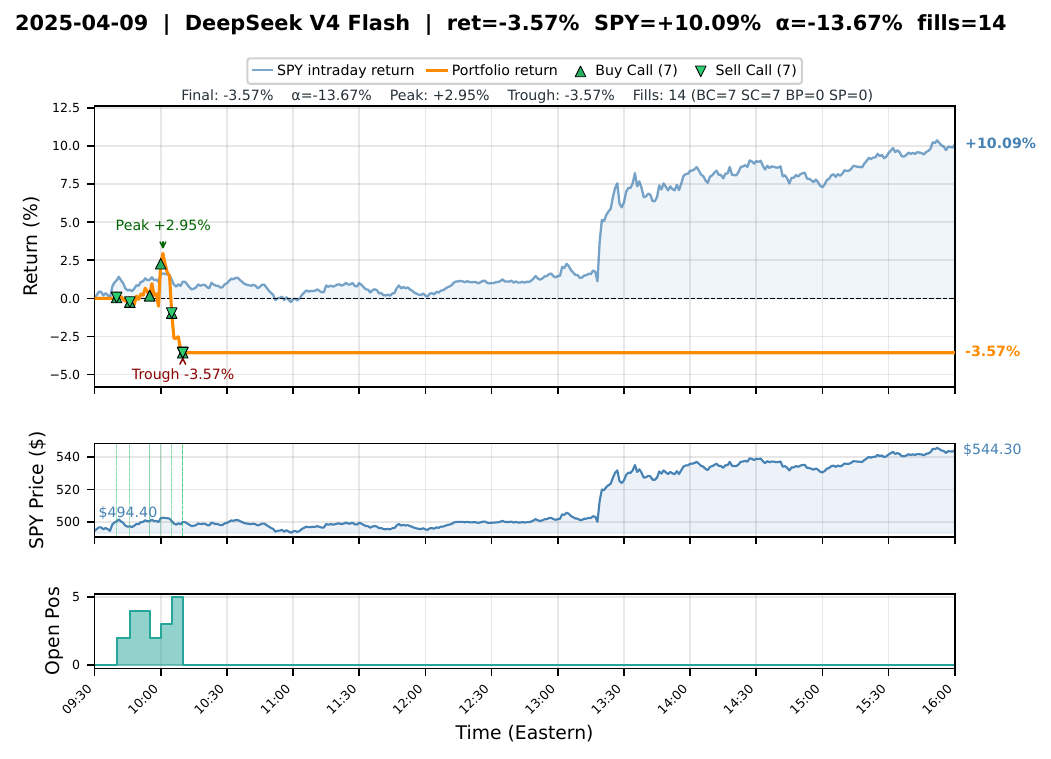}
    \end{subfigure}

    \caption{Intraday trading samples of DeepSeek V4 Flash, on 15 January 2025 (EST) and 9 April 2025 (EDT). Markers are buys as upward triangles and sells as downward triangles, green for calls and red or orange for puts, with per-side counts in the legend. Times are the Eastern session clock, converted from the recorded UTC instants with a DST-aware zone.}
    \label{fig:combined_intraday_ds}
\end{figure}

\begin{figure}[H]
    \centering
    \begin{subfigure}[b]{0.9\linewidth}
        \centering
        \includegraphics[width=\linewidth]{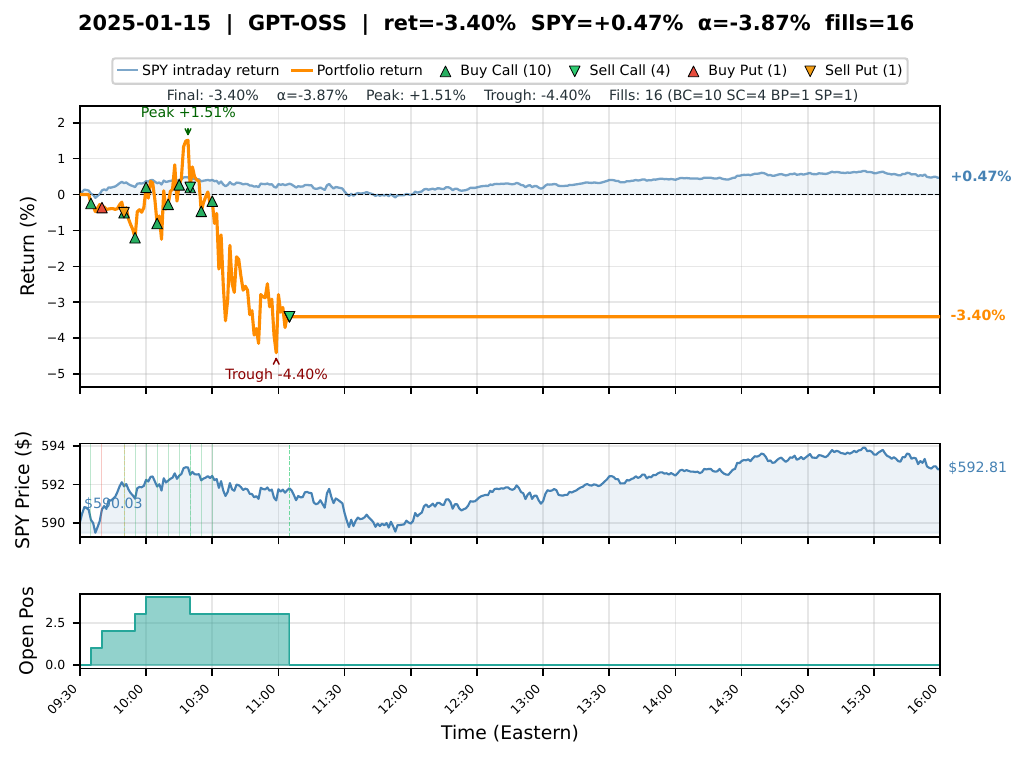}
    \end{subfigure}

    \vspace{4pt}

    \begin{subfigure}[b]{0.9\linewidth}
        \centering
        \includegraphics[width=\linewidth]{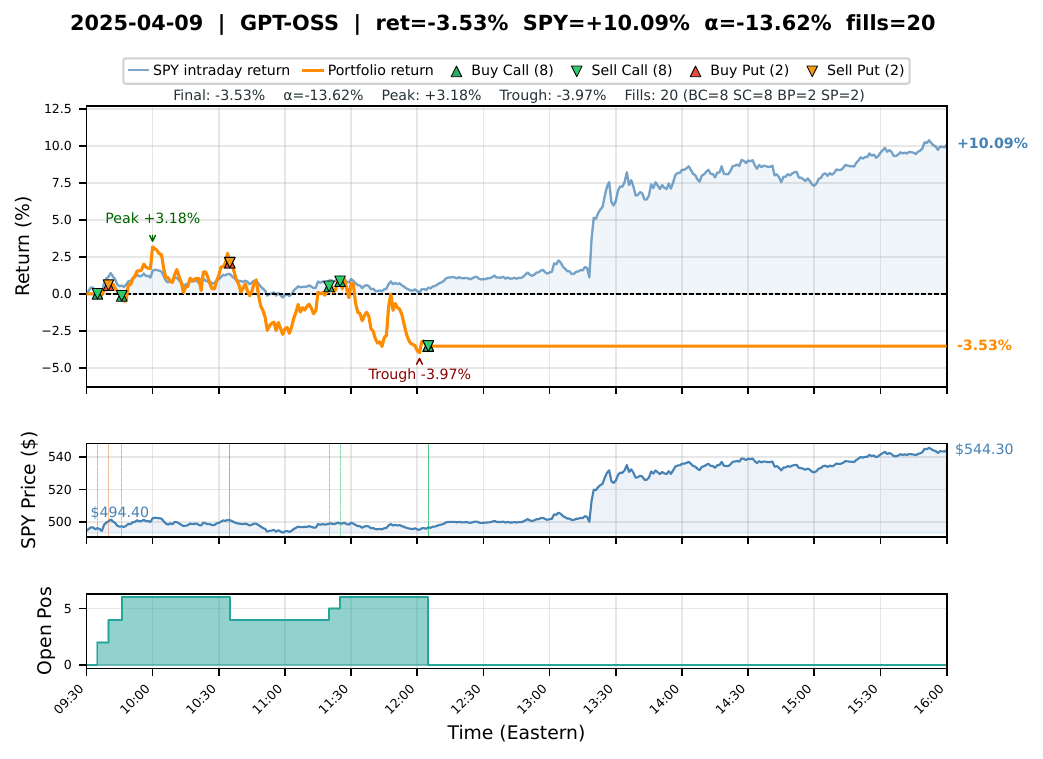}
    \end{subfigure}

    \caption{Intraday trading samples of GPT-OSS 120B, on the same two sessions as Figure~\ref{fig:combined_intraday_ds}.}
    \label{fig:combined_intraday_gptoss}
\end{figure}

We provide several recorded 0DTE sessions. Each panel set shows the SPY
intraday return against the portfolio return with every action marked, the
absolute SPY path with a guide at each fill, and the open option position
count. Figure~\ref{fig:combined_intraday_ds} shows DeepSeek-V4-Flash and
Figure~\ref{fig:combined_intraday_gptoss} shows GPT-OSS-120B on the same two
sessions. The two dates sit on either side of the 2025 daylight-saving change,
so the session clock can be checked in both regimes.

\section{Execution Quality and Quoted Liquidity in 0DTE}
\label{appendix:liquidity}

Every return in this paper is a return \emph{net of whatever execution cost the
simulator charged}, and that charge was never stated. This appendix establishes
it empirically against market data, characterises the liquidity of the
contracts the agents selected, and bounds how far the reported results depend
on the difference.

\paragraph{Data and validation.}
We join every recorded 0DTE fill to the prevailing OPRA quote for that exact
contract. The archive holds 15-minute National Best Bid and Offer snapshots for
SPY over the full evaluation period, with bid and ask price, displayed size at
each side, and the OCC contract identifier; quotes are restricted to contracts
expiring on the session date, so the join is 0DTE throughout. Of the $18{,}904$
option fills recorded across $102$ sessions and all eight
models, $18{,}890$
($99.93\%$) match a two-sided quote; the $14$ exceptions are contracts with no
two-sided market in the relevant bucket. Fill counts agree exactly with the
counts underlying Table~\ref{tab:consolidated_metrics} in $98.7\%$ of episodes.
Episode returns are taken on the fixed \$10{,}000 of episode capital, so one
percentage point of return is \$100 exactly.

Two properties of the archive shape the analysis and are stated here rather
than buried. First, the snapshot grid omits the top of each hour, giving $20$
buckets per session rather than $26$. Second, agents decide on a five-minute
clock, so only $4{,}023$ fills ($21.3\%$) land exactly on a snapshot. We
therefore separate the two questions the data can answer at different
strengths: the execution-rule question uses only the exactly-matched fills,
because it depends on the price \emph{level}, which moves within a bucket;
the liquidity and cost questions use all fills, because they depend on spread
\emph{width}, which is less sensitive to a short lag. The median quoted spread is $1.42\%$, $1.46\%$
and $1.61\%$ of mid for fills at lag zero, one to five minutes, and six to
fifteen minutes respectively, so the cost bound below is insensitive to the
$79\%$ of fills that sit a few minutes from a snapshot.

\paragraph{The simulator charges a nominal, not a realistic, spread.}
Recorded fill prices sit on a whole cent in $73.4\%$ of cases, and in exactly
those cases the applied slippage is zero. Where slippage is applied it is
\$0.0001 per contract and correctly signed: buyers pay above the round price
and sellers receive below it. The median half-spread actually quoted on these
contracts is \$0.0050, about $50$ times larger. The execution model is
therefore directionally right and economically negligible, and the returns
reported elsewhere in this paper should be read as excluding spread cost.

One artefact deserves explicit warning, because it invites the opposite
conclusion. Measured as a position within the quote, fills appear to cluster at
the touches and almost never at the mid. This is a consequence of tick
granularity, not of execution: $69.5\%$ of these quotes are one cent wide, and
a penny-wide quote has a half-cent mid that a whole-cent fill cannot occupy.
The clustering is not evidence that the simulator fills adversely.

\begin{table}[t]
\centering
\small
\setlength{\tabcolsep}{4pt}
\caption{0DTE execution quality against the prevailing OPRA 15-minute NBBO.
\emph{Spread} is the quoted spread as a percentage of mid and \emph{Depth} the
displayed size at the near touch, both medians over that model's fills.
\emph{Charge} is the mean per-episode cost of paying one half-spread on every
fill, in percentage points of the \$10{,}000 episode capital. \emph{Reported}
is the mean episode return as recorded, \emph{Charged} the same mean after that
cost with a $95\%$ bootstrap interval, and \emph{Full} the mean after a full
spread. $p$ is a two-sided sign-flip test that the charged mean is zero, Holm
adjusted across the eight models. For the pooled row, the interval and
unadjusted $p$ resample the 102 common trading dates jointly across models,
preserving within-date dependence.}
\label{tab:liquidity}
\begin{adjustbox}{max width=\linewidth}
\begin{tabular}{lrrrrrrrr}
\toprule
Model & Fills & Spread & Depth & Charge & Reported & Charged & $p_{\mathrm{Holm}}$ & Full \\
\midrule
DeepSeek-V4-Flash & 32.1 & 1.53\% & 136 & $-0.247$ & $-0.731$ & $-0.978$ {\scriptsize $[-1.63, -0.25]$} & 0.058 & $-1.226$ \\
GPT-OSS-120B & 33.6 & 1.48\% & 136 & $-0.300$ & $+0.141$ & $-0.159$ {\scriptsize $[-1.48, +1.26]$} & 1.000 & $-0.458$ \\
Qwen3-235B-A22B & 24.7 & 1.57\% & 100 & $-0.292$ & $-0.363$ & $-0.655$ {\scriptsize $[-1.10, -0.20]$} & 0.058 & $-0.946$ \\
Llama-3.3-70B-Instruct & 23.5 & 1.46\% & 98 & $-0.271$ & $-0.594$ & $-0.865$ {\scriptsize $[-1.67, +0.04]$} & 0.308 & $-1.136$ \\
MiniMax-M3 & 14.7 & 1.57\% & 126 & $-0.130$ & $+0.141$ & $+0.010$ {\scriptsize $[-0.50, +0.59]$} & 1.000 & $-0.120$ \\
GLM-5.3-Flash & 20.3 & 1.44\% & 116 & $-0.174$ & $+0.107$ & $-0.066$ {\scriptsize $[-0.97, +0.99]$} & 1.000 & $-0.240$ \\
GPT-5.6-Terra & 16.6 & 1.32\% & 104 & $-0.144$ & $-0.180$ & $-0.324$ {\scriptsize $[-1.01, +0.41]$} & 1.000 & $-0.469$ \\
Gemini-3-Flash & 19.7 & 1.55\% & 114 & $-0.185$ & $-0.111$ & $-0.296$ {\scriptsize $[-0.89, +0.29]$} & 1.000 & $-0.481$ \\
\midrule
Pooled & 23.1 & --- & --- & $-0.218$ & $-0.199$ & $-0.417$ {\scriptsize $[-0.95, +0.17]$} & 0.148 & $-0.634$ \\
\bottomrule
\end{tabular}
\end{adjustbox}
\end{table}

\paragraph{The selected contracts are liquid, so the half-spread is the
right bound.}
Across all fills the median quoted spread is \$0.010, or $1.49\%$ of mid, with
a $90$th percentile of $4.38\%$; the median displayed size at the near touch is
$116$ contracts. Recorded orders are small---the median order is one contract
and the largest is $17$---so displayed size covers the order in $99.75\%$ of
fills. Market impact beyond the touch is therefore not the binding
consideration at these sizes, and charging one half-spread per fill is an
appropriate bound rather than a conservative one. Per-model spread and depth
are given in Table~\ref{tab:liquidity}. Median relative spreads range from
$1.32\%$ to $1.57\%$ across the eight models, so both contract selection
and trading intensity contribute to model-level spread charges.

\paragraph{Charging the spread leaves no reliably positive mean.}
Table~\ref{tab:liquidity} recomputes the 0DTE episode means after charging each
fill the half-spread it was assumed not to pay. The charge is small in absolute
terms---a median of $0.160$ and a mean of $0.218$ percentage points per
episode---but the reported means are smaller still. All three models with a
positive reported mean lose it: only MiniMax-M3 remains above zero, at
$+0.010$ percentage points, with an interval including zero and a
sign-flip $p$-value above $0.9$. Under a full spread all eight models are
negative. GPT-5.6-Terra changes from $-0.180$ to $-0.324$ points after the
half-spread charge, and Gemini-3-Flash from $-0.111$ to $-0.296$.
No individual model is significant at $5\%$ after Holm correction across
eight models; the smallest adjusted value is about $0.058$.

Pooled across the panel, the mean moves from $-0.199$ to $-0.417$
percentage points. Pooled inference averages the eight model returns on
each common trading date and resamples or sign-flips those $102$ daily
averages, retaining dependence between models facing the same session.
The resulting interval includes zero. Treating all $816$ model--session
records as independent would give an artificially narrow interval and is
not the pooled test reported here. Spread charges worsen the observed
means but do not establish a statistically reliable negative pooled mean.

The charge also reorders the panel, and the mechanism is behavioural rather
than statistical. Trading intensity correlates $0.84$ with the per-episode
charge across the eight models, so the most active agents were the most
flattered by the nominal execution model: GPT-OSS-120B trades $33.6$ times per
session against MiniMax-M3's $14.7$, and falls from a share of first place on
reported means to third once the spread is charged. Any leaderboard computed
under a zero-spread execution model partly ranks turnover.

\paragraph{Scope, and what this does not establish.}
This is a re-pricing of recorded actions, not a counterfactual: it charges the
agents' own fills a realistic spread, and does not ask what an agent would have
decided had it observed one. The 15-minute snapshot resolution means queue
position, latency and partial-fill dynamics remain outside the analysis, and
displayed depth bounds market impact without measuring it. The analysis is
restricted to 0DTE because SPY coverage is complete there; the quote archive
omits one of the five Overlay underlyings entirely and $13$ of the $24$
Earnings names, so neither task supports an equivalent treatment, and we do not
report a partial one. Extending liquidity-aware evaluation beyond re-pricing
requires the agent to see quoted spread and depth at decision time; we plan to
support this through optional deterministic L1 replay in the data layer, which
would let spread and displayed size enter the observation and the feasibility
constraints rather than only the post-hoc accounting.

\clearpage
\section*{NeurIPS Paper Checklist}

\begin{enumerate}

\item {\bf Claims}
    \item[] Question: Do the main claims made in the abstract and introduction accurately reflect the paper's contributions and scope?
    \item[] Answer: \answerYes{} %
    \item[] Justification: {As shown in Abstract and Introduction.}

\item {\bf Limitations}
    \item[] Question: Does the paper discuss the limitations of the work performed by the authors?
    \item[] Answer: \answerYes{} %
    \item[] Justification:  We thoroughly discussed the potential impact of our work in Appendix.

\item {\bf Theory assumptions and proofs}
    \item[] Question: For each theoretical result, does the paper provide the full set of assumptions and a complete (and correct) proof?
    \item[] Answer: \answerNA{}{} %
    \item[] Justification:  The paper does not include theoretical results; the formulas in the Appendix are descriptive and accompanied by corresponding explanations.

    \item {\bf Experimental result reproducibility}
    \item[] Question: Does the paper fully disclose all the information needed to reproduce the main experimental results of the paper to the extent that it affects the main claims and/or conclusions of the paper (regardless of whether the code and data are provided or not)?
    \item[] Answer: \answerYes{} %
    \item[] Justification: As detailed in Experimental Setting (Section~\ref{sec:experimental}) and the open-source code with a sample dataset. 

\item {\bf Open access to data and code}
    \item[] Question: Does the paper provide open access to the data and code, with sufficient instructions to faithfully reproduce the main experimental results, as described in supplemental material?
    \item[] Answer: \answerYes{} %
    \item[] Justification:  We provide a publicly accessible repository for data sample, code, and instructions.

\item {\bf Experimental setting/details}
    \item[] Question: Does the paper specify all the training and test details (e.g., data splits, hyperparameters, how they were chosen, type of optimizer) necessary to understand the results?
    \item[] Answer: \answerYes{} %
    \item[] Justification: We discuss in Section~\ref{sec:experimental} and Section~\ref{appendix:data}.

\item {\bf Experiment statistical significance}
    \item[] Question: Does the paper report error bars suitably and correctly defined or other appropriate information about the statistical significance of the experiments?
    \item[] Answer: \answerYes{} %
    \item[] Justification: Every reported mean carries a percentile 95\% bootstrap interval from 10{,}000 episode resamples. Cross-model and ablation comparisons are paired on shared episodes and tested with two-sided Monte Carlo sign-flip tests using 50{,}000 draws, with Holm correction applied within each comparison family; the adjusted $p$-values are reported in Section~\ref{sec:analysis} and Appendix~\ref{appendix:additional_results}. The intervals quantify variation across the recorded 2025 market episodes conditional on one selected rollout per configuration, and therefore do not measure repeated-generation variability; this is stated as a limitation.

\item {\bf Experiments compute resources}
    \item[] Question: For each experiment, does the paper provide sufficient information on the computer resources (type of compute workers, memory, time of execution) needed to reproduce the experiments?
    \item[] Answer: \answerYes{} %
    \item[] Justification: Our work does not need to train models and only needs to conduct model inference. We call a single hosted inference provider for six of the eight evaluated models and a separate gateway for the other two, and report token usage for each, in millions, in Table~\ref{tab:token-merged}.
    
\item {\bf Code of ethics}
    \item[] Question: Does the research conducted in the paper conform, in every respect, with the NeurIPS Code of Ethics \url{https://neurips.cc/public/EthicsGuidelines}?
    \item[] Answer: \answerYes{} %
    \item[] Justification:  We thoroughly discussed the potential impact of our work in Appendix, and ensured the compliance with the NeurIPS code of ethics.

\item {\bf Broader impacts}
    \item[] Question: Does the paper discuss both potential positive societal impacts and negative societal impacts of the work performed?
    \item[] Answer: \answerYes{} %
    \item[] Justification: We thoroughly discussed the potential impact of our work in Appendix~\ref{appendix:board}.
    
\item {\bf Safeguards}
    \item[] Question: Does the paper describe safeguards that have been put in place for responsible release of data or models that have a high risk for misuse (e.g., pre-trained language models, image generators, or scraped datasets)?
    \item[] Answer: \answerNA{} %
    \item[] Justification: Our data or models don't have risk for misuse.

\item {\bf Licenses for existing assets}
    \item[] Question: Are the creators or original owners of assets (e.g., code, data, models), used in the paper, properly credited and are the license and terms of use explicitly mentioned and properly respected?
    \item[] Answer: \answerYes{} %
    \item[] Justification:  We have fully complied with the licensing terms and usage policies of all third-party datasets and assets used in this work.

\item {\bf New assets}
    \item[] Question: Are new assets introduced in the paper well documented and is the documentation provided alongside the assets?
    \item[] Answer: \answerYes{} %
    \item[] Justification:  See supplementary and our code repository.

\item {\bf Crowdsourcing and research with human subjects}
    \item[] Question: For crowdsourcing experiments and research with human subjects, does the paper include the full text of instructions given to participants and screenshots, if applicable, as well as details about compensation (if any)? 
    \item[] Answer: \answerNA{} %
    \item[] Justification: This work doesn't include this kind of experiment.

\item {\bf Institutional review board (IRB) approvals or equivalent for research with human subjects}
    \item[] Question: Does the paper describe potential risks incurred by study participants, whether such risks were disclosed to the subjects, and whether Institutional Review Board (IRB) approvals (or an equivalent approval/review based on the requirements of your country or institution) were obtained?
    \item[] Answer: \answerNA{} %
    \item[] Justification: This work doesn't include this kind of experiment.

\item {\bf Declaration of LLM usage}
    \item[] Question: Does the paper describe the usage of LLMs if it is an important, original, or non-standard component of the core methods in this research? Note that if the LLM is used only for writing, editing, or formatting purposes and does \emph{not} impact the core methodology, scientific rigor, or originality of the research, declaration is not required.
    \item[] Answer: \answerYes{} %
    \item[] Justification: LLMs are a core component of our research methodology, which focuses on benchmarking and evaluating their performance in real-time investment.

\end{enumerate}

\end{document}